%% file: top.tex
\documentclass[10pt,twocolumn,letterpaper]{article}

\usepackage{cvpr}              

\input{preamble}
\definecolor{cvprblue}{rgb}{0.21,0.49,0.74}
\usepackage[pagebackref,breaklinks,colorlinks,citecolor=cvprblue]{hyperref}

\def\paperID{867} 
\def\confName{CVPR}
\def\confYear{2024}

\title{\methodName: Disentangling the Dynamic and Static World for End-to-End Driving}

\author{
Simon Doll$^{1,2}$, \quad Niklas Hanselmann$^{1,2}$, \quad Lukas Schneider$^{1}$, \quad Richard Schulz$^{1}$,\\
\quad Marius Cordts$^{1}$, \quad Markus Enzweiler$^3$, \quad Hendrik P.A. Lensch$^2$ \\
 \\
$^1$Mercedes-Benz AG, \quad $^2$University of Tübingen,
$^3$ Esslingen University of Applied Sciences\\
}

\begin{document}
\maketitle
\input{sections/0_abstract}    
\input{sections/1_intro}
\input{sections/2_related}
\input{sections/3_method}

\input{sections/4_experiments}
\input{sections/5_conclusion}

{
    \small
    \bibliographystyle{ieeenat_fullname}
    \bibliography{bibliography_long, bibliography}
}

\input{sections/X_suppl}

\end{document}

%% file: preamble.tex
\usepackage[accsupp]{axessibility}
\usepackage[dvipsnames]{xcolor}
\usepackage[nolist, nohyperlinks]{acronym}
\usepackage{amsmath,amssymb, mathtools}
\usepackage{pifont}

\usepackage{colortbl}
\usepackage{adjustbox}

\usepackage{tikz}
\usepackage{fontawesome}

\input{shortcuts}
\input{acronyms}

\input{tikz_shortcuts}

\usepackage{siunitx}

%% file: shortcuts.tex
\newcommand{\methodName}{\textsc{DualAD}}
\newcommand{\methodNameSOne}{\textsc{DualAD-I}}
\newcommand{\methodNameSTwo}{\textsc{DualAD-II}}

\newcommand{\uniadSOne}{UniAD-I}
\newcommand{\uniadSTwo}{UniAD-II}

\newcommand{\cmark}{\ding{51}}%
\newcommand{\xmark}{\ding{55}}%

\newcommand{\tf}[2]{{}^{#1}{\mathbf{T}_{#2}}}

\newcommand{\figref}[1]{Fig.~\ref{#1}}
\newcommand{\secref}[1]{Section~\ref{#1}}

\newcommand{\tabref}[1]{Table~\ref{#1}}

\def\mb{\mathbf}

\makeatletter
\DeclareRobustCommand\onedot{\futurelet\@let@token\@onedot}
\def\@onedot{\ifx\@let@token.\else.\null\fi\xspace}
\def\eg{e.g\onedot} 
\def\ie{i.e\onedot}

\makeatother

\newcommand{\boldparagraph}[1]{\vspace{0.2cm}\noindent{\bf #1:} }

\definecolor{darkgreen}{rgb}{0,0.7,0}
\definecolor{tablegrey}{gray}{0.8}

\definecolor{ellisred}{rgb}{0.87,0.44,0.38} 
\definecolor{ellisgreen}{rgb}{0.69,0.90,0.52} 
\definecolor{elliscyan}{rgb}{0.29,0.77,0.74} 
\definecolor{ellisorange}{rgb}{0.89,0.55,0.28} 
\definecolor{ellisblue}{rgb}{0.41,0.61,0.86} 

\definecolor{Gray}{gray}{0.9}
\newcommand{\highlight}{\cellcolor{Gray}}

%% file: acronyms.tex
\begin{acronym}
    \acro{bev}[BEV]{bird's-eye view}
    \acro{sota}[SOTA]{\textit{state-of-the-art}}
    \acro{map}[mAP]{\textit{mean Average Precision}}
    \acro{nds}[NDS]{\textit{nuScenes Detection Score}}
    \acro{mate}[mATE]{\textit{mean Average Translation Error}}
    \acro{maoe}[mAOE]{\textit{mean Average Orientation Error}}
    \acro{mase}[mASE]{\textit{mean Average Scale Error}}
    \acro{mave}[mAVE]{\textit{mean Average Velocity Error}}
    \acro{amota}[AMOTA]{\textit{Average Multi Object Tracking Accuracy}}
    \acro{amotp}[AMOTP]{\textit{Average Multi Object Tracking Precision}}
    \acro{ids}[IDS]{\textit{number of identity switches}}
    \acro{iou}[IoU]{\textit{Intersection over Union}}
    \acro{epa}[EPA]{\textit{End-to-end Prediction Accuracy}}
    \acro{minade}[minADE]{\textit{minimum Average Displacement Error}}
    \acro{minfde}[minFDE]{\textit{minimum Final Displacement Error}}
    \acro{tid}[TID]{\textit{Track Initialization Duration}}
    \acro{lgd}[LGD]{\textit{Longest Gap Duration}}
    
\end{acronym}

%% file: tikz_shortcuts.tex
\definecolor{darkmidnightblue}{rgb}{0.0, 0.2, 0.4}
\definecolor{gold}{rgb}{0.706,0.627,0.412}
\definecolor{red}{rgb}{0.647,0.117,0.215}
\definecolor{anthracite}{rgb}{0.196, 0.255, 0.294}
\definecolor{blue}{rgb}{0.000, 0.412, 0.667}
\definecolor{light blue}{rgb}{0.314, 0.667, 0.784}
\definecolor{dark blue}{rgb}{0.255, 0.353, 0.549}
\definecolor{dark green}{rgb}{0.196, 0.431, 0.118}
\definecolor{light green}{rgb}{0.490, 0.647, 0.294}
\definecolor{violet}{rgb}{0.686, 0.431, 0.588}
\definecolor{rust}{rgb}{0.784, 0.314, 0.235}
\definecolor{brown}{rgb}{0.569, 0.412, 0.275}

\definecolor{mycolor}{rgb}{1,0.2,0.3}

\definecolor{sns_blue}{rgb}{0.12156863, 0.46666667, 0.70588235}
\definecolor{sns_orange}{rgb}{1.        , 0.49803922, 0.05490196}
\definecolor{sns_green}{rgb}{0.17254902, 0.62745098, 0.17254902}
\definecolor{sns_red}{rgb}{0.83921569, 0.15294118, 0.15686275}
\definecolor{sns_purple}{rgb}{0.58039216, 0.40392157, 0.74117647}
\definecolor{sns_brown}{rgb}{0.54901961, 0.3372549 , 0.29411765}
\definecolor{sns_rose}{rgb}{0.89019608, 0.46666667, 0.76078431}
\definecolor{sns_grey}{rgb}{0.49803922, 0.49803922, 0.49803922}
\definecolor{sns_light_green}{rgb}{0.7372549 , 0.74117647, 0.13333333}
\definecolor{sns_light_blue}{rgb}{0.09019608, 0.74509804, 0.81176471}

\definecolor{sns_dark_grey}{rgb}{0,0.2,0.2}
\definecolor{task_dark_anthracite}{rgb}{0.192, 0.263, 0.329}

\definecolor{light_grey}{rgb}{0.502, 0.490, 0.443}
\definecolor{light_orange}{rgb}{100., 0.882, 0.784}
\definecolor{track_query_goldgrey}{rgb}{0.949, 0.929, 0.863}
\definecolor{track_query_blue}{rgb}{0.396, 0.706, 0.796}
\definecolor{track_query_dark_blue}{rgb}{0.239, 0.604, 0.710}
\definecolor{detection_query_anthracite}{rgb}{0.392, 0.463, 0.529}
\definecolor{decoder_track_query_blue}{rgb}{0.094, 0.325, 0.486}
\definecolor{decoder_track_query_dark_blue}{rgb}{0.075, 0.259, 0.388}
\definecolor{lmm_green}{rgb}{0.337, 0.655, 0.416}
\definecolor{lmm_dark_green}{rgb}{0.267, 0.522, 0.329}
\definecolor{gmm_blue}{rgb}{0.298, 0.447, 0.690}
\definecolor{gmm_dark_blue}{rgb}{0.235, 0.357, 0.549}
\definecolor{gt_query_red}{rgb}{0.765, 0.314, 0.333}
\definecolor{gt_query_dark_red}{rgb}{0.639, 0.220, 0.235}

\definecolor{detection_query_grey}{rgb}{0.878, 0.890, 0.906}

\usetikzlibrary{%
    arrows,%
    shapes,%
    chains,%
    calligraphy,%
    fadings,%
    matrix,%
    positioning,
    scopes,%
    shadows, %
    calc, %
    decorations,%
    decorations.text,%
    decorations.pathmorphing,%
    decorations.pathreplacing,%
    shapes.arrows,%
    math
}

%% file: sections/0_abstract.tex
\begin{abstract}
State-of-the-art approaches for autonomous driving integrate multiple sub-tasks of the overall driving task into a single pipeline that can be trained in an end-to-end fashion by passing latent representations between the different modules.
In contrast to previous approaches that rely on a unified grid to represent the belief state of the scene, we propose dedicated representations to disentangle dynamic agents and static scene elements. This allows us to explicitly compensate for the effect of both ego and object motion between consecutive time steps and to flexibly propagate the belief state through time. Furthermore, dynamic objects can not only attend to the input camera images, but also directly benefit from the inferred static scene structure via a novel dynamic-static cross-attention. Extensive experiments on the challenging nuScenes benchmark demonstrate the benefits of the proposed dual-stream design, especially for modelling highly dynamic agents in the scene, and highlight the improved temporal consistency of our approach. Our method titled DualAD not only outperforms independently trained single-task networks, but also improves over previous state-of-the-art end-to-end models by a large margin on all tasks along the functional chain of driving.
\end{abstract}

%% file: sections/1_intro.tex
\section{Introduction}\label{sec:introduction}
Autonomous systems have evolved from strictly modular and largely hand-crafted pipelines towards a more holistic learning-centric paradigm~\cite{sadat2020perceive, casas2021mp3}. While the former relies on explicitly defined interfaces between modules, the latter tackles the entire driving task in an end-to-end fashion.
Nevertheless, recent work has shown the benefits of retaining a modular structure including typical sub-tasks such as perception, prediction and planning while allowing latent features to serve as interfaces between the modules~\cite{hu2023planning, jiang2023vad}.

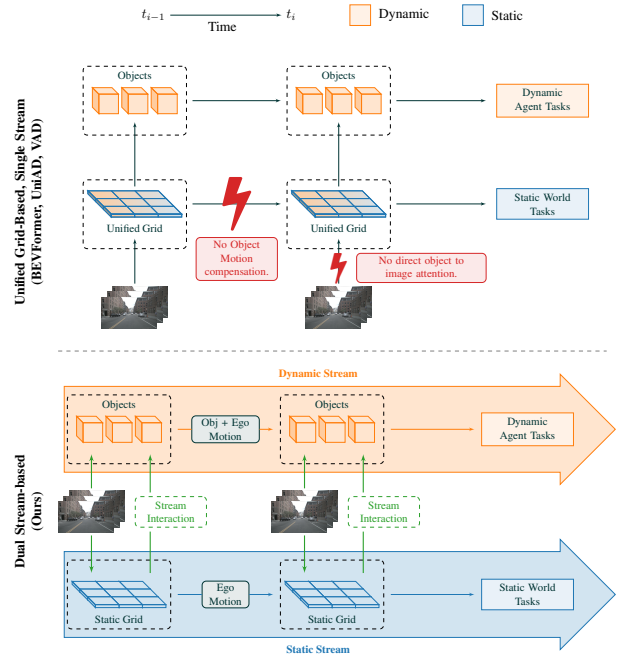
\begin{figure}[t]
    \centering
    \resizebox{\columnwidth}{!}{%
      \input{diagrams/teaser.tikz}
    }
    \caption{\textbf{Comparison of Representation Design} of unified grid-based approaches and our dual-stream design. 
    By explicitly disentangling dynamic and static representations, the dynamic stream can aggregate highly descriptive features. This is achieved through direct attention to image features, as well as explicit compensation for object and ego motion, which is not feasible with unified grids.}
    \label{fig:teaser}
\end{figure}

In contrast to independent, task-specific modules with fixed pre-defined interfaces, an end-to-end approach enables the
joint optimization of the entire pipeline, learning not only the parameters in each module but also \textit{the interfaces between them}. 
The chosen space of each module's latent representation restricts the set of interfaces which can be learned, allowing to model inductive biases about the scene structure, \eg consistent motion of dynamic elements, or to incorporate task specific properties.
However, this places additional importance on choosing well-suited intermediate representations, since they heavily affect information flow and the performance of subsequent modules. Hence, these representations should be carefully tailored to their corresponding semantic entities in the driving scene to achieve a high performing end-to-end architecture.

To model dynamic agents in the scene, a prevalent approach is to leverage attention with object-centric queries that detect an individual object in the environment~\cite{wang2022detr3d, liu2022petr, doll2022spatialdetr}. Furthermore, recent works~\cite{zhang2022mutr3d, doll2023star, wang2023exploring} have demonstrated the benefits of incorporating temporal information to consistently model object dynamics and to account for temporal occlusions. In such work, object-queries provide dedicated latent representations that each describe a single object. Its belief state can then be propagated through time by explicitly compensating for the ego and estimated object motion between two consecutive timestamps~\cite{doll2023star, wang2023exploring}. 

The most common alternative is to use \ac{bev}-grid queries~\cite{li2022bevformer, hu2023planning, jiang2023vad} as an intermediate representation, with subsequent tasks relying solely on this representation.
However, such a grid is not coupled to semantic instances and instead represents a spatial area of the scene. Hence, the motion of agents cannot be explicitly modeled and compensated for, see~\figref{fig:teaser}. This is due to the fact that each grid-cell could potentially represent multiple entities with different rigid motion transforms or even completely static elements, depending on grid resolution and object sizes.  
While grid-based representations are well-suited for static world perception~\cite{li2022bevformer, liao2022maptr}, exclusively relying on them to aggregate sensor measurements and temporal information hampers the perception of highly dynamic agents. 

\boldparagraph{Contributions} In this work, we propose a dual-stream approach to leverage the potential of object-centric representations for dynamic agents combined with a \ac{bev}-grid representation for static scene elements.
This dual-stream design explicitly applies object and ego motion compensation to dynamic agents and allows object-queries and \ac{bev}-queries to simultaneously attend to the camera images of the current timestamp. Besides self-attention and cross-attention with camera images, we introduce a new dynamic-static cross-attention-block that allows object-queries to attend to the \ac{bev}-queries fostering the consistency between the streams.

Our proposed approach termed \methodName\ allows for robust and temporally consistent perception.
On the challenging nuScenes dataset~\cite{caesar2020nuscenes} \methodName\ outperforms specialized \ac{sota} models for various perception tasks by a large margin. The integration with recent end-to-end frameworks~\cite{hu2023planning, jiang2023vad} reveals the importance of disentangled representations for dynamic agents and static world elements, and exhibits significant performance gains along the entire functional chain.
Extensive ablation studies highlight the importance of the dual-stream design for all driving tasks, while especially improving temporal consistency and the perception of highly dynamic agents. 


%% file: diagrams/teaser.tikz
{
\tikzset{
    static_world_tasks/.style={
        rectangle,
        fill = sns_blue!10,
        draw = sns_blue,
        text=sns_dark_grey,
        font=\fontsize{8}{10}\selectfont,
        text width=60pt,
        align=center
    },
    dynamic_world_tasks/.style={
        rectangle,
        fill = sns_orange!10,
        draw = sns_orange,
        text=sns_dark_grey,
        font=\fontsize{8}{10}\selectfont,
        text width=60pt,
        align=center
    },
    static_square/.style={
        rectangle,
        fill = sns_blue!10,
        draw = sns_blue,
        minimum width=15pt,
        minimum height=15pt
    },
    dynamic_square/.style={
        rectangle,
        fill = sns_orange!10,
        draw = sns_orange,
        minimum width=15pt,
        minimum height=15pt
    },
    label/.style={
        rectangle,
        rounded corners=3pt,
        inner sep=2pt,
        fill = sns_dark_grey!10,
        draw = sns_dark_grey,
        text=sns_dark_grey,
        font=\fontsize{8}{10}\selectfont,
        align=center
    },
    red_label/.style={
        rectangle,
        rounded corners=3pt,
        inner sep=2pt,
        fill = sns_red!10,
        draw = sns_red,
        text=sns_red,
        font=\fontsize{8}{10}\selectfont,
        align=center
    },
    label_text/.style={
        text=sns_dark_grey,
        font=\fontsize{8}{10}\selectfont,
        align=center
    },
    stream_interaction/.style={
        text=sns_green,
        font=\fontsize{8}{10}\selectfont,
        align=center,
        text width=45pt,
        rectangle,
        rounded corners=3pt,
        dashed,
        draw=sns_green,
    },
    dashed_round_rectangle/.style={
        rectangle,
        rounded corners=3pt,
        dashed,
        draw=black,
        align=center
    },
    dirline/.style={
            draw,
            -latex',
        },
    double_dirline/.style={
            draw,
            latex'-latex'
        },
    brace/.style={decorate, decoration={brace, amplitude=5pt}},
}

\vspace{12pt}
\begin{tikzpicture}
    \newcommand{\cuboid}[8]{
        \begin{scope}
            \newcommand\cuboidCol{#1}
            \newcommand\cuboidWeight{#2}
            \newcommand\cuboidLx{#3}
            \newcommand\cuboidBy{#4}
            \newcommand\cuboidWidth{#5}
            \newcommand\cuboidHeight{#6}
            \newcommand\cuboidDepth{#7}
            \newcommand\shear{#8}
            \newcommand\cuboidRx{#3+\cuboidWidth}
            \newcommand\cuboidTy{#4+\cuboidHeight}
            \newcommand\cuboidDt{\cuboidDepth*\shear}

            \fill[\cuboidCol!20]
                (\cuboidLx, \cuboidBy) --
                (\cuboidRx, \cuboidBy) --
                (\cuboidRx, \cuboidTy) --
                (\cuboidLx, \cuboidTy) --
                cycle;
            \fill[\cuboidCol!20]
                (\cuboidLx,      \cuboidTy) --
                (\cuboidRx,      \cuboidTy) --
                (\cuboidRx-\cuboidDt, \cuboidTy+\cuboidDt) --
                (\cuboidLx-\cuboidDt, \cuboidTy+\cuboidDt) --
                cycle;
            \fill[\cuboidCol!40]
                (\cuboidLx,      \cuboidBy) --
                (\cuboidLx,      \cuboidTy) --
                (\cuboidLx-\cuboidDt, \cuboidTy+\cuboidDt) --
                (\cuboidLx-\cuboidDt, \cuboidBy+\cuboidDt) --
                cycle;
            \draw[\cuboidCol, line width=\cuboidWeight]
                (\cuboidLx, \cuboidBy) --
                (\cuboidRx, \cuboidBy) --
                (\cuboidRx, \cuboidTy) --
                (\cuboidLx, \cuboidTy) --
                cycle;
            \draw[\cuboidCol, line width=\cuboidWeight]
                (\cuboidLx,      \cuboidTy) --
                (\cuboidRx,      \cuboidTy) --
                (\cuboidRx-\cuboidDt, \cuboidTy+\cuboidDt) --
                (\cuboidLx-\cuboidDt, \cuboidTy+\cuboidDt) --
                cycle;
            \draw[\cuboidCol, line width=\cuboidWeight]
                (\cuboidLx,      \cuboidBy) --
                (\cuboidLx,      \cuboidTy) --
                (\cuboidLx-\cuboidDt, \cuboidTy+\cuboidDt) --
                (\cuboidLx-\cuboidDt, \cuboidBy+\cuboidDt) --
                cycle;
        \end{scope}
    }

    \newcommand{\wireFrameCuboid}[8]{
        \begin{scope}
            \newcommand\cuboidCol{#1}
            \newcommand\cuboidWeight{#2}
            \newcommand\cuboidLx{#3}
            \newcommand\cuboidBy{#4}
            \newcommand\cuboidWidth{#5}
            \newcommand\cuboidHeight{#6}
            \newcommand\cuboidDepth{#7}
            \newcommand\shear{#8}
            \newcommand\cuboidRx{#3+\cuboidWidth}
            \newcommand\cuboidTy{#4+\cuboidHeight}
            \newcommand\cuboidDt{\cuboidDepth*\shear}

            \draw[\cuboidCol, line width=\cuboidWeight]
                (\cuboidLx, \cuboidBy) --
                (\cuboidRx, \cuboidBy) --
                (\cuboidRx, \cuboidTy) --
                (\cuboidLx, \cuboidTy) --
                cycle;
            \draw[\cuboidCol, line width=\cuboidWeight]
                (\cuboidLx,      \cuboidTy) --
                (\cuboidRx,      \cuboidTy) --
                (\cuboidRx-\cuboidDt, \cuboidTy+\cuboidDt) --
                (\cuboidLx-\cuboidDt, \cuboidTy+\cuboidDt) --
                cycle;
            \draw[\cuboidCol, line width=\cuboidWeight]
                (\cuboidLx,      \cuboidBy) --
                (\cuboidLx,      \cuboidTy) --
                (\cuboidLx-\cuboidDt, \cuboidTy+\cuboidDt) --
                (\cuboidLx-\cuboidDt, \cuboidBy+\cuboidDt) --
                cycle;
        \end{scope}
    }
    \newcommand{\wireFrameCuboidFrontTop}[8]{
        \begin{scope}
            \newcommand\cuboidCol{#1}
            \newcommand\cuboidWeight{#2}
            \newcommand\cuboidLx{#3}
            \newcommand\cuboidBy{#4}
            \newcommand\cuboidWidth{#5}
            \newcommand\cuboidHeight{#6}
            \newcommand\cuboidDepth{#7}
            \newcommand\shear{#8}
            \newcommand\cuboidRx{#3+\cuboidWidth}
            \newcommand\cuboidTy{#4+\cuboidHeight}
            \newcommand\cuboidDt{\cuboidDepth*\shear}

            \draw[\cuboidCol, line width=\cuboidWeight]
                (\cuboidLx, \cuboidBy) --
                (\cuboidRx, \cuboidBy) --
                (\cuboidRx, \cuboidTy) --
                (\cuboidLx, \cuboidTy) --
                cycle;
            \draw[\cuboidCol, line width=\cuboidWeight]
                (\cuboidLx,      \cuboidTy) --
                (\cuboidRx,      \cuboidTy) --
                (\cuboidRx-\cuboidDt, \cuboidTy+\cuboidDt) --
                (\cuboidLx-\cuboidDt, \cuboidTy+\cuboidDt) --
                cycle;
        \end{scope}
    }

    \newcommand{\wireFrameCuboidLeftTop}[8]{
        \begin{scope}
            \newcommand\cuboidCol{#1}
            \newcommand\cuboidWeight{#2}
            \newcommand\cuboidLx{#3}
            \newcommand\cuboidBy{#4}
            \newcommand\cuboidWidth{#5}
            \newcommand\cuboidHeight{#6}
            \newcommand\cuboidDepth{#7}
            \newcommand\shear{#8}
            \newcommand\cuboidRx{#3+\cuboidWidth}
            \newcommand\cuboidTy{#4+\cuboidHeight}
            \newcommand\cuboidDt{\cuboidDepth*\shear}

            \draw[\cuboidCol, line width=\cuboidWeight]
                (\cuboidLx,      \cuboidTy) --
                (\cuboidRx,      \cuboidTy) --
                (\cuboidRx-\cuboidDt, \cuboidTy+\cuboidDt) --
                (\cuboidLx-\cuboidDt, \cuboidTy+\cuboidDt) --
                cycle;
            \draw[\cuboidCol, line width=\cuboidWeight]
                (\cuboidLx,      \cuboidBy) --
                (\cuboidLx,      \cuboidTy) --
                (\cuboidLx-\cuboidDt, \cuboidTy+\cuboidDt) --
                (\cuboidLx-\cuboidDt, \cuboidBy+\cuboidDt) --
                cycle;
        \end{scope}
    }

    \newcommand{\wireFrameCuboidTop}[8]{
        \begin{scope}
            \newcommand\cuboidCol{#1}
            \newcommand\cuboidWeight{#2}
            \newcommand\cuboidLx{#3}
            \newcommand\cuboidBy{#4}
            \newcommand\cuboidWidth{#5}
            \newcommand\cuboidHeight{#6}
            \newcommand\cuboidDepth{#7}
            \newcommand\shear{#8}
            \newcommand\cuboidRx{#3+\cuboidWidth}
            \newcommand\cuboidTy{#4+\cuboidHeight}
            \newcommand\cuboidDt{\cuboidDepth*\shear}

            \draw[\cuboidCol, line width=\cuboidWeight]
                (\cuboidLx,      \cuboidTy) --
                (\cuboidRx,      \cuboidTy) --
                (\cuboidRx-\cuboidDt, \cuboidTy+\cuboidDt) --
                (\cuboidLx-\cuboidDt, \cuboidTy+\cuboidDt) --
                cycle;
        \end{scope}
    }

    \newcommand{\cuboidGradient}[9]{
        \begin{scope}
            \newcommand\cuboidColA{#1}
            \newcommand\cuboidColB{#2}
            \newcommand\cuboidWeight{#3}
            \newcommand\cuboidLx{#4}
            \newcommand\cuboidBy{#5}
            \newcommand\cuboidWidth{#6}
            \newcommand\cuboidHeight{#7}
            \newcommand\cuboidDepth{#8}
            \newcommand\shear{#9}
            \newcommand\cuboidRx{#4+\cuboidWidth}
            \newcommand\cuboidTy{#5+\cuboidHeight}
            \newcommand\cuboidDt{\cuboidDepth*\shear}

            \fill[left color=\cuboidColA!40, right color=\cuboidColB!20]
                (\cuboidLx, \cuboidBy) --
                (\cuboidRx, \cuboidBy) --
                (\cuboidRx, \cuboidTy) --
                (\cuboidLx, \cuboidTy) --
                cycle;
            \fill[left color=\cuboidColA!40, right color=\cuboidColB!20]
                (\cuboidLx,      \cuboidTy) --
                (\cuboidRx,      \cuboidTy) --
                (\cuboidRx-\cuboidDt, \cuboidTy+\cuboidDt) --
                (\cuboidLx-\cuboidDt, \cuboidTy+\cuboidDt) --
                cycle;
            \fill[color=\cuboidColA!60]
                (\cuboidLx,      \cuboidBy) --
                (\cuboidLx,      \cuboidTy) --
                (\cuboidLx-\cuboidDt, \cuboidTy+\cuboidDt) --
                (\cuboidLx-\cuboidDt, \cuboidBy+\cuboidDt) --
                cycle;
            \draw[\cuboidColB, line width=\cuboidWeight]
                (\cuboidLx, \cuboidBy) --
                (\cuboidRx, \cuboidBy) --
                (\cuboidRx, \cuboidTy) --
                (\cuboidLx, \cuboidTy) --
                cycle;
            \draw[\cuboidColB, line width=\cuboidWeight]
                (\cuboidLx,      \cuboidTy) --
                (\cuboidRx,      \cuboidTy) --
                (\cuboidRx-\cuboidDt, \cuboidTy+\cuboidDt) --
                (\cuboidLx-\cuboidDt, \cuboidTy+\cuboidDt) --
                cycle;
            \draw[\cuboidColB, line width=\cuboidWeight]
                (\cuboidLx,      \cuboidBy) --
                (\cuboidLx,      \cuboidTy) --
                (\cuboidLx-\cuboidDt, \cuboidTy+\cuboidDt) --
                (\cuboidLx-\cuboidDt, \cuboidBy+\cuboidDt) --
                cycle;
        \end{scope}
    }

    \newcommand{\cube}[6]{
        \cuboid{#1}{#2}{#3}{#4}{#5}{#5}{#5}{#6};
    }

    \newcommand{\cubeTriple}[3]{
        \begin{scope}
            \newcommand\cubeSize{0.5}
            \newcommand\cubeShear{0.33}
            \newcommand\cubeLineWidth{0.1}
            \newcommand\cubeDistanceScale{0.75}
            \newcommand\cubeTripleColor{#1}
            \foreach \ix in {3,2,1}
                \cube{\cubeTripleColor}{\cubeLineWidth}{\ix * \cubeDistanceScale + #2}{#3}{\cubeSize}{\cubeShear};
        \end{scope}
    }
    \renewcommand{\grid}[3]{
        \begin{scope}
            \newcommand\gridColor{#1}
            \newcommand\gridWeight{0.1}
            \newcommand\gridPosX{#2}
            \newcommand\gridPosY{#3}
            \newcommand\gridCellWidth{0.6}
            \newcommand\gridCellHeight{0.075}
            \newcommand\gridShear{0.33}
            \newcommand\gridDt{\gridCellWidth*\gridShear}

            \foreach \i in {2,1,0} {
                \foreach \j in {2,1,0} {
                    \cuboid
                        {\gridColor}{\gridWeight}
                        {\gridPosX - \j * \gridDt + \i * \gridCellWidth}
                        {\gridPosY + \j * \gridDt}
                        {\gridCellWidth}{\gridCellHeight}{\gridCellWidth}{\gridShear};
                }
            }
        \end{scope}
    }

    \newcommand{\gridGradient}[4]{
        \begin{scope}
            \newcommand\gridColorA{#1}
            \newcommand\gridColorB{#2}
            \newcommand\gridWeight{0.05}
            \newcommand\gridPosX{#3}
            \newcommand\gridPosY{#4}
            \newcommand\gridCellWidth{0.6}
            \newcommand\gridCellHeight{0.075}
            \newcommand\gridShear{0.33}
            \newcommand\gridDt{\gridCellWidth*\gridShear}

            \cuboidGradient
                {\gridColorA}{\gridColorB}{\gridWeight}
                {\gridPosX}
                {\gridPosY}
                {\gridCellWidth * 3}{\gridCellHeight}{\gridCellWidth * 3}
                {\gridShear};

            \wireFrameCuboid
                {\gridColorB}{\gridWeight}
                {\gridPosX}
                {\gridPosY}
                {\gridCellWidth}{\gridCellHeight}{\gridCellWidth}{\gridShear};

            \foreach \i in {2,1} {
                \foreach \j in {0} {
                    \wireFrameCuboidFrontTop
                        {\gridColorB}{\gridWeight}
                        {\gridPosX - \j * \gridDt + \i * \gridCellWidth}
                        {\gridPosY + \j * \gridDt}
                        {\gridCellWidth}{\gridCellHeight}{\gridCellWidth}{\gridShear};
                }
            }

            \foreach \i in {0} {
                \foreach \j in {2,1} {
                    \wireFrameCuboidLeftTop
                        {\gridColorB}{\gridWeight}
                        {\gridPosX - \j * \gridDt + \i * \gridCellWidth}
                        {\gridPosY + \j * \gridDt}
                        {\gridCellWidth}{\gridCellHeight}{\gridCellWidth}{\gridShear};
                }
            }

            \foreach \i in {2,1} {
                \foreach \j in {2,1} {
                    \wireFrameCuboidTop
                        {\gridColorB}{\gridWeight}
                        {\gridPosX - \j * \gridDt + \i * \gridCellWidth}
                        {\gridPosY + \j * \gridDt}
                        {\gridCellWidth}{\gridCellHeight}{\gridCellWidth}{\gridShear};
                }
            }
        \end{scope}
    }
    \newcommand{\gridCellGradient}[4]{
        \begin{scope}
            \newcommand\gridColorA{#1}
            \newcommand\gridColorB{#2}
            \newcommand\gridWeight{0.1}
            \newcommand\gridPosX{#3}
            \newcommand\gridPosY{#4}
            \newcommand\gridCellWidth{0.5}
            \newcommand\gridCellHeight{0.075}
            \newcommand\gridShear{0.2}
            \newcommand\gridDt{\gridCellWidth*\gridShear}

            \foreach \i in {2,1,0} {
                \foreach \j in {2,1,0} {
                    \cuboidGradient
                    {\gridColorA}{\gridColorB}{\gridWeight}
                    {\gridPosX - \j * \gridDt + \i * \gridCellWidth}
                    {\gridPosY + \j * \gridDt}
                    {\gridCellWidth}{\gridCellHeight}{\gridCellWidth}
                    {\gridShear};
                }
            }
        \end{scope}
    }

    \newcommand{\imageBatch}{
        \begin{scope}
            \newcommand\imageSize{40pt};
            \node[]
            (layer_3) at (0.2*\imageSize, 0.2*\imageSize)
            {\includegraphics[width=\imageSize]{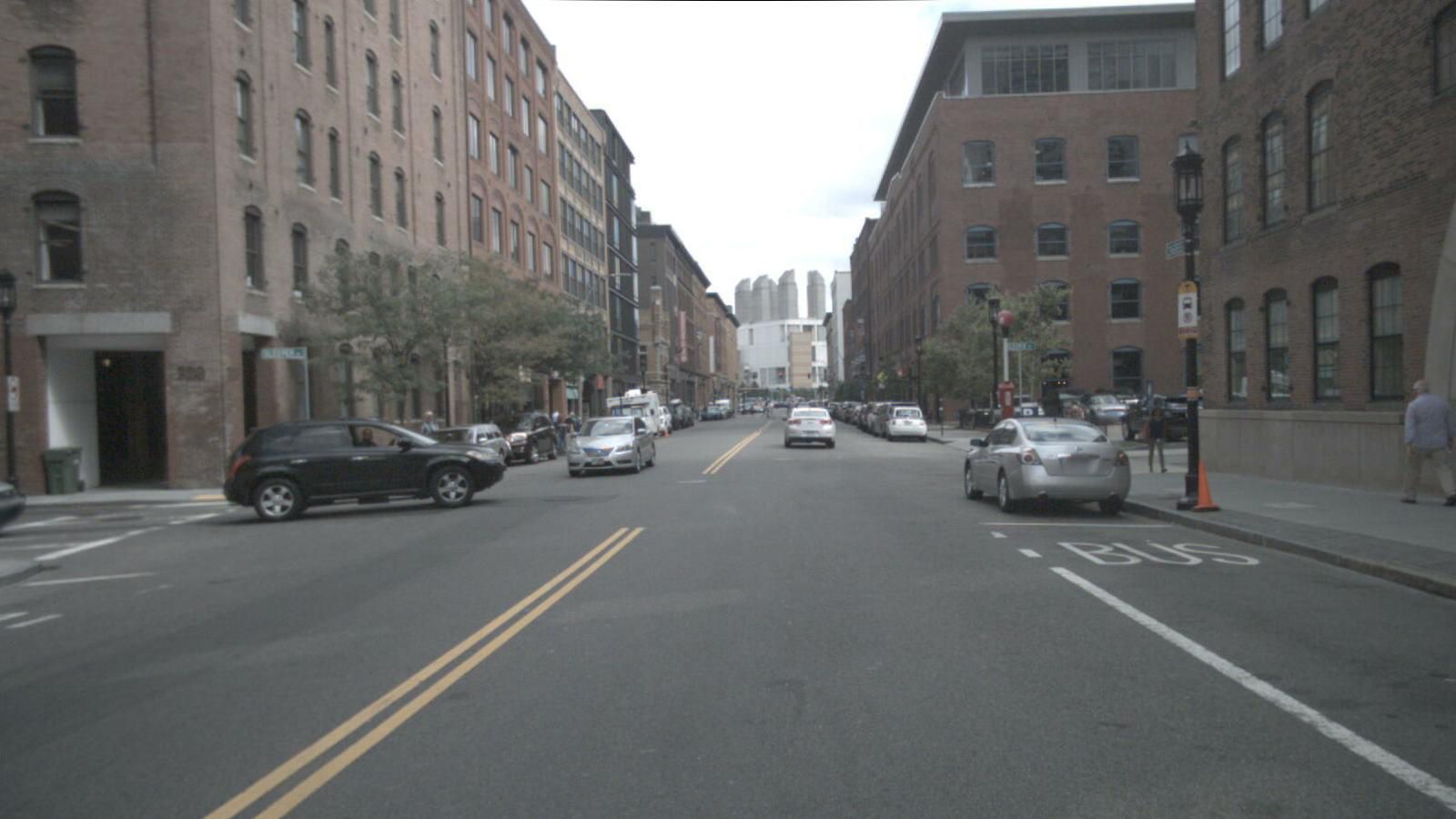}};
            \node[]
            (layer_2) at (0.1*\imageSize, 0.1*\imageSize)
            {\includegraphics[width=\imageSize]{diagrams/imgs/camera_image.jpg}};
            \node[text width=\imageSize]
            (layer_1) at (0,0)
            {\includegraphics[width=\imageSize]{diagrams/imgs/camera_image.jpg}};
        \end{scope}
    }




    \newcommand\dashedLineMargin{10pt}
    \node[](bottom)
    {
        \begin{tikzpicture}
            \newcommand\arrowWidth{80pt}
            \newcommand\arrowHeight{400pt}
            \newcommand\arrowHeadExtend{1pt}
            \newcommand\arrowDistance{120pt}
            \newcommand\arrowLabelDistance{40pt}
            \newcommand\arrowrMarginLeft{40pt}
            \newcommand\arrowrMarginRight{64pt}

            \node[single arrow, draw=sns_blue, fill=sns_blue!20,
                minimum width=\arrowWidth, minimum height=\arrowHeight,
                single arrow head extend = \arrowHeadExtend]
                    (static_stream_arrow) at (0,0) {};
            \node[font=\fontsize{8}{10}\selectfont, color=sns_blue,
                below of=static_stream_arrow, node distance=\arrowLabelDistance]
                    (label_static_stream_arrow) {\textbf{Static Stream}};

            \node[single arrow, draw=sns_orange, fill=sns_orange!20,
                minimum width=\arrowWidth, minimum height=\arrowHeight,
                single arrow head extend = \arrowHeadExtend,
                above of=static_stream_arrow, node distance=\arrowDistance]
                    (dynamic_stream_arrow) {};
            \node[font=\fontsize{8}{10}\selectfont, color=sns_orange,
                    above of=dynamic_stream_arrow, node distance=\arrowLabelDistance]
                    (label_dynamic_stream_arrow) {\textbf{Dynamic Stream}};

            \node[] (static_grid) at ({$(static_stream_arrow.west) + (\arrowrMarginLeft, 0pt)$} |- static_stream_arrow)
            {
                \begin{tikzpicture}
                    \grid{sns_blue}{0}{0};
                \end{tikzpicture}
            };
            \node[dashed_round_rectangle, minimum width=75pt, minimum height=50pt]
            (static_grid_highlight) at (static_grid)
            {};
            \node[label_text, yshift=7.5pt]
            (static_grid_label) at (static_grid_highlight.south)
            {Static Grid};
            \node[xshift=10pt] (static_grid_compensated) at (static_stream_arrow)
            {
                \begin{tikzpicture}
                    \grid{sns_blue}{0}{0};
                \end{tikzpicture}
            };
            \node[dashed_round_rectangle, minimum width=75pt, minimum height=50pt]
            (static_grid_compensated_highlight) at (static_grid_compensated)
            {};
            \node[label_text, yshift=7.5pt]
            (static_grid_compensated_label) at (static_grid_compensated_highlight.south)
            {Static Grid};

            \node[static_world_tasks]
            (static_world_tasks) at ({$(static_stream_arrow.east) - (\arrowrMarginRight, 0pt)$} |- static_stream_arrow)
            {Static World Tasks};

            \node[] (dynamic_agents) at ({$(dynamic_stream_arrow.west) + (\arrowrMarginLeft, 0pt)$} |- dynamic_stream_arrow)
            {
                \begin{tikzpicture}
                    \cubeTriple{sns_orange}{0}{0};
                \end{tikzpicture}
            };
            \node[dashed_round_rectangle, minimum width=75pt, minimum height=50pt]
            (dynamic_agents_highlight) at (dynamic_agents)
            {};
            \node[label_text, yshift=-7.5pt]
            (dynamic_agents_label) at (dynamic_agents_highlight.north)
            {Objects};

            \node[xshift=10pt] (dynamic_agents_compensated) at (dynamic_stream_arrow)
            {
                \begin{tikzpicture}
                    \cubeTriple{sns_orange}{0}{0};
                \end{tikzpicture}
            };
            \node[dashed_round_rectangle, minimum width=75pt, minimum height=50pt]
            (dynamic_agents_compensated_highlight) at (dynamic_agents_compensated)
            {};
            \node[label_text, yshift=-7.5pt]
            (dynamic_agents_compensated_label) at (dynamic_agents_compensated_highlight.north)
            {Objects};
            \node[dynamic_world_tasks]
            (dynamic_agents_tasks) at ({$(dynamic_stream_arrow.east) - (\arrowrMarginRight, 0pt)$} |- dynamic_stream_arrow)
            {Dynamic Agent Tasks};

            \node[xshift=-20pt] (image_batch_1) at ($(static_grid)!0.5!(dynamic_agents)$)
            {
                \begin{tikzpicture}
                    \imageBatch
                \end{tikzpicture}

            };
            \node[xshift=-20pt] (image_batch_2) at ($(static_grid_compensated)!0.5!(dynamic_agents_compensated)$)
            {
                \begin{tikzpicture}
                    \imageBatch
                \end{tikzpicture}

            };
            \node[stream_interaction]
            (stream_interaction_1) at (static_grid.east |- image_batch_1)
            {Stream \\ Interaction};
            \node[stream_interaction    ]
            (stream_interaction_2) at (static_grid_compensated.east |- image_batch_2.east)
            {Stream \\ Interaction};

            \node[label, text width=30pt]
            (ego_motion_only) at ($(static_grid)!0.5!(static_grid_compensated)$)
            {Ego Motion};
            \node[label, text width=45pt]
            (obj_ego_motion) at ($(dynamic_agents)!0.5!(dynamic_agents_compensated)$)
            {Obj + Ego \\ Motion};

            \draw[dirline, sns_blue] ([xshift=5pt] static_grid.east) -- (ego_motion_only) -- ([xshift=-5pt] static_grid_compensated.west);
            \draw[dirline, sns_blue] ([xshift=5pt] static_grid_compensated.east) -- ([xshift=-5pt] static_world_tasks.west);

            \draw[dirline, sns_orange] ([xshift=5pt] static_grid.east |- dynamic_agents.east) -- (obj_ego_motion) -- ([xshift=-5pt] static_grid_compensated.west |- dynamic_agents_compensated.west);
            \draw[dirline, sns_orange] ([xshift=5pt] static_grid_compensated.east |- dynamic_agents_compensated.east) -- ([xshift=-5pt] dynamic_agents_tasks.west);

            \draw[sns_green, thick, dirline]
                ([yshift=-5pt] image_batch_1.north) --([yshift=-5pt] image_batch_1 |- dynamic_agents.south);
            \draw[sns_green, thick, dirline]
                ([yshift=5pt] image_batch_1.south) -- ([yshift=2.5pt] image_batch_1 |- static_grid.north);
            \draw[sns_green, thick, dirline]
                ([yshift=-8pt, xshift=-15pt] stream_interaction_1 |- image_batch_1.north) -- ([yshift=-5pt, xshift=-15pt] stream_interaction_1 |- dynamic_agents.south);
            \draw[sns_green, thick]
                ([yshift=-3pt, xshift=-15pt] stream_interaction_1.south) -- ([yshift=2.5pt, xshift=-15pt] stream_interaction_1 |- static_grid.north);

            \draw[sns_green, thick, dirline]
                ([yshift=-5pt] image_batch_2.north) -- ([yshift=-5pt] image_batch_2 |- dynamic_agents_compensated.south);
            \draw[sns_green, thick, dirline]
                ([yshift=5pt] image_batch_2.south) -- ([yshift=2.5pt] image_batch_2 |- static_grid_compensated.north);
            \draw[sns_green, thick, dirline]
                ([yshift=-8pt, xshift=-15pt] stream_interaction_2 |- image_batch_2.north) -- ([yshift=-5pt, xshift=-15pt] stream_interaction_2 |- dynamic_agents_compensated.south);
            \draw[sns_green, thick]
                ([yshift=-3pt, xshift=-15pt] stream_interaction_2.south) -- ([yshift=2.5pt, xshift=-15pt] stream_interaction_2 |- static_grid_compensated.north);
        \end{tikzpicture}
    };
    \node[above of = bottom, node distance = 225pt](top)
    {
        \newcommand\columnDistance{148pt}
        \newcommand\rowDistance{75pt}
        \begin{tikzpicture}
            \node[]
            (image_batch_1)
            {
                \begin{tikzpicture}
                    \imageBatch
                \end{tikzpicture}

            };
            \node[right of=image_batch_1, node distance=\columnDistance]
            (image_batch_2)
            {
                \begin{tikzpicture}
                    \imageBatch
                \end{tikzpicture}

            };

            \node[above of=image_batch_1, node distance=\rowDistance]
            (static_grid)
            {
                \begin{tikzpicture}
                    \gridGradient{sns_orange}{sns_blue}{0}{0};
                \end{tikzpicture}
            };
            \node[dashed_round_rectangle, minimum width=75pt, minimum height=50pt]
            (static_grid_highlight) at ($(static_grid) + (0, -7.5pt)$)
            {};
            \node[label_text, yshift=15pt]
            (static_grid_label) at (static_grid_highlight.south)
            {Unified Grid};

            \node[above of=image_batch_2, node distance=\rowDistance]
            (static_grid_compensated)
            {
                \begin{tikzpicture}
                    \gridGradient{sns_orange}{sns_blue}{0}{0};
                \end{tikzpicture}
            };
            \node[dashed_round_rectangle, minimum width=75pt, minimum height=50pt]
            (static_grid_compensated_highlight) at ($(static_grid_compensated) + (0, -7.5pt)$)
            {};
            \node[label_text, yshift=15pt]
            (static_grid_compensated_label) at (static_grid_compensated_highlight.south)
            {Unified Grid};

            \node[static_world_tasks, right of=static_grid_compensated, node distance=\columnDistance]
            (static_world_tasks)
            {Static World Tasks};

            \node[above of=static_grid, node distance=\rowDistance] (dynamic_agents)
            {
                \begin{tikzpicture}
                    \cubeTriple{sns_orange}{0}{0};
                \end{tikzpicture}
            };
            \node[dashed_round_rectangle, minimum width=75pt, minimum height=50pt]
            (dynamic_agents_highlight) at (dynamic_agents)
            {};
            \node[label_text, yshift=-7.5pt]
            (dynamic_agents_label) at (dynamic_agents_highlight.north)
            {Objects};

            \node[above of=static_grid_compensated, node distance=\rowDistance] (dynamic_agents_compensated)
            {
                \begin{tikzpicture}
                    \cubeTriple{sns_orange}{0}{0};
                \end{tikzpicture}
            };
            \node[dashed_round_rectangle, minimum width=75pt, minimum height=50pt]
            (dynamic_agents_compensated_highlight) at (dynamic_agents_compensated)
            {};
            \node[label_text, yshift=-7.5pt]
            (dynamic_agents_compensated_label) at (dynamic_agents_compensated_highlight.north)
            {Objects};

            \node[dynamic_world_tasks, right of=dynamic_agents_compensated, node distance=\columnDistance]
            (dynamic_agents_tasks)
            {Dynamic Agent Tasks};

            \draw[dirline, sns_dark_grey] ([yshift=-5pt] image_batch_1.north) -- ([yshift=7.5pt] static_grid_highlight.south);
            \draw[dirline, sns_dark_grey] ([yshift=-5pt] image_batch_2.north) -- ([yshift=7.5pt] static_grid_compensated_highlight.south);

            \draw[dirline, sns_dark_grey] ([yshift=-5pt] static_grid_highlight.north) -- ([yshift=-2.5pt] dynamic_agents.south);
            \draw[dirline, sns_dark_grey] ([yshift=-5pt] static_grid_compensated_highlight.north) -- ([yshift=-2.5pt] dynamic_agents_compensated.south);

            \draw[dirline, sns_dark_grey] ([xshift=7.5pt] dynamic_agents.east) -- ([xshift=-7.5pt] dynamic_agents_compensated.west);

            \draw[dirline, sns_dark_grey] ({$(dynamic_agents.east) + (7.5pt, 0pt)$} |- static_grid.center) -- ({$(dynamic_agents_compensated.west) + (-7.5pt, 0pt)$} |- static_grid_compensated.center);

            \draw[dirline, sns_dark_grey] ([xshift=7.5pt] dynamic_agents_compensated.east) -- ([xshift=-7.5pt] dynamic_agents_tasks.west);
            \draw[dirline, sns_dark_grey] ({$(dynamic_agents_compensated.east) + (7.5pt, 0pt)$} |- static_grid.center) -- ([xshift=-7.5pt] static_world_tasks.west);

            \node[font=\fontsize{42}{48}\selectfont, text=sns_red]
            (temporal_bolt) at ($(static_grid)!0.5!(static_grid_compensated)$)
            {
                \faBolt{}
            };
            \node[red_label, yshift=-40pt, text width=55pt, align=center, text=sns_red]
            (temporal_bolt_label) at ($(static_grid)!0.5!(static_grid_compensated)$)
            {
                No Object \\ Motion \\ compensation.
            };

            \node[font=\fontsize{24}{28}\selectfont, text=sns_red]
            (temporal_bolt) at ($(image_batch_2)!0.5!(static_grid_compensated) + (0pt, -9pt)$)
            {
                \faBolt{}
            };
            \node[red_label, xshift=60pt, text width=90pt, align=center, text=sns_red]
            (temporal_bolt_label) at ($(image_batch_2)!0.5!(static_grid_compensated) + (0pt, -9pt)$)
            {
                No direct object to image attention.
            };
        \end{tikzpicture}
    };
    \node[above of = top, node distance = 132pt](time)
    {
        \begin{tikzpicture}
            \node[]
                (time_i){$t_i$};
            \node[left of=time_i, node distance = 100pt]
                (time_i_1){$t_{i-1}$};
            \node[dynamic_square, right of=time_i, node distance = 50pt]
                (dynamic_square)
                {};
            \node[right of=dynamic_square, node distance = 10pt, anchor=west]
                (dynamic_legend)
                {Dynamic};
            \node[static_square, right of=dynamic_legend, node distance = 50pt]
                (static_square)
                {};
            \node[right of=static_square, node distance = 10pt, anchor=west]
                (static_legend)
                {Static};

            \node[]
                (time label) at ($(time_i)!0.5!(time_i_1) + (0pt, -8pt)$)
                {
                    Time
                };

            \draw[dirline, sns_dark_grey] (time_i_1) -- (time_i);
        \end{tikzpicture}
    };

    \node[left of=top, node distance = 220pt, text width=200pt, align=center, rotate=90]
        (label_top) {\textbf{Unified Grid-Based, Single Stream \\ (BEVFormer, UniAD, VAD)}};

    \node[left of=bottom, node distance = 220pt, text width=100pt, align=center, rotate=90]
        (label_bottom) {\textbf{Dual Stream-based \\ (Ours)}};

    \coordinate (dashed_line_y) at ($(bottom.north)!0.5!(top.south)$);
    \draw[dashed, sns_grey] ([xshift=\dashedLineMargin] bottom.west |- dashed_line_y) -- ([xshift=-\dashedLineMargin] bottom.east |- dashed_line_y);
\end{tikzpicture}
}

%% file: sections/2_related.tex
\section{Related Work}
Accurate and consistent perception forms the basis for autonomous driving. 
We structure related literature into three categories: (i) specialized models for \textit{dynamic agents} that perform 3D object detection and 3D multiple object tracking, (ii) models that reason about \textit{static scene elements} and perform online mapping, and (iii) \textit{multi-task end-to-end} models that jointly perform the aforementioned tasks in a single model and can be optimized end-to-end.



\boldparagraph{Perception of Dynamic Agents}
Based on pioneering works~\cite{carion2020end, wang2022detr3d}, recent specialized models for 3D object detection utilize a transformer-based architecture with a set of object-queries to detect objects in the scene~\cite{li2022bevformer,doll2022spatialdetr, liu2022petr, wang2023object}. 
Several extensions have been proposed \eg to reduce the memory footprint and to increase the convergence speed, resulting in improved overall performance~\cite{zhu2020deformable, li2022dn}.
Incorporating temporal information for the perception of dynamic agents can be achieved by query propagation and therefore implicit tracking~\cite{li2022bevformer, qing2023dort, wang2023exploring} combined with various tracking-by-detection approaches~\cite{yin2021center, wang2021immortal}, or by tracking-by-attention~\cite{doll2023star, zhang2022mutr3d}.
For such query propagation, it is crucial to follow an object-centric paradigm. This allows to aggregate descriptive features for each object by performing attention directly between object-queries and sensor measurements, as well as explicitly compensate for the motion of objects between consecutive time steps~\cite{doll2023star, wang2023exploring}.

Another line of work utilizes an intermediate grid of \ac{bev}-queries to propagate information through time~\cite{li2022bevformer, hu2023planning, jiang2023vad}. 
In such approaches, each \ac{bev}-query always represents the same area in the grid and is not coupled to a specific semantic element.
Dynamic agents are then detected using queries attending to this grid. 
However, since compensating for the motion of dynamic agents in the grid is not directly possible, we opt for an object-centric approach to model dynamic agents in our dual-stream design.

\boldparagraph{Perception of Static Scene Elements}
Inspired by recent works on 2D panoptic segmentation~\cite{li2022panoptic}, current works that perform online map segmentation rely on \ac{bev}-grid-queries, coupled with a transformer-decoder architecture to perform \ac{bev} map segmentation~\cite{li2022bevformer, hu2023planning}.

Another class of approaches tries to model map perception tasks in a vectorized fashion, where map elements are directly modeled as a sequence of points, \eg by leveraging map queries~\cite{liu2023vectormapnet, liao2022maptr, vad_github}.
As both variants rely on a temporal \ac{bev}-grid to achieve a temporally consistent performance, we follow this concept for static world perception.

\boldparagraph{Multi-Task End-to-End Models}
Most recently, different approaches~\cite{hu2023planning, jiang2023vad} proposed to model the driving task as a modular pipeline that is trainable end-to-end. This allows to optimize the individual modules as well as their interfaces towards the final driving task. The modules are typically connected by transformer mechanisms, effectively defining interfaces in terms of query, key and value triplets.

Inspired by the aforementioned works, we propose a dual-stream transformer that can be used as the foundation for various perception tasks as well as for end-to-end multi-task driving.
We simultaneously use object-centric queries to represent dynamic agents in the scene, while modelling static scene elements with \ac{bev}-grid-queries. 
This explicitly disentangles the representation of static and dynamic elements in the scene, resulting in a higher temporal consistency, especially for highly dynamic agents.
The resulting architecture combines the potential of \ac{sota} approaches for dynamic object perception as well as static perception in a single model, and can directly be integrated with recent multi-task models to train the entire stack end-to-end.


%% file: sections/3_method.tex
\section{Method}
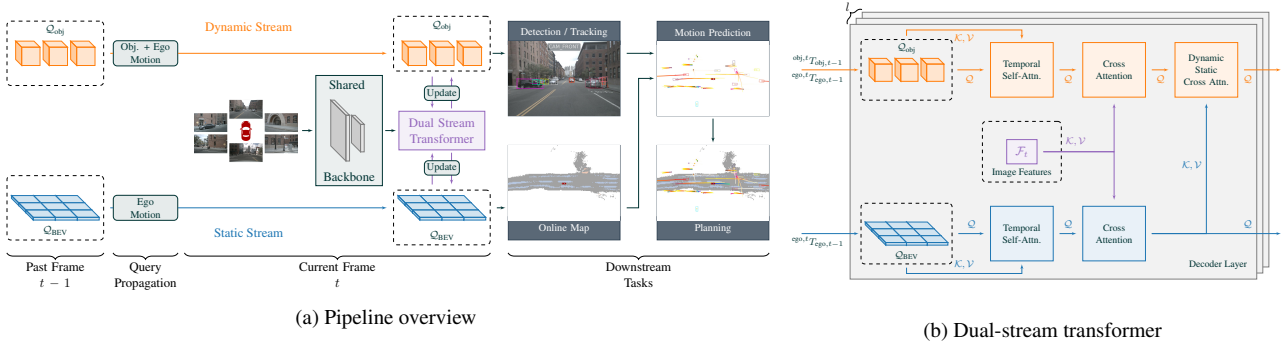
\begin{figure*}
     \centering
     \begin{subfigure}[c]{0.6\textwidth}
         \centering
         \resizebox{\textwidth}{!}{%
         \input{diagrams/architecture.tikz}
         }
         \caption{Pipeline overview}
         \label{fig:architecture}
     \end{subfigure}%
     \hfill
     \begin{subfigure}[c]{0.4\textwidth}
         \centering
         \resizebox{\textwidth}{!}{%
         \input{diagrams/transformer.tikz}
         }
         \begin{minipage}{.1cm}
            \vfill
         \end{minipage}
         \caption{Dual-stream transformer}
         \label{fig:transformer_layer}
     \end{subfigure}%
    \caption{\textbf{\methodName\ Architecture}: Two separate representations are chosen for dynamic agents ($\mathcal{Q}_{\text{obj}}$) and static elements ($\mathcal{Q}_{\text{BEV}}$) as shown in~\figref{fig:architecture}. Self- and cross-attention is simultaneously performed in the proposed dual-stream transformer as shown in~\figref{fig:transformer_layer}, paired with the novel dynamic-static cross-attention block to allow the dynamic agents to benefit from the inferred scene structure.}
    \label{fig:overview}
\end{figure*}

As shown in~\figref{fig:architecture}, our proposed approach \methodName\ comprises a transformer-decoder-based perception architecture that uses two streams to explicitly model dynamic objects in an object-centric and static scene elements in a grid-based fashion.
The resulting dynamic and static world representations enable various tasks relevant to driving such as 3D object detection and tracking, map segmentation, motion prediction as well as planning. Furthermore, our approach permits an end-to-end optimization of the entire driving stack as proposed in~\cite{hu2023planning, jiang2023vad}.

At each time step $t$ a set of $N$ multi-view camera images $\mathcal{I}_t$ is fed into a shared image feature extractor.
The resulting image features $\mathcal{F}_t$ are used by both, the dynamic object as well as the static stream. The former reasons about dynamic agents in the scene, like cars or pedestrians. These agents are represented by a set of object-queries $q_{\text{obj}} \in \mathcal{Q}_{\text{obj}}$ that can be decoded into a bounding box $b_t = \begin{bmatrix} x, y, z, w, l, h, \theta, v_x, v_y \end{bmatrix}$ as well as the predicted class $c$ of the agent.
In parallel, a grid of \ac{bev}-queries $q_{\text{BEV}} \in \mathcal{Q}_{\text{BEV}}$ with dimensions $H_{\text{BEV}} \times W_{\text{BEV}}$ uses $\mathcal{F}_t$ to reason about the static scene. The resulting \ac{bev} representation is used to perform panoptic segmentation of the road topology, \eg drivable space or lane markings, utilizing a segmentation head as proposed in~\cite{li2022panoptic, hu2023planning}.

Interaction between the two streams is enabled by novel dynamic-static cross-attention blocks (see~\figref{fig:transformer_layer}) where the object-queries $q_{\text{obj}}$ attend to the \ac{bev}-queries $q_{\text{BEV}}$ representing the static scene structure. Since the temporal representations for the dynamic and the static world are disentangled, we can explicitly compensate object and ego motion for the object-centric queries of dynamic agents while the static \ac{bev}-queries only require a propagation that is dependent on the motion of the ego vehicle.

\subsection{Dual Stream Design for End-to-End Driving}\label{sec:world_representations}
Finding a well-suited representation for the belief state of the scene is key for any transformer-based end-to-end trainable driving stack as motivated in~\secref{sec:introduction}.
In comparison to traditional pipelines, the end-to-end paradigm allows the interfaces to be optimized towards subsequent modules in the pipeline. Nevertheless, the chosen space of latent representations heavily affects the ability to model the relevant semantic entities and their relations~\cite{wang2023exploring, jiang2023vad}.

Whereas unified \ac{bev}-grid representations can appropriately handle static content, representing highly dynamic objects in a \ac{bev}-grid is ill-posed, since each cell might describe multiple entities with different motion patterns, static scene elements or even a combination of both. We therefore argue that dynamic objects and static scene content should be represented separately, and propose a dual-stream architecture consisting of a dynamic and a static stream.

\boldparagraph{Dynamic Stream}
In \methodName\, dynamic objects are modelled with an object-centric representation by using a single object-query $q_{\text{obj}}$ to describe an individual object in the scene~\cite{wang2022detr3d,liu2022petr,doll2022spatialdetr, doll2023star, wang2023exploring}. 
To obteain a highly descriptive representation, we propose that each object-query should directly perform cross-attention to the image features $\mathcal{F}_t$.
In contrast to unified methods in which only the \ac{bev}-grid queries directly attend to images~\cite{li2022bevformer,hu2023planning,jiang2023vad}, this enables to exploit the high spatial resolution of the image features for more precise detection and tracking.

Following the arguments in~\cite{doll2023star,wang2023exploring}, we propose to propagate the latent queries $q_{\text{obj}}$ to the next timestamp by compensating for the motion between the two timestamps via a latent transformation that depends on the geometric motion.
In contrast to static scene parts, the observed motion of objects consists of two separate components: (1) the motion $\tf{\text{ego}_{t+1}}{\text{ego}_t}$ of the ego vehicle and (2) the motion $\tf{obj_{t+1}}{obj_t}$ of the dynamic object itself.
For more details on query propagation, we kindly refer the reader to~\cite{wang2023exploring,doll2023star}.

In detail, our approach uses the top-k propagated object-queries of each time step as priors in the subsequent frame, following an implicit tracking approach as in StreamPETR~\cite{wang2023exploring} to account for temporary occlusions and to consistently track objects in the scene.
In contrast to tracking-by-attention~\cite{doll2023star,zhang2022mutr3d,hu2023planning} in which only matched objects are propagated to the next frame, this allows our model to maintain multiple hypotheses for the same object and does not require explicit track handling. To obtain explicit object identities, our model can be combined with any tracking-by-detection approach.

\boldparagraph{Static Stream}
We use a \ac{bev} grid-based representation to model static scene elements. A dense, spatially regular representation is well-suited for non-moving objects in the surrounding area. 
Since all elements in the grid are assumed to be static, updates over time are incorporated by applying a rigid transform to the BEV-grid computed from the ego motion $\tf{\text{ego}_{t+1}}{\text{ego}_t}$. We sample grid features differentiably via interpolation and use deformable temporal grid-attention as proposed in~\cite{li2022bevformer}.
Map segmentation is then performed with a decoder-only segmentation head~\cite{li2022panoptic, liao2022maptr}. 
This significantly simplifies the map segmentation head as compared to unified grid-based approaches, since those typically require an additional grid-encoder~\cite{li2022bevformer, hu2023planning}.

\subsection{Modelling Interactions between the Dynamic and Static World}
Explicitly disentangling dynamic agents and static scene elements results in two independent streams of the model. Both streams rely on shared image features $\mathcal{F}_t$, but perform self-attention and cross-attention to these features separately.
To enable the network to leverage mutual information between static scene elements and dynamic agents, we propose an additional attention block that performs dynamic-static cross-attention between the streams. 

As shown in~\figref{fig:transformer_layer}, this is achieved by performing deformable attention~\cite{zhu2020deformable} between object-queries $q_{\text{obj}}$ and \ac{bev}-queries $q_{\text{BEV}}$ of the current timestamp that are close to the position of the object~\cite{li2022bevformer}.
In doing so, the dynamic objects can infer their state update more precisely by considering not only the sensor information but also the aggregated static \ac{bev}-grid, \eg by incorporating the estimated information on road layout and lane topology.

\boldparagraph{Movable Belief State Through Space and Time}
Our dual stream design enables incorporating even unsynchronized sensor input. Whenever sensor information is available, potentially at arbitrary time intervals, the belief state of static and dynamic parts can be propagated to that timestamp considering ego and object motion. The novel sensor data is then easily integrated via cross-attention to the available image features to update the inferred scene state. 
 
Hence, our approach facilitates incorporating sensor measurements at different points in time while simultaneously keeping a temporally consistent representation of the scene. Our model can also handle cases, where the set of sensors varies at each time step. This is especially relevant for non-synchronized sensors, \eg due to different sensing rates, or even sensor failures. 
Additionally, this enables to use ground truth annotations that are synchronized with only a subset of the sensors, as well as getting a model output at timestamps between sensor measurements, which might be beneficial for real-time applications~\cite{wang2023we}.

%% file: diagrams/architecture.tikz
\tikzset{
    static_world_tasks/.style={
        rectangle,
        fill = sns_blue!10,
        draw = sns_blue,
        text=sns_dark_grey,
        font=\fontsize{8}{10}\selectfont,
        text width=60pt,
        align=center
    },
    dynamic_world_tasks/.style={
        rectangle,
        fill = sns_orange!10,
        draw = sns_orange,
        text=sns_dark_grey,
        font=\fontsize{8}{10}\selectfont,
        text width=60pt,
        align=center
    },
    task/.style={
        rectangle,
        fill = task_dark_anthracite,
        draw = task_dark_anthracite,
        text=white,
        fill opacity=0.8,
        text opacity=1.0,
        font=\fontsize{8}{10}\selectfont,
        text width=80pt,
        minimum height=18pt,
        minimum width=80pt,
        align=center
    },
    dual_transformer/.style={
        rectangle,
        fill = sns_purple!10,
        draw = sns_purple,
        text = sns_purple,
        minimum width=50pt,
        minimum height=30pt,
        align=center
    },
    shared_backbone/.style={
        rectangle,
        fill = sns_dark_grey!10,
        draw = sns_dark_grey,
        text = sns_dark_grey,
        minimum width=40pt,
        minimum height=80pt,
        align=center,
        fill opacity=0.8,
        text opacity=1.0
    },
    label/.style={
        rectangle,
        rounded corners=3pt,
        inner sep=2pt,
        fill = sns_dark_grey!10,
        draw = sns_dark_grey,
        text=sns_dark_grey,
        font=\fontsize{8}{10}\selectfont,
        align=center
    },
    label_text/.style={
        text=sns_dark_grey,
        font=\fontsize{8}{10}\selectfont,
        align=center
    },
    stream_interaction/.style={
        text=sns_green,
        font=\fontsize{8}{10}\selectfont,
        align=center,
        text width=45pt,
        rectangle,
        rounded corners=3pt,
        dashed,
        draw=sns_green,
    },
    dashed_round_rectangle/.style={
        rectangle,
        rounded corners=3pt,
        dashed,
        draw=black,
        align=center
    },
    dirline/.style={
            draw,
            -latex',
        },
    double_dirline/.style={
            draw,
            latex'-latex'
        },
    brace/.style={decorate, decoration={brace, amplitude=5pt}},
	bracenode_left/.style={midway, left=2pt, rotate=90, anchor=south},
	bracenode_right/.style={midway, right=2pt, rotate=90, anchor=north},
	bracenode_top/.style={midway, anchor=south, yshift=4pt},
	bracenode_bottom/.style={midway, anchor=north, yshift=-4pt},
}

\vspace{12pt}
\begin{tikzpicture}
    \newcommand{\cuboid}[8]{
        \begin{scope}
            \newcommand\cuboidCol{#1}
            \newcommand\cuboidWeight{#2}
            \newcommand\cuboidLx{#3}
            \newcommand\cuboidBy{#4}
            \newcommand\cuboidWidth{#5}
            \newcommand\cuboidHeight{#6}
            \newcommand\cuboidDepth{#7}
            \newcommand\shear{#8}
            \newcommand\cuboidRx{#3+\cuboidWidth}
            \newcommand\cuboidTy{#4+\cuboidHeight}
            \newcommand\cuboidDt{\cuboidDepth*\shear}

            \fill[\cuboidCol!20]
                (\cuboidLx, \cuboidBy) --
                (\cuboidRx, \cuboidBy) --
                (\cuboidRx, \cuboidTy) --
                (\cuboidLx, \cuboidTy) --
                cycle;
            \fill[\cuboidCol!20]
                (\cuboidLx,      \cuboidTy) --
                (\cuboidRx,      \cuboidTy) --
                (\cuboidRx-\cuboidDt, \cuboidTy+\cuboidDt) --
                (\cuboidLx-\cuboidDt, \cuboidTy+\cuboidDt) --
                cycle;
            \fill[\cuboidCol!40]
                (\cuboidLx,      \cuboidBy) --
                (\cuboidLx,      \cuboidTy) --
                (\cuboidLx-\cuboidDt, \cuboidTy+\cuboidDt) --
                (\cuboidLx-\cuboidDt, \cuboidBy+\cuboidDt) --
                cycle;
            \draw[\cuboidCol, line width=\cuboidWeight]
                (\cuboidLx, \cuboidBy) --
                (\cuboidRx, \cuboidBy) --
                (\cuboidRx, \cuboidTy) --
                (\cuboidLx, \cuboidTy) --
                cycle;
            \draw[\cuboidCol, line width=\cuboidWeight]
                (\cuboidLx,      \cuboidTy) --
                (\cuboidRx,      \cuboidTy) --
                (\cuboidRx-\cuboidDt, \cuboidTy+\cuboidDt) --
                (\cuboidLx-\cuboidDt, \cuboidTy+\cuboidDt) --
                cycle;
            \draw[\cuboidCol, line width=\cuboidWeight]
                (\cuboidLx,      \cuboidBy) --
                (\cuboidLx,      \cuboidTy) --
                (\cuboidLx-\cuboidDt, \cuboidTy+\cuboidDt) --
                (\cuboidLx-\cuboidDt, \cuboidBy+\cuboidDt) --
                cycle;
        \end{scope}
    }

    \newcommand{\wireFrameCuboid}[8]{
        \begin{scope}
            \newcommand\cuboidCol{#1}
            \newcommand\cuboidWeight{#2}
            \newcommand\cuboidLx{#3}
            \newcommand\cuboidBy{#4}
            \newcommand\cuboidWidth{#5}
            \newcommand\cuboidHeight{#6}
            \newcommand\cuboidDepth{#7}
            \newcommand\shear{#8}
            \newcommand\cuboidRx{#3+\cuboidWidth}
            \newcommand\cuboidTy{#4+\cuboidHeight}
            \newcommand\cuboidDt{\cuboidDepth*\shear}

            \draw[\cuboidCol, line width=\cuboidWeight]
                (\cuboidLx, \cuboidBy) --
                (\cuboidRx, \cuboidBy) --
                (\cuboidRx, \cuboidTy) --
                (\cuboidLx, \cuboidTy) --
                cycle;
            \draw[\cuboidCol, line width=\cuboidWeight]
                (\cuboidLx,      \cuboidTy) --
                (\cuboidRx,      \cuboidTy) --
                (\cuboidRx-\cuboidDt, \cuboidTy+\cuboidDt) --
                (\cuboidLx-\cuboidDt, \cuboidTy+\cuboidDt) --
                cycle;
            \draw[\cuboidCol, line width=\cuboidWeight]
                (\cuboidLx,      \cuboidBy) --
                (\cuboidLx,      \cuboidTy) --
                (\cuboidLx-\cuboidDt, \cuboidTy+\cuboidDt) --
                (\cuboidLx-\cuboidDt, \cuboidBy+\cuboidDt) --
                cycle;
        \end{scope}
    }
    \newcommand{\wireFrameCuboidFrontTop}[8]{
        \begin{scope}
            \newcommand\cuboidCol{#1}
            \newcommand\cuboidWeight{#2}
            \newcommand\cuboidLx{#3}
            \newcommand\cuboidBy{#4}
            \newcommand\cuboidWidth{#5}
            \newcommand\cuboidHeight{#6}
            \newcommand\cuboidDepth{#7}
            \newcommand\shear{#8}
            \newcommand\cuboidRx{#3+\cuboidWidth}
            \newcommand\cuboidTy{#4+\cuboidHeight}
            \newcommand\cuboidDt{\cuboidDepth*\shear}

            \draw[\cuboidCol, line width=\cuboidWeight]
                (\cuboidLx, \cuboidBy) --
                (\cuboidRx, \cuboidBy) --
                (\cuboidRx, \cuboidTy) --
                (\cuboidLx, \cuboidTy) --
                cycle;
            \draw[\cuboidCol, line width=\cuboidWeight]
                (\cuboidLx,      \cuboidTy) --
                (\cuboidRx,      \cuboidTy) --
                (\cuboidRx-\cuboidDt, \cuboidTy+\cuboidDt) --
                (\cuboidLx-\cuboidDt, \cuboidTy+\cuboidDt) --
                cycle;
        \end{scope}
    }

    \newcommand{\wireFrameCuboidLeftTop}[8]{
        \begin{scope}
            \newcommand\cuboidCol{#1}
            \newcommand\cuboidWeight{#2}
            \newcommand\cuboidLx{#3}
            \newcommand\cuboidBy{#4}
            \newcommand\cuboidWidth{#5}
            \newcommand\cuboidHeight{#6}
            \newcommand\cuboidDepth{#7}
            \newcommand\shear{#8}
            \newcommand\cuboidRx{#3+\cuboidWidth}
            \newcommand\cuboidTy{#4+\cuboidHeight}
            \newcommand\cuboidDt{\cuboidDepth*\shear}

            \draw[\cuboidCol, line width=\cuboidWeight]
                (\cuboidLx,      \cuboidTy) --
                (\cuboidRx,      \cuboidTy) --
                (\cuboidRx-\cuboidDt, \cuboidTy+\cuboidDt) --
                (\cuboidLx-\cuboidDt, \cuboidTy+\cuboidDt) --
                cycle;
            \draw[\cuboidCol, line width=\cuboidWeight]
                (\cuboidLx,      \cuboidBy) --
                (\cuboidLx,      \cuboidTy) --
                (\cuboidLx-\cuboidDt, \cuboidTy+\cuboidDt) --
                (\cuboidLx-\cuboidDt, \cuboidBy+\cuboidDt) --
                cycle;
        \end{scope}
    }

    \newcommand{\wireFrameCuboidTop}[8]{
        \begin{scope}
            \newcommand\cuboidCol{#1}
            \newcommand\cuboidWeight{#2}
            \newcommand\cuboidLx{#3}
            \newcommand\cuboidBy{#4}
            \newcommand\cuboidWidth{#5}
            \newcommand\cuboidHeight{#6}
            \newcommand\cuboidDepth{#7}
            \newcommand\shear{#8}
            \newcommand\cuboidRx{#3+\cuboidWidth}
            \newcommand\cuboidTy{#4+\cuboidHeight}
            \newcommand\cuboidDt{\cuboidDepth*\shear}

            \draw[\cuboidCol, line width=\cuboidWeight]
                (\cuboidLx,      \cuboidTy) --
                (\cuboidRx,      \cuboidTy) --
                (\cuboidRx-\cuboidDt, \cuboidTy+\cuboidDt) --
                (\cuboidLx-\cuboidDt, \cuboidTy+\cuboidDt) --
                cycle;
        \end{scope}
    }

    \newcommand{\cuboidGradient}[9]{
        \begin{scope}
            \newcommand\cuboidColA{#1}
            \newcommand\cuboidColB{#2}
            \newcommand\cuboidWeight{#3}
            \newcommand\cuboidLx{#4}
            \newcommand\cuboidBy{#5}
            \newcommand\cuboidWidth{#6}
            \newcommand\cuboidHeight{#7}
            \newcommand\cuboidDepth{#8}
            \newcommand\shear{#9}
            \newcommand\cuboidRx{#4+\cuboidWidth}
            \newcommand\cuboidTy{#5+\cuboidHeight}
            \newcommand\cuboidDt{\cuboidDepth*\shear}

            \fill[left color=\cuboidColA!40, right color=\cuboidColB!20]
                (\cuboidLx, \cuboidBy) --
                (\cuboidRx, \cuboidBy) --
                (\cuboidRx, \cuboidTy) --
                (\cuboidLx, \cuboidTy) --
                cycle;
            \fill[left color=\cuboidColA!40, right color=\cuboidColB!20]
                (\cuboidLx,      \cuboidTy) --
                (\cuboidRx,      \cuboidTy) --
                (\cuboidRx-\cuboidDt, \cuboidTy+\cuboidDt) --
                (\cuboidLx-\cuboidDt, \cuboidTy+\cuboidDt) --
                cycle;
            \fill[color=\cuboidColA!60]
                (\cuboidLx,      \cuboidBy) --
                (\cuboidLx,      \cuboidTy) --
                (\cuboidLx-\cuboidDt, \cuboidTy+\cuboidDt) --
                (\cuboidLx-\cuboidDt, \cuboidBy+\cuboidDt) --
                cycle;
            \draw[\cuboidColB, line width=\cuboidWeight]
                (\cuboidLx, \cuboidBy) --
                (\cuboidRx, \cuboidBy) --
                (\cuboidRx, \cuboidTy) --
                (\cuboidLx, \cuboidTy) --
                cycle;
            \draw[\cuboidColB, line width=\cuboidWeight]
                (\cuboidLx,      \cuboidTy) --
                (\cuboidRx,      \cuboidTy) --
                (\cuboidRx-\cuboidDt, \cuboidTy+\cuboidDt) --
                (\cuboidLx-\cuboidDt, \cuboidTy+\cuboidDt) --
                cycle;
            \draw[\cuboidColB, line width=\cuboidWeight]
                (\cuboidLx,      \cuboidBy) --
                (\cuboidLx,      \cuboidTy) --
                (\cuboidLx-\cuboidDt, \cuboidTy+\cuboidDt) --
                (\cuboidLx-\cuboidDt, \cuboidBy+\cuboidDt) --
                cycle;
        \end{scope}
    }

    \newcommand{\cube}[6]{
        \cuboid{#1}{#2}{#3}{#4}{#5}{#5}{#5}{#6};
    }

    \newcommand{\cubeTriple}[3]{
        \begin{scope}
            \newcommand\cubeSize{0.5}
            \newcommand\cubeShear{0.33}
            \newcommand\cubeLineWidth{0.1}
            \newcommand\cubeDistanceScale{0.75}
            \newcommand\cubeTripleColor{#1}
            \foreach \ix in {3,2,1}
                \cube{\cubeTripleColor}{\cubeLineWidth}{\ix * \cubeDistanceScale + #2}{#3}{\cubeSize}{\cubeShear};
        \end{scope}
    }

    \newcommand{\backbone}[3]{
        \begin{scope}
            \newcommand\backboneColor{#1}
            \newcommand\backboneWeight{0.1}
            \newcommand\backboneShear{0.33}
            \newcommand\backoneX{#2}
            \newcommand\backboneY{#3}
            \newcommand\inputLayerSize{1.25}
            \newcommand\secondLayerSize{0.75}
            \newcommand\layerThickness{0.1}
            \newcommand\backboneShift{0.2}
            \tikzmath{
                \inputLayerOffset = 0.5 * \inputLayerSize;
                \secondLayerOffset = 0.5 * \secondLayerSize;
            }

            \cuboid
                {\backboneColor}{\backboneWeight}
                {\backoneX + \backboneShift}{\backboneY - \secondLayerOffset}
                {\layerThickness}{\secondLayerSize}{\secondLayerSize}{\backboneShear};

            \cuboid
                {\backboneColor}{\backboneWeight}
                {\backoneX - \backboneShift}{\backboneY - \inputLayerOffset}
                {\layerThickness}{\inputLayerSize}{\inputLayerSize}{\backboneShear};

            \end{scope}
    }

    \renewcommand{\grid}[3]{
        \begin{scope}
            \newcommand\gridColor{#1}
            \newcommand\gridWeight{0.1}
            \newcommand\gridPosX{#2}
            \newcommand\gridPosY{#3}
            \newcommand\gridCellWidth{0.6}
            \newcommand\gridCellHeight{0.075}
            \newcommand\gridShear{0.33}
            \newcommand\gridDt{\gridCellWidth*\gridShear}

            \foreach \i in {2,1,0} {
                \foreach \j in {2,1,0} {
                    \cuboid
                        {\gridColor}{\gridWeight}
                        {\gridPosX - \j * \gridDt + \i * \gridCellWidth}
                        {\gridPosY + \j * \gridDt}
                        {\gridCellWidth}{\gridCellHeight}{\gridCellWidth}{\gridShear};
                }
            }
        \end{scope}
    }

    \newcommand{\gridGradient}[4]{
        \begin{scope}
            \newcommand\gridColorA{#1}
            \newcommand\gridColorB{#2}
            \newcommand\gridWeight{0.05}
            \newcommand\gridPosX{#3}
            \newcommand\gridPosY{#4}
            \newcommand\gridCellWidth{0.6}
            \newcommand\gridCellHeight{0.075}
            \newcommand\gridShear{0.33}
            \newcommand\gridDt{\gridCellWidth*\gridShear}

            \cuboidGradient
                {\gridColorA}{\gridColorB}{\gridWeight}
                {\gridPosX}
                {\gridPosY}
                {\gridCellWidth * 3}{\gridCellHeight}{\gridCellWidth * 3}
                {\gridShear};

            \wireFrameCuboid
                {\gridColorB}{\gridWeight}
                {\gridPosX}
                {\gridPosY}
                {\gridCellWidth}{\gridCellHeight}{\gridCellWidth}{\gridShear};

            \foreach \i in {2,1} {
                \foreach \j in {0} {
                    \wireFrameCuboidFrontTop
                        {\gridColorB}{\gridWeight}
                        {\gridPosX - \j * \gridDt + \i * \gridCellWidth}
                        {\gridPosY + \j * \gridDt}
                        {\gridCellWidth}{\gridCellHeight}{\gridCellWidth}{\gridShear};
                }
            }

            \foreach \i in {0} {
                \foreach \j in {2,1} {
                    \wireFrameCuboidLeftTop
                        {\gridColorB}{\gridWeight}
                        {\gridPosX - \j * \gridDt + \i * \gridCellWidth}
                        {\gridPosY + \j * \gridDt}
                        {\gridCellWidth}{\gridCellHeight}{\gridCellWidth}{\gridShear};
                }
            }

            \foreach \i in {2,1} {
                \foreach \j in {2,1} {
                    \wireFrameCuboidTop
                        {\gridColorB}{\gridWeight}
                        {\gridPosX - \j * \gridDt + \i * \gridCellWidth}
                        {\gridPosY + \j * \gridDt}
                        {\gridCellWidth}{\gridCellHeight}{\gridCellWidth}{\gridShear};
                }
            }
        \end{scope}
    }

    \newcommand{\gridCellGradient}[4]{
        \begin{scope}
            \newcommand\gridColorA{#1}
            \newcommand\gridColorB{#2}
            \newcommand\gridWeight{0.1}
            \newcommand\gridPosX{#3}
            \newcommand\gridPosY{#4}
            \newcommand\gridCellWidth{0.5}
            \newcommand\gridCellHeight{0.075}
            \newcommand\gridShear{0.2}
            \newcommand\gridDt{\gridCellWidth*\gridShear}

            \foreach \i in {2,1,0} {
                \foreach \j in {2,1,0} {
                    \cuboidGradient
                    {\gridColorA}{\gridColorB}{\gridWeight}
                    {\gridPosX - \j * \gridDt + \i * \gridCellWidth}
                    {\gridPosY + \j * \gridDt}
                    {\gridCellWidth}{\gridCellHeight}{\gridCellWidth}
                    {\gridShear};
                }
            }
        \end{scope}
    }

    \newcommand{\imageBatch}{
        \begin{scope}
            \newcommand\imageSize{40pt};
            \node[]
            (layer_3) at (0.2*\imageSize, 0.2*\imageSize)
            {\includegraphics[width=\imageSize]{camera_image.jpg}};
            \node[]
            (layer_2) at (0.1*\imageSize, 0.1*\imageSize)
            {\includegraphics[width=\imageSize]{camera_image.jpg}};
            \node[text width=\imageSize]
            (layer_1) at (0,0)
            {\includegraphics[width=\imageSize]{camera_image.jpg}};
        \end{scope}
    }




    \newcommand\dashedLineMargin{10pt}
    \node[](bottom)
    {
        \begin{tikzpicture}
            \newcommand\compensationDistance{70pt}
            \newcommand\compensationWidth{50pt}
            \newcommand\compastedStateDistance{230pt}
            \newcommand\streamDistance{120pt}
            \newcommand\temporalDistance{35pt}
            \newcommand\sharedBackboneDistance{80pt}
            \newcommand\cameraSetupImageSize{80pt}
            \newcommand\taskDistance{100pt}
            \newcommand\taskImageSize{86.65pt}

            \node[] (static_grid)
            {
                \begin{tikzpicture}
                    \grid{sns_blue}{0}{0};
                \end{tikzpicture}
            };
            \node[dashed_round_rectangle, minimum width=75pt, minimum height=50pt]
            (static_grid_highlight) at (static_grid)
            {};
            \node[label_text, yshift=7.5pt]
            (static_grid_label) at (static_grid_highlight.south)
            {$\mathcal{Q}_{\text{BEV}}$};

            \node[label, text width=30pt, right of=static_grid, node distance=\compensationDistance,  minimum width=\compensationWidth]
            (ego_motion_only)
            {Ego Motion};

            \node[right of=ego_motion_only, node distance=\compastedStateDistance]
            (static_grid_compensated)
            {
                \begin{tikzpicture}
                    \grid{sns_blue}{0}{0};
                \end{tikzpicture}
            };
            \node[dashed_round_rectangle, minimum width=75pt, minimum height=42.5pt]
            (static_grid_compensated_highlight) at ($(static_grid_compensated) + (0pt, -7.5pt)$)
            {};
            \node[label_text, yshift=7.5pt]
            (static_grid_compensated_label) at (static_grid_compensated_highlight.south)
            {$\mathcal{Q}_{\text{BEV}}$};

            \node[text=sns_blue]
            (static_stream_label) at ($(static_grid_label)!0.5!(static_grid_compensated_label)$)
            {Static Stream};


            \node[above of=static_grid, node distance=\streamDistance]
            (dynamic_agents)
            {
                \begin{tikzpicture}
                    \cubeTriple{sns_orange}{0}{0};
                \end{tikzpicture}
            };
            \node[dashed_round_rectangle, minimum width=75pt, minimum height=50pt]
            (dynamic_agents_highlight) at (dynamic_agents)
            {};
            \node[label_text, yshift=-7.5pt]
            (dynamic_agents_label) at (dynamic_agents_highlight.north)
            {$\mathcal{Q}_{\text{obj}}$};

            \node[label, text width=45pt, right of=dynamic_agents, node distance=\compensationDistance, minimum width=\compensationWidth]
            (obj_ego_motion)
            {Obj. + Ego \\ Motion};

            \node[right of=obj_ego_motion, node distance=\compastedStateDistance]
            (dynamic_agents_compensated)
            {
                \begin{tikzpicture}
                    \cubeTriple{sns_orange}{0}{0};
                \end{tikzpicture}
            };
            \node[dashed_round_rectangle, minimum width=75pt, minimum height=42.5pt]
            (dynamic_agents_compensated_highlight) at ($(dynamic_agents_compensated) + (0pt, 7.5pt)$)
            {};
            \node[label_text, yshift=-7.5pt]
            (dynamic_agents_compensated_label) at (dynamic_agents_compensated_highlight.north)
            {$\mathcal{Q}_{\text{obj}}$};

            \node[text=sns_orange]
            (dynamic_stream_label) at ($(dynamic_agents_label)!0.5!(dynamic_agents_compensated_label)$)
            {Dynamic Stream};

            \node[dual_transformer, text width=60pt, align=center]
            (dual_transformer) at ($(static_grid_compensated)!0.5!(dynamic_agents_compensated)$)
            {Dual Stream \\ Transformer};

            \node[shared_backbone, text width=45pt, left of=dual_transformer, node distance=0.9*\sharedBackboneDistance]
            (shared_backbone)
            {Shared \vspace*{60pt} \\ Backbone};

            \node[align=center]
            (shared_backbone_layers) at (shared_backbone)
            {
                \begin{tikzpicture}
                    \backbone{sns_grey}{0}{0};
                \end{tikzpicture}
            };

            \node[left of=shared_backbone, node distance=\sharedBackboneDistance]
            (camera_setup)
            {
                \includegraphics[width=\cameraSetupImageSize]{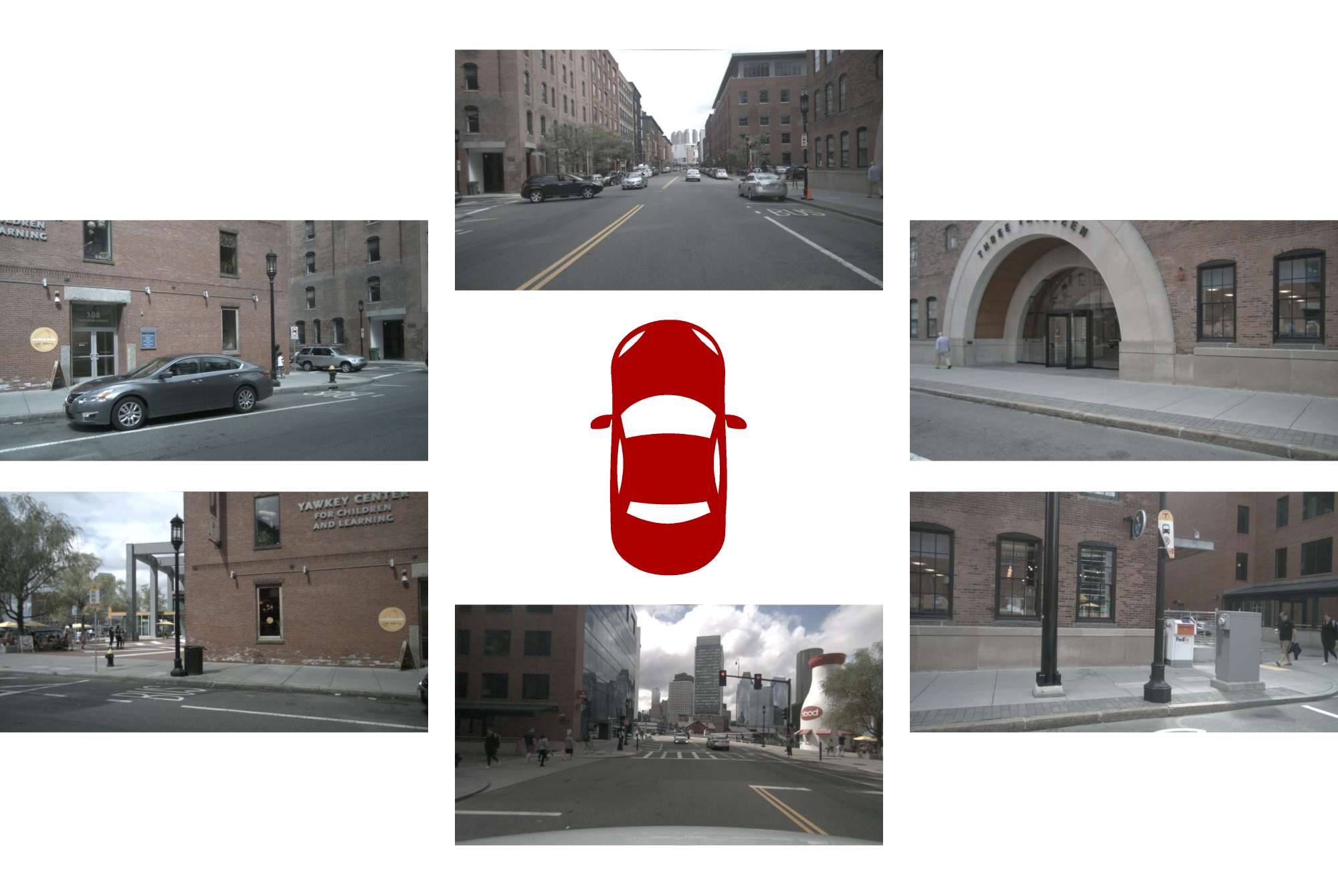}
            };

            \node[task, anchor=north west, yshift=12pt]
            (detection_tracking) at ($(dynamic_agents_compensated.north east) + (17pt, 0pt)$)
            {Detection / Tracking};
            \node[yshift=-0.5*\taskImageSize+8pt, draw=task_dark_anthracite, inner sep=0pt]
            (detection_tracking_img) at (detection_tracking)
            {
                \includegraphics[width=\taskImageSize]{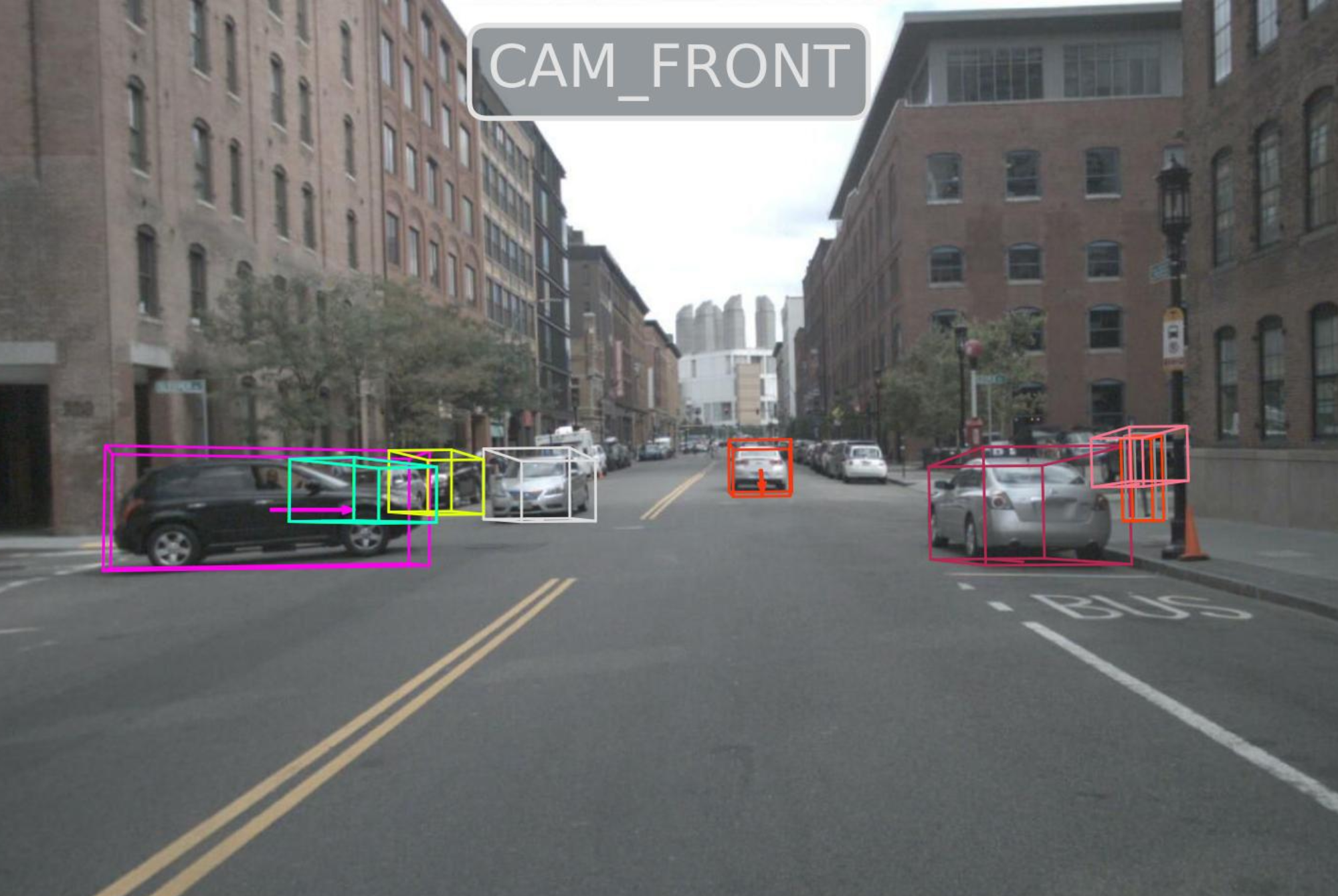}
            };

            \node[task, yshift=-12pt, anchor=south]
            (online_map) at (detection_tracking |- static_grid_compensated.south)
            {Online Map};
            \node[yshift= 0.5*\taskImageSize-8pt, draw=task_dark_anthracite, inner sep=0pt]
            (online_map_img) at (online_map)
            {
                \includegraphics[width=\taskImageSize]{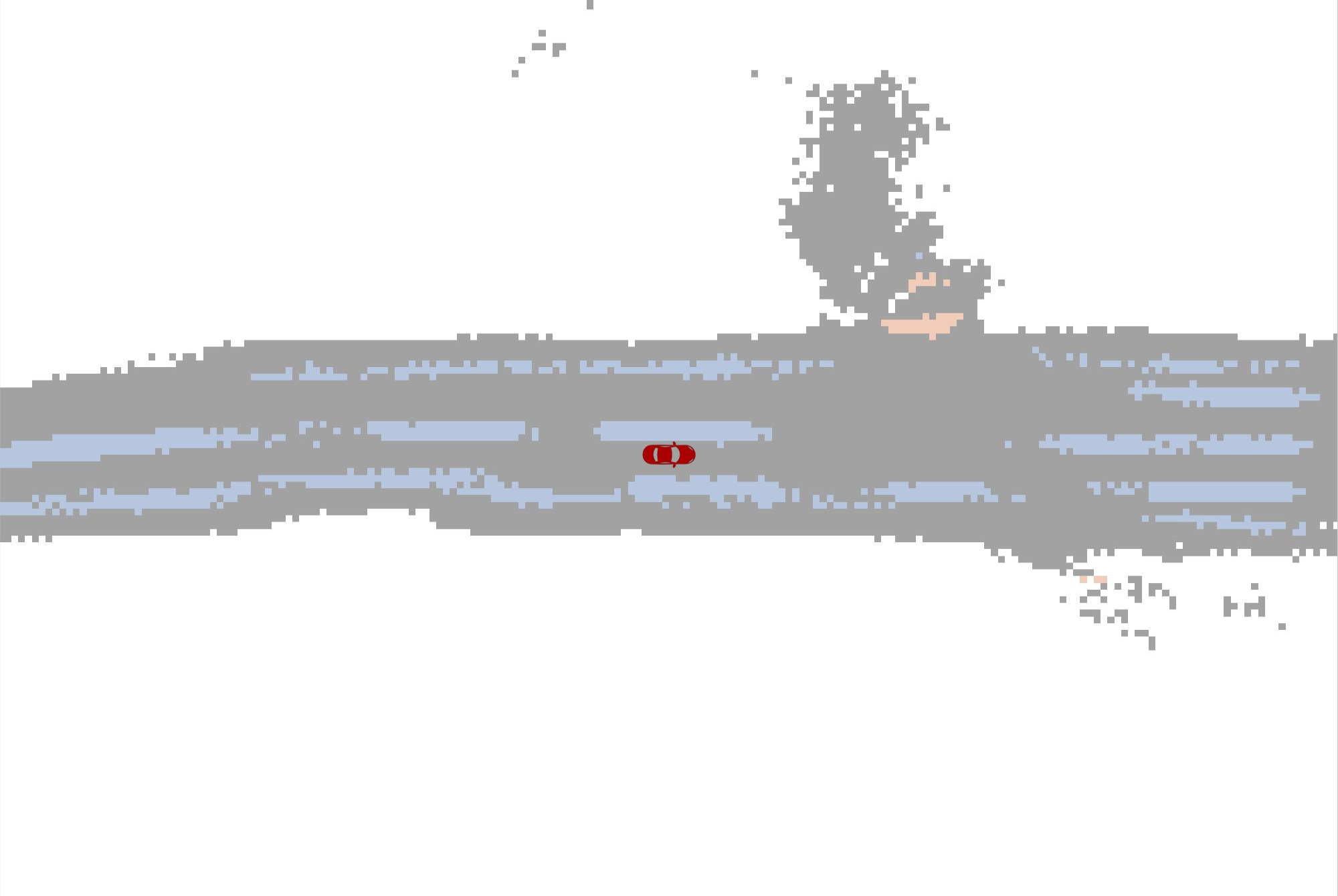}
            };

            \node[task, right=of detection_tracking]
            (motion_prediction) 
            {Motion Prediction};
            \node[yshift=-0.5*\taskImageSize+8pt, draw=task_dark_anthracite, inner sep=0pt]
            (motion_prediction_img) at (motion_prediction)
            {
                \includegraphics[width=\taskImageSize]{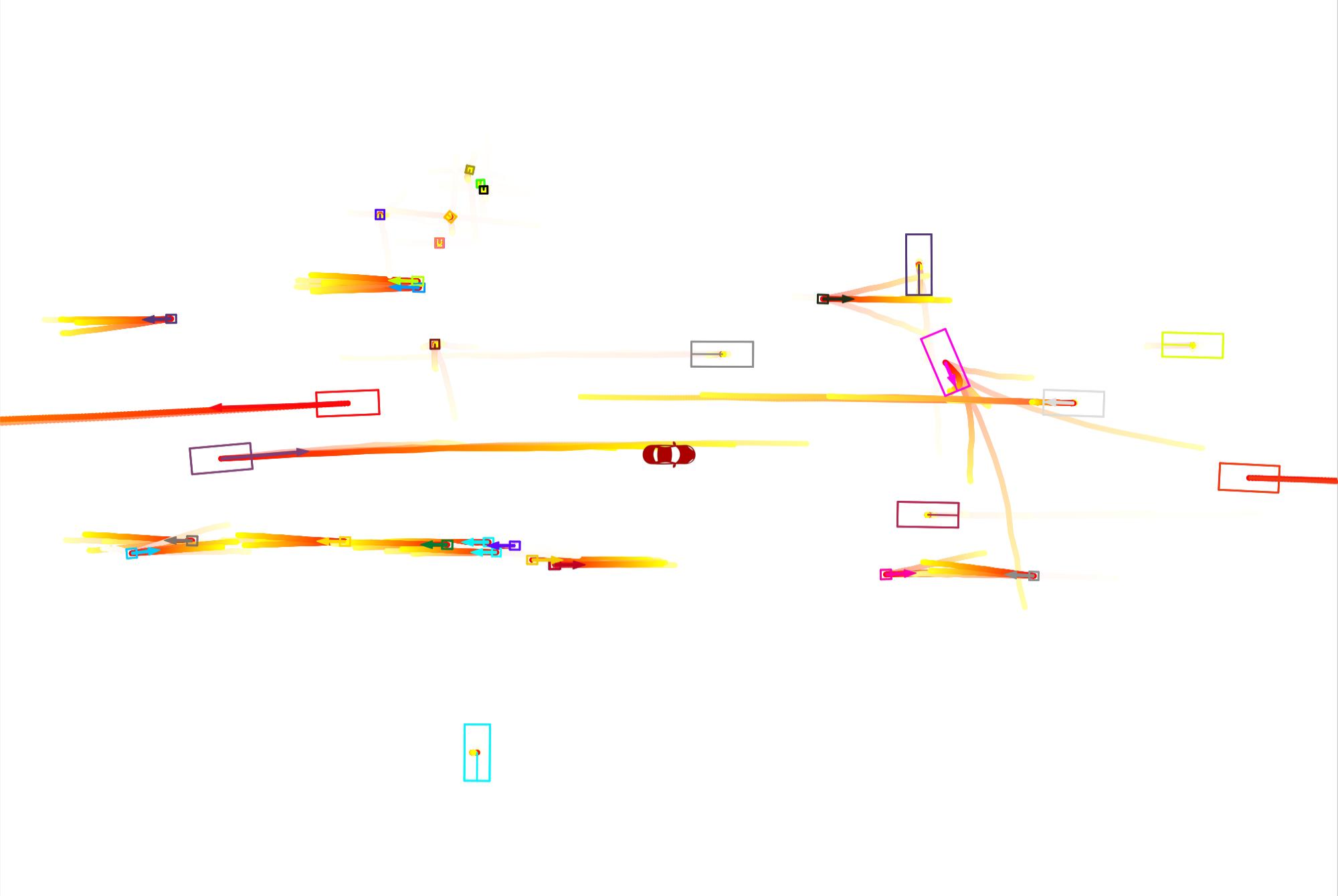}
            };

            \node[task, right=of online_map]
            (planning) 
            {Planning};
            \node[yshift= 0.5*\taskImageSize-8pt, draw=task_dark_anthracite, inner sep=0pt]
            (planning_img) at (planning)
            {
                \includegraphics[width=\taskImageSize]{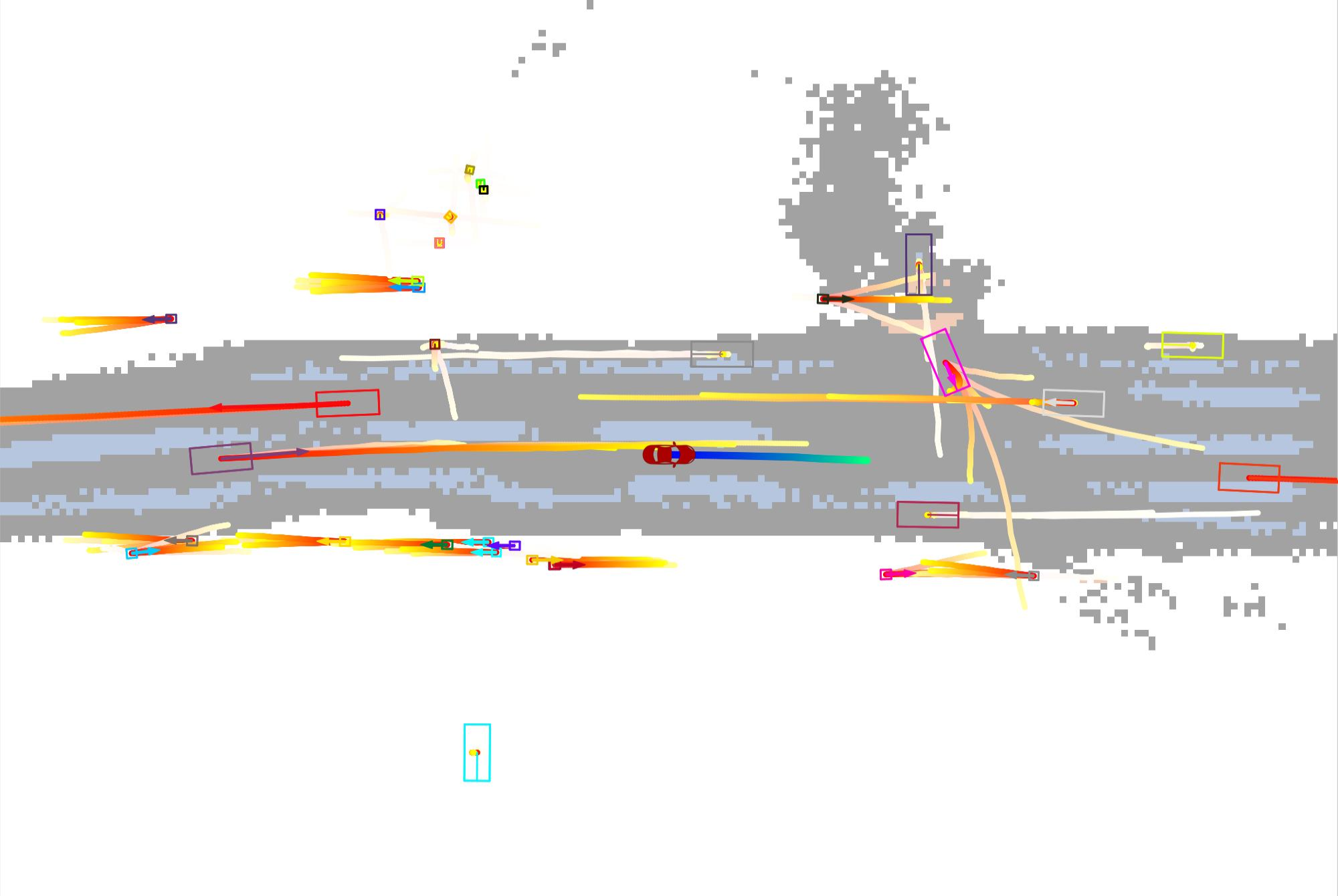}
            };

            \node[below of=static_grid, node distance=\temporalDistance] (temporal_row) {};
            \draw[brace] (static_grid.east |- temporal_row) -- (static_grid.west |- temporal_row)
                node [bracenode_bottom, text width=80pt, align = center] { Past Frame \\ $t - 1$};

            \draw[brace] (ego_motion_only.east |- temporal_row) -- (ego_motion_only.west |- temporal_row)
                node [bracenode_bottom, text width=80pt, align = center] { Query \\ Propagation };

            \draw[brace] (static_grid_compensated.east |- temporal_row) -- ({$(ego_motion_only.east) + (5pt, 0pt)$} |- temporal_row)
                node [bracenode_bottom, text width=80pt, align = center] { Current Frame \\ $t$ };

            \draw[brace] (planning.east |- temporal_row) -- (online_map.west |- temporal_row)
                node [bracenode_bottom, text width=80pt, align = center] { Downstream \\ Tasks };

            \draw[dirline, sns_blue] ([xshift=5pt] static_grid.east) -- (ego_motion_only) -- ([xshift=-5pt] static_grid_compensated.west);
            \draw[dirline, sns_orange] ([xshift=5pt] static_grid.east |- dynamic_agents.east) -- (obj_ego_motion) --
                ([xshift=-5pt] static_grid_compensated.west |- dynamic_agents_compensated.west);

            \draw[dirline, sns_dark_grey] (camera_setup) -- ([xshift=-2pt] shared_backbone.west);
            \draw[dirline, sns_dark_grey] ([xshift=2pt] shared_backbone.east) -- ([xshift=-2pt] dual_transformer.west);

            \draw[dirline, sns_dark_grey] ([xshift=5pt] dynamic_agents_compensated.east) --([xshift=-2pt] detection_tracking.west |- dynamic_agents_compensated);
            \draw[dirline, sns_dark_grey] ([xshift=5pt] detection_tracking.east |- dynamic_agents_compensated) -- ([xshift=-2pt] motion_prediction.west |- dynamic_agents_compensated);

            \draw[dirline, sns_dark_grey] ([xshift=5pt] online_map.east |- static_grid_compensated) -| ($(online_map)!0.5!(motion_prediction)$) |- ([xshift=-2pt] motion_prediction.west |- motion_prediction_img);
            \draw[dirline, sns_dark_grey] ([xshift=5pt] static_grid_compensated.east) -- ([xshift=-2pt] online_map.west |- static_grid_compensated);

            \draw[dirline, sns_dark_grey] ([yshift=-2pt] motion_prediction_img.south) -- ([yshift=2pt] planning_img.north);

            \draw[dirline, sns_purple] ($(dynamic_agents_compensated.south) + (-7.5pt, -5pt)$)
                -- ({$(dynamic_agents_compensated.south) +  (-7.5pt, -5pt)$} |- {$(dual_transformer.north) + (0pt, 2pt)$});
            \draw[dirline, sns_purple] ({$(dynamic_agents_compensated.south) + (7.5pt, -5pt)$} |- {$(dual_transformer.north) + (0pt, 2pt)$})
                -- ($(dynamic_agents_compensated.south) + (7.5pt, -5pt)$);
            \node[label]
                (update_agents_label) at ($(dual_transformer)!0.5!(dynamic_agents_compensated) + (0pt,-1pt)$)
                {Update};

            \draw[dirline, sns_purple] ($(static_grid_compensated.north) + (-7.5pt, 5pt)$)
                -- ({$(static_grid_compensated.north) +  (-7.5pt, 5pt)$} |- {$(dual_transformer.south) + (0pt, -2pt)$});
            \draw[dirline, sns_purple] ({$(static_grid_compensated.north) + (7.5pt, -2pt)$} |- {$(dual_transformer.south) + (0pt, -2pt)$})
                -- ($(static_grid_compensated.north) + (7.5pt, 5pt)$);
            \node[label]
                (update_grid_label) at ($(dual_transformer)!0.5!(static_grid_compensated) + (0pt, 1pt)$)
                {Update};


        \end{tikzpicture}
    };

\end{tikzpicture}

%% file: diagrams/transformer.tikz
\tikzset{
    label/.style={
        rectangle,
        rounded corners=3pt,
        inner sep=2pt,
        fill = sns_dark_grey!10,
        draw = sns_dark_grey,
        text=sns_dark_grey,
        font=\fontsize{8}{10}\selectfont,
        align=center
    },
    label_text/.style={
        text=sns_dark_grey,
        font=\fontsize{8}{10}\selectfont,
        align=center
    },
    image_features/.style={
        rectangle,
        fill = sns_purple!10,
        draw = sns_purple,
        text = sns_purple,
        minimum width=25pt,
        minimum height=20pt,
        align=center
    },
    static_world_function/.style={
        rectangle,
        fill = sns_blue!10,
        draw = sns_blue,
        text=sns_dark_grey,
        font=\fontsize{8}{10}\selectfont,
        text width=45pt,
        minimum width=45pt,
        minimum height=45pt,
        align=center
    },
    dynamic_world_function/.style={
        rectangle,
        fill = sns_orange!10,
        draw = sns_orange,
        text=sns_dark_grey,
        font=\fontsize{8}{10}\selectfont,
        text width=45pt,
        minimum width=45pt,
        minimum height=45pt,
        align=center
    },
    dashed_round_rectangle/.style={
        rectangle,
        rounded corners=3pt,
        dashed,
        draw=black,
        align=center
    },
    decoder_layer/.style={
        rectangle,
        fill = sns_grey!10,
        draw = sns_grey,
        text=sns_grey,
        font=\fontsize{8}{10}\selectfont,
        text width=45pt,
        minimum width=335pt,
        minimum height=210pt,
        align=center
    },
    dirline/.style={
            draw,
            -latex',
        },
    double_dirline/.style={
            draw,
            latex'-latex'
        },
    brace/.style={decorate, decoration={brace, amplitude=5pt}},
	bracenode_left/.style={midway, left=2pt, rotate=90, anchor=south},
	bracenode_right/.style={midway, right=2pt, rotate=90, anchor=north},
	bracenode_top/.style={midway, anchor=south, yshift=4pt},
	bracenode_bottom/.style={midway, anchor=north, yshift=-4pt},
}

\vspace{12pt}
\begin{tikzpicture}
    \newcommand{\cuboid}[8]{
        \begin{scope}
            \newcommand\cuboidCol{#1}
            \newcommand\cuboidWeight{#2}
            \newcommand\cuboidLx{#3}
            \newcommand\cuboidBy{#4}
            \newcommand\cuboidWidth{#5}
            \newcommand\cuboidHeight{#6}
            \newcommand\cuboidDepth{#7}
            \newcommand\shear{#8}
            \newcommand\cuboidRx{#3+\cuboidWidth}
            \newcommand\cuboidTy{#4+\cuboidHeight}
            \newcommand\cuboidDt{\cuboidDepth*\shear}

            \fill[\cuboidCol!20]
                (\cuboidLx, \cuboidBy) --
                (\cuboidRx, \cuboidBy) --
                (\cuboidRx, \cuboidTy) --
                (\cuboidLx, \cuboidTy) --
                cycle;
            \fill[\cuboidCol!20]
                (\cuboidLx,      \cuboidTy) --
                (\cuboidRx,      \cuboidTy) --
                (\cuboidRx-\cuboidDt, \cuboidTy+\cuboidDt) --
                (\cuboidLx-\cuboidDt, \cuboidTy+\cuboidDt) --
                cycle;
            \fill[\cuboidCol!40]
                (\cuboidLx,      \cuboidBy) --
                (\cuboidLx,      \cuboidTy) --
                (\cuboidLx-\cuboidDt, \cuboidTy+\cuboidDt) --
                (\cuboidLx-\cuboidDt, \cuboidBy+\cuboidDt) --
                cycle;
            \draw[\cuboidCol, line width=\cuboidWeight]
                (\cuboidLx, \cuboidBy) --
                (\cuboidRx, \cuboidBy) --
                (\cuboidRx, \cuboidTy) --
                (\cuboidLx, \cuboidTy) --
                cycle;
            \draw[\cuboidCol, line width=\cuboidWeight]
                (\cuboidLx,      \cuboidTy) --
                (\cuboidRx,      \cuboidTy) --
                (\cuboidRx-\cuboidDt, \cuboidTy+\cuboidDt) --
                (\cuboidLx-\cuboidDt, \cuboidTy+\cuboidDt) --
                cycle;
            \draw[\cuboidCol, line width=\cuboidWeight]
                (\cuboidLx,      \cuboidBy) --
                (\cuboidLx,      \cuboidTy) --
                (\cuboidLx-\cuboidDt, \cuboidTy+\cuboidDt) --
                (\cuboidLx-\cuboidDt, \cuboidBy+\cuboidDt) --
                cycle;
        \end{scope}
    }

    \newcommand{\wireFrameCuboid}[8]{
        \begin{scope}
            \newcommand\cuboidCol{#1}
            \newcommand\cuboidWeight{#2}
            \newcommand\cuboidLx{#3}
            \newcommand\cuboidBy{#4}
            \newcommand\cuboidWidth{#5}
            \newcommand\cuboidHeight{#6}
            \newcommand\cuboidDepth{#7}
            \newcommand\shear{#8}
            \newcommand\cuboidRx{#3+\cuboidWidth}
            \newcommand\cuboidTy{#4+\cuboidHeight}
            \newcommand\cuboidDt{\cuboidDepth*\shear}

            \draw[\cuboidCol, line width=\cuboidWeight]
                (\cuboidLx, \cuboidBy) --
                (\cuboidRx, \cuboidBy) --
                (\cuboidRx, \cuboidTy) --
                (\cuboidLx, \cuboidTy) --
                cycle;
            \draw[\cuboidCol, line width=\cuboidWeight]
                (\cuboidLx,      \cuboidTy) --
                (\cuboidRx,      \cuboidTy) --
                (\cuboidRx-\cuboidDt, \cuboidTy+\cuboidDt) --
                (\cuboidLx-\cuboidDt, \cuboidTy+\cuboidDt) --
                cycle;
            \draw[\cuboidCol, line width=\cuboidWeight]
                (\cuboidLx,      \cuboidBy) --
                (\cuboidLx,      \cuboidTy) --
                (\cuboidLx-\cuboidDt, \cuboidTy+\cuboidDt) --
                (\cuboidLx-\cuboidDt, \cuboidBy+\cuboidDt) --
                cycle;
        \end{scope}
    }
    \newcommand{\wireFrameCuboidFrontTop}[8]{
        \begin{scope}
            \newcommand\cuboidCol{#1}
            \newcommand\cuboidWeight{#2}
            \newcommand\cuboidLx{#3}
            \newcommand\cuboidBy{#4}
            \newcommand\cuboidWidth{#5}
            \newcommand\cuboidHeight{#6}
            \newcommand\cuboidDepth{#7}
            \newcommand\shear{#8}
            \newcommand\cuboidRx{#3+\cuboidWidth}
            \newcommand\cuboidTy{#4+\cuboidHeight}
            \newcommand\cuboidDt{\cuboidDepth*\shear}

            \draw[\cuboidCol, line width=\cuboidWeight]
                (\cuboidLx, \cuboidBy) --
                (\cuboidRx, \cuboidBy) --
                (\cuboidRx, \cuboidTy) --
                (\cuboidLx, \cuboidTy) --
                cycle;
            \draw[\cuboidCol, line width=\cuboidWeight]
                (\cuboidLx,      \cuboidTy) --
                (\cuboidRx,      \cuboidTy) --
                (\cuboidRx-\cuboidDt, \cuboidTy+\cuboidDt) --
                (\cuboidLx-\cuboidDt, \cuboidTy+\cuboidDt) --
                cycle;
        \end{scope}
    }

    \newcommand{\wireFrameCuboidLeftTop}[8]{
        \begin{scope}
            \newcommand\cuboidCol{#1}
            \newcommand\cuboidWeight{#2}
            \newcommand\cuboidLx{#3}
            \newcommand\cuboidBy{#4}
            \newcommand\cuboidWidth{#5}
            \newcommand\cuboidHeight{#6}
            \newcommand\cuboidDepth{#7}
            \newcommand\shear{#8}
            \newcommand\cuboidRx{#3+\cuboidWidth}
            \newcommand\cuboidTy{#4+\cuboidHeight}
            \newcommand\cuboidDt{\cuboidDepth*\shear}

            \draw[\cuboidCol, line width=\cuboidWeight]
                (\cuboidLx,      \cuboidTy) --
                (\cuboidRx,      \cuboidTy) --
                (\cuboidRx-\cuboidDt, \cuboidTy+\cuboidDt) --
                (\cuboidLx-\cuboidDt, \cuboidTy+\cuboidDt) --
                cycle;
            \draw[\cuboidCol, line width=\cuboidWeight]
                (\cuboidLx,      \cuboidBy) --
                (\cuboidLx,      \cuboidTy) --
                (\cuboidLx-\cuboidDt, \cuboidTy+\cuboidDt) --
                (\cuboidLx-\cuboidDt, \cuboidBy+\cuboidDt) --
                cycle;
        \end{scope}
    }

    \newcommand{\wireFrameCuboidTop}[8]{
        \begin{scope}
            \newcommand\cuboidCol{#1}
            \newcommand\cuboidWeight{#2}
            \newcommand\cuboidLx{#3}
            \newcommand\cuboidBy{#4}
            \newcommand\cuboidWidth{#5}
            \newcommand\cuboidHeight{#6}
            \newcommand\cuboidDepth{#7}
            \newcommand\shear{#8}
            \newcommand\cuboidRx{#3+\cuboidWidth}
            \newcommand\cuboidTy{#4+\cuboidHeight}
            \newcommand\cuboidDt{\cuboidDepth*\shear}

            \draw[\cuboidCol, line width=\cuboidWeight]
                (\cuboidLx,      \cuboidTy) --
                (\cuboidRx,      \cuboidTy) --
                (\cuboidRx-\cuboidDt, \cuboidTy+\cuboidDt) --
                (\cuboidLx-\cuboidDt, \cuboidTy+\cuboidDt) --
                cycle;
        \end{scope}
    }

    \newcommand{\cuboidGradient}[9]{
        \begin{scope}
            \newcommand\cuboidColA{#1}
            \newcommand\cuboidColB{#2}
            \newcommand\cuboidWeight{#3}
            \newcommand\cuboidLx{#4}
            \newcommand\cuboidBy{#5}
            \newcommand\cuboidWidth{#6}
            \newcommand\cuboidHeight{#7}
            \newcommand\cuboidDepth{#8}
            \newcommand\shear{#9}
            \newcommand\cuboidRx{#4+\cuboidWidth}
            \newcommand\cuboidTy{#5+\cuboidHeight}
            \newcommand\cuboidDt{\cuboidDepth*\shear}

            \fill[left color=\cuboidColA!40, right color=\cuboidColB!20]
                (\cuboidLx, \cuboidBy) --
                (\cuboidRx, \cuboidBy) --
                (\cuboidRx, \cuboidTy) --
                (\cuboidLx, \cuboidTy) --
                cycle;
            \fill[left color=\cuboidColA!40, right color=\cuboidColB!20]
                (\cuboidLx,      \cuboidTy) --
                (\cuboidRx,      \cuboidTy) --
                (\cuboidRx-\cuboidDt, \cuboidTy+\cuboidDt) --
                (\cuboidLx-\cuboidDt, \cuboidTy+\cuboidDt) --
                cycle;
            \fill[color=\cuboidColA!60]
                (\cuboidLx,      \cuboidBy) --
                (\cuboidLx,      \cuboidTy) --
                (\cuboidLx-\cuboidDt, \cuboidTy+\cuboidDt) --
                (\cuboidLx-\cuboidDt, \cuboidBy+\cuboidDt) --
                cycle;
            \draw[\cuboidColB, line width=\cuboidWeight]
                (\cuboidLx, \cuboidBy) --
                (\cuboidRx, \cuboidBy) --
                (\cuboidRx, \cuboidTy) --
                (\cuboidLx, \cuboidTy) --
                cycle;
            \draw[\cuboidColB, line width=\cuboidWeight]
                (\cuboidLx,      \cuboidTy) --
                (\cuboidRx,      \cuboidTy) --
                (\cuboidRx-\cuboidDt, \cuboidTy+\cuboidDt) --
                (\cuboidLx-\cuboidDt, \cuboidTy+\cuboidDt) --
                cycle;
            \draw[\cuboidColB, line width=\cuboidWeight]
                (\cuboidLx,      \cuboidBy) --
                (\cuboidLx,      \cuboidTy) --
                (\cuboidLx-\cuboidDt, \cuboidTy+\cuboidDt) --
                (\cuboidLx-\cuboidDt, \cuboidBy+\cuboidDt) --
                cycle;
        \end{scope}
    }

    \newcommand{\cube}[6]{
        \cuboid{#1}{#2}{#3}{#4}{#5}{#5}{#5}{#6};
    }

    \newcommand{\cubeTriple}[3]{
        \begin{scope}
            \newcommand\cubeSize{0.5}
            \newcommand\cubeShear{0.33}
            \newcommand\cubeLineWidth{0.1}
            \newcommand\cubeDistanceScale{0.75}
            \newcommand\cubeTripleColor{#1}
            \foreach \ix in {3,2,1}
                \cube{\cubeTripleColor}{\cubeLineWidth}{\ix * \cubeDistanceScale + #2}{#3}{\cubeSize}{\cubeShear};
        \end{scope}
    }

    \newcommand{\backbone}[3]{
        \begin{scope}
            \newcommand\backboneColor{#1}
            \newcommand\backboneWeight{0.1}
            \newcommand\backboneShear{0.33}
            \newcommand\backoneX{#2}
            \newcommand\backboneY{#3}
            \newcommand\inputLayerSize{1.25}
            \newcommand\secondLayerSize{0.75}
            \newcommand\layerThickness{0.1}
            \newcommand\backboneShift{0.2}
            \tikzmath{
                \inputLayerOffset = 0.5 * \inputLayerSize;
                \secondLayerOffset = 0.5 * \secondLayerSize;
            }

            \cuboid
                {\backboneColor}{\backboneWeight}
                {\backoneX + \backboneShift}{\backboneY - \secondLayerOffset}
                {\layerThickness}{\secondLayerSize}{\secondLayerSize}{\backboneShear};

            \cuboid
                {\backboneColor}{\backboneWeight}
                {\backoneX - \backboneShift}{\backboneY - \inputLayerOffset}
                {\layerThickness}{\inputLayerSize}{\inputLayerSize}{\backboneShear};

            \end{scope}
    }

    \renewcommand{\grid}[3]{
        \begin{scope}
            \newcommand\gridColor{#1}
            \newcommand\gridWeight{0.1}
            \newcommand\gridPosX{#2}
            \newcommand\gridPosY{#3}
            \newcommand\gridCellWidth{0.6}
            \newcommand\gridCellHeight{0.075}
            \newcommand\gridShear{0.33}
            \newcommand\gridDt{\gridCellWidth*\gridShear}

            \foreach \i in {2,1,0} {
                \foreach \j in {2,1,0} {
                    \cuboid
                        {\gridColor}{\gridWeight}
                        {\gridPosX - \j * \gridDt + \i * \gridCellWidth}
                        {\gridPosY + \j * \gridDt}
                        {\gridCellWidth}{\gridCellHeight}{\gridCellWidth}{\gridShear};
                }
            }
        \end{scope}
    }

    \newcommand{\gridGradient}[4]{
        \begin{scope}
            \newcommand\gridColorA{#1}
            \newcommand\gridColorB{#2}
            \newcommand\gridWeight{0.05}
            \newcommand\gridPosX{#3}
            \newcommand\gridPosY{#4}
            \newcommand\gridCellWidth{0.6}
            \newcommand\gridCellHeight{0.075}
            \newcommand\gridShear{0.33}
            \newcommand\gridDt{\gridCellWidth*\gridShear}

            \cuboidGradient
                {\gridColorA}{\gridColorB}{\gridWeight}
                {\gridPosX}
                {\gridPosY}
                {\gridCellWidth * 3}{\gridCellHeight}{\gridCellWidth * 3}
                {\gridShear};

            \wireFrameCuboid
                {\gridColorB}{\gridWeight}
                {\gridPosX}
                {\gridPosY}
                {\gridCellWidth}{\gridCellHeight}{\gridCellWidth}{\gridShear};

            \foreach \i in {2,1} {
                \foreach \j in {0} {
                    \wireFrameCuboidFrontTop
                        {\gridColorB}{\gridWeight}
                        {\gridPosX - \j * \gridDt + \i * \gridCellWidth}
                        {\gridPosY + \j * \gridDt}
                        {\gridCellWidth}{\gridCellHeight}{\gridCellWidth}{\gridShear};
                }
            }

            \foreach \i in {0} {
                \foreach \j in {2,1} {
                    \wireFrameCuboidLeftTop
                        {\gridColorB}{\gridWeight}
                        {\gridPosX - \j * \gridDt + \i * \gridCellWidth}
                        {\gridPosY + \j * \gridDt}
                        {\gridCellWidth}{\gridCellHeight}{\gridCellWidth}{\gridShear};
                }
            }

            \foreach \i in {2,1} {
                \foreach \j in {2,1} {
                    \wireFrameCuboidTop
                        {\gridColorB}{\gridWeight}
                        {\gridPosX - \j * \gridDt + \i * \gridCellWidth}
                        {\gridPosY + \j * \gridDt}
                        {\gridCellWidth}{\gridCellHeight}{\gridCellWidth}{\gridShear};
                }
            }
        \end{scope}
    }

    \newcommand{\gridCellGradient}[4]{
        \begin{scope}
            \newcommand\gridColorA{#1}
            \newcommand\gridColorB{#2}
            \newcommand\gridWeight{0.1}
            \newcommand\gridPosX{#3}
            \newcommand\gridPosY{#4}
            \newcommand\gridCellWidth{0.5}
            \newcommand\gridCellHeight{0.075}
            \newcommand\gridShear{0.2}
            \newcommand\gridDt{\gridCellWidth*\gridShear}

            \foreach \i in {2,1,0} {
                \foreach \j in {2,1,0} {
                    \cuboidGradient
                    {\gridColorA}{\gridColorB}{\gridWeight}
                    {\gridPosX - \j * \gridDt + \i * \gridCellWidth}
                    {\gridPosY + \j * \gridDt}
                    {\gridCellWidth}{\gridCellHeight}{\gridCellWidth}
                    {\gridShear};
                }
            }
        \end{scope}
    }

    \newcommand{\imageBatch}{
        \begin{scope}
            \newcommand\imageSize{40pt};
            \node[]
            (layer_3) at (0.2*\imageSize, 0.2*\imageSize)
            {\includegraphics[width=\imageSize]{camera_image.jpg}};
            \node[]
            (layer_2) at (0.1*\imageSize, 0.1*\imageSize)
            {\includegraphics[width=\imageSize]{camera_image.jpg}};
            \node[text width=\imageSize]
            (layer_1) at (0,0)
            {\includegraphics[width=\imageSize]{camera_image.jpg}};
        \end{scope}
    }

    \newcommand\streamDistance{135pt}
    \newcommand\horizontalBlockDistance{95pt}
    \newcommand\decoderLayerShift{5pt}

    \node[decoder_layer, xshift=2*\decoderLayerShift+10pt, yshift=2*\decoderLayerShift]
    (decoder_layer_2)
    {};
    \node[decoder_layer, xshift=\decoderLayerShift+10pt, yshift=\decoderLayerShift]
    (decoder_layer_1)
    {};
    \node[decoder_layer, xshift=10pt]
    (decoder_layer_0)
    {};
    \node[label_text, xshift=-32pt, yshift=12pt]
    (decoder_layer_label) at (decoder_layer_0.east |- decoder_layer_0.south)
    {Decoder Layer};

    \draw[brace] ({$(decoder_layer_0.west) + (-2pt, 0pt)$} |- decoder_layer_0.north) -- (decoder_layer_2.west |- {$(decoder_layer_2.north) + (0pt, 2pt)$})
        node [anchor=east, align = center, xshift=-4pt, yshift=2pt] { $l$ };

    \node[]
    (layer_functions)
    {
    \begin{tikzpicture}
        \node[] (static_grid)
        {
            \begin{tikzpicture}
                \grid{sns_blue}{0}{0};
            \end{tikzpicture}
        };
        \node[dashed_round_rectangle, minimum width=75pt, minimum height=50pt]
        (static_grid_highlight) at (static_grid)
        {};
        \node[label_text, yshift=7.5pt]
        (static_grid_label) at (static_grid_highlight.south)
        {$\mathcal{Q}_{\text{BEV}}$};

        \node[static_world_function, right of=static_grid, node distance=\horizontalBlockDistance]
        (static_temporal_self_attention)
        {Temporal \\ Self-Attn.};

        \node[static_world_function, right of=static_temporal_self_attention, node distance=0.8*\horizontalBlockDistance]
        (static_cross_attention)
        {Cross \\ Attention};

        \node[above of=static_grid, node distance=\streamDistance]
        (dynamic_agents)
        {
            \begin{tikzpicture}
                \cubeTriple{sns_orange}{0}{0};
            \end{tikzpicture}
        };
        \node[dashed_round_rectangle, minimum width=75pt, minimum height=50pt]
        (dynamic_agents_highlight) at (dynamic_agents)
        {};
        \node[label_text, yshift=-7.5pt]
        (dynamic_agents_label) at (dynamic_agents_highlight.north)
        {$\mathcal{Q}_{\text{obj}}$};

        \node[dynamic_world_function, right of=dynamic_agents, node distance=\horizontalBlockDistance]
        (dynamic_temporal_self_attention)
        {Temporal \\ Self-Attn.};

        \node[dynamic_world_function, right of=dynamic_temporal_self_attention, node distance=0.8*\horizontalBlockDistance]
        (dynamic_cross_attention)
        {Cross \\ Attention};

        \node[dynamic_world_function, right of=dynamic_cross_attention, node distance=0.8*\horizontalBlockDistance]
        (dynamic_static_cross_attention)
        {Dynamic \\ Static \\ Cross Attn.};

        \node[image_features]
        (image_features) at ($(static_temporal_self_attention)!0.5!(dynamic_temporal_self_attention)$)
        {$\mathcal{F}_t$};
        \node[dashed_round_rectangle, minimum width=65pt, minimum height=50pt]
        (image_features_highlight) at (image_features)
        {};
        \node[label_text, yshift=-7.5pt]
        (image_features_label) at (image_features.south)
        {Image Features};

        \node[left of=static_grid, node distance=\horizontalBlockDistance] (static_in){};
        \node[left of=dynamic_agents, node distance=\horizontalBlockDistance] (dynamic_in){};
        \node[right of=dynamic_static_cross_attention, node distance=70pt] (dynamic_out){};
        \node[](static_out) at (dynamic_out |- static_cross_attention){};

        \draw[dirline, sns_blue] ([xshift=5pt] static_in.east) -- ({$(dynamic_agents.west) + (-5pt, 0pt)$} |- static_in.center);
        \draw[dirline, sns_blue] ([xshift=5pt] static_grid.east) -- ([xshift=-5pt] static_temporal_self_attention.west);
        \draw[dirline, sns_blue] ([xshift=5pt] static_temporal_self_attention.east) -- ([xshift=-5pt] static_cross_attention.west);
        \draw[dirline, sns_blue] ([xshift=5pt] static_cross_attention.east) -| ([yshift=-5pt] dynamic_static_cross_attention.south);
        \draw[dirline, sns_blue] ([xshift=5pt] static_cross_attention.east) -- ([xshift=-5pt] static_out.west);
        \draw[dirline, sns_blue] ([yshift=-15pt] static_grid.south) |-
             ($(static_grid.south)!0.5!(static_temporal_self_attention.south) + (0pt,-15pt)$) -|
                ([yshift=-2.5pt] static_temporal_self_attention.south);

        \draw[dirline, sns_orange] ([xshift=5pt] dynamic_in.east) -- ([xshift=-5pt] dynamic_agents.west);
        \draw[dirline, sns_orange] ([xshift=7.5pt] dynamic_agents.east) -- ([xshift=-5pt] dynamic_temporal_self_attention.west);
        \draw[dirline, sns_orange] ([xshift=5pt] dynamic_temporal_self_attention.east) -- ([xshift=-5pt] dynamic_cross_attention.west);
        \draw[dirline, sns_orange] ([xshift=5pt] dynamic_cross_attention.east) -- ([xshift=-5pt] dynamic_static_cross_attention.west);
        \draw[dirline, sns_orange] ([xshift=5pt] dynamic_static_cross_attention.east) -- ([xshift=-5pt] dynamic_out.west);
        \draw[dirline, sns_orange] ([yshift=15pt] dynamic_agents.north) |-
             ($(dynamic_agents.north)!0.5!(dynamic_temporal_self_attention.north) + (0pt,15pt)$) -|
                ([yshift=2.5pt] dynamic_temporal_self_attention.north);

        \draw[dirline, sns_purple] ([xshift=5pt] image_features.east) -| ([yshift=5pt] static_cross_attention.north);
        \draw[dirline, sns_purple] ([xshift=5pt] image_features.east) -| ([yshift=-5pt] dynamic_cross_attention.south);

        \node[label_text, above right of=dynamic_in, node distance=10pt, xshift=15pt]
        (obj_T_obj)
        {${}^{\text{obj},t}T_{\text{obj},t-1}$};
        \node[label_text, below right of=dynamic_in, node distance=10pt, xshift=15pt]
        (ego_T_ego)
        {${}^{\text{ego},t}T_{\text{ego},t-1}$};
        \node[label_text, below right of=static_in, node distance=10pt, xshift=15pt]
        (ego_T_ego)
        {${}^{\text{ego},t}T_{\text{ego},t-1}$};

        \node[label_text,sns_orange]
        (kv_dynamic_agents) at ($(dynamic_agents.north)!0.5!(dynamic_temporal_self_attention.north) + (0pt,8pt)$)
        {$\mathcal{K},\mathcal{V}$};
        \node[label_text,sns_orange]
        (q_dynamic_agents) at ($(dynamic_agents.east)!0.5!(dynamic_temporal_self_attention.west) + (0pt,-8pt)$)
        {$\mathcal{Q}$};
        \node[label_text,sns_orange]
        (q_dynamic_temporal_self_attention) at ($(dynamic_temporal_self_attention.east)!0.5!(dynamic_cross_attention.west) + (0pt,-8pt)$)
        {$\mathcal{Q}$};
        \node[label_text,sns_orange]
        (q_dynamic_static_temporal_self_attention) at ($(dynamic_cross_attention.east)!0.5!(dynamic_static_cross_attention.west) + (0pt,-8pt)$)
        {$\mathcal{Q}$};
        \node[label_text,sns_orange]
        (q_dynamic_out) at ($(dynamic_static_cross_attention.east) + (8pt,-8pt)$)
        {$\mathcal{Q}$};
        \node[label_text,sns_blue]
        (kv_static_grid) at ($(static_grid.south)!0.5!(static_temporal_self_attention.south) + (0pt,-8pt)$)
        {$\mathcal{K},\mathcal{V}$};
        \node[label_text,sns_blue]
        (q_static_grid) at ($(static_grid.east)!0.5!(static_temporal_self_attention.west) + (0pt,8pt)$)
        {$\mathcal{Q}$};
        \node[label_text,sns_blue]
        (q_static_temporal_self_attention) at ($(static_temporal_self_attention.east)!0.5!(static_cross_attention.west) + (0pt,8pt)$)
        {$\mathcal{Q}$};
        \node[label_text,sns_blue]
        (q_static_out) at (q_dynamic_out |- {$(static_cross_attention.west) + (0pt,8pt)$})
        {$\mathcal{Q}$};
        \node[label_text,sns_blue]
        (kv_static_grid) at ($(static_cross_attention.east)!0.5!(dynamic_static_cross_attention.south east)$)
        {$\mathcal{K},\mathcal{V}$};

        \node[label_text,sns_purple]
        (kv_static_grid) at ({$(image_features.east)!0.5!(dynamic_cross_attention.south)$} |- {$(image_features.west) + (0pt,8pt) $})
        {$\mathcal{K},\mathcal{V}$};
    \end{tikzpicture}
    };
\end{tikzpicture}

%% file: sections/4_experiments.tex
\section{Experiments}
We evaluate the performance of \methodName\ on the challenging and well-established nuScenes dataset~\cite{caesar2020nuscenes}. Additionally, we integrate our proposed approach into two \ac{sota} end-to-end trainable driving frameworks, \ie  UniAD~\cite{hu2023planning} and VAD~\cite{jiang2023vad}.  We perform extensive ablation studies to evaluate the effect of our design choices and provide additional insights as well as qualitative results.

\boldparagraph{Dataset}
We utilize the large-scale nuScenes dataset~\cite{caesar2020nuscenes} consisting of 1000 scenes and use the official train- and val-set split. We adopt the official task definitions for the object detection task~\cite{nuscenes_detection_benchmark} and object tracking task~\cite{nuscenes_tracking_benchmark}, respectively, and follow other recent works~\cite{gu2023vip3d, hu2023planning, jiang2023vad} for the definition of the motion prediction and planning objectives.

\boldparagraph{Metrics}
For object detection, we report the main metrics \ac{map} and \ac{nds} computed on all ten classes of the dataset, as well as true positive metrics such as the \ac{mate}, \ac{maoe}, and \ac{mave} as defined in~\cite{nuscenes_detection_benchmark}. For object tracking, we follow the official metric definition in~\cite{nuscenes_tracking_benchmark} and report \ac{amota} and  \ac{amotp} as well as recall and \ac{ids}.
For map segmentation, we closely follow~\cite{hu2023planning} and report \ac{bev} segmentation \ac{iou} for different classes. We refer the reader to~\cite{hu2023planning} for additional details on the metric and class definitions.
For motion prediction, we report the \ac{epa}~\cite{gu2023vip3d} as main metric as well as the true positive metrics \ac{minade} and \ac{minfde}.
For open-loop planning, we report the L2 distance to the ego trajectory as well as collision rates for $\SI{1}{\second}$ and $\SI{3}{\second}$, respectively.
A more detailed evaluation including additional metrics for the different configurations of our approach can be found in the supplementary.

\boldparagraph{Training Configuration}
We closely follow the settings in~\cite{wang2023exploring, hu2023planning, jiang2023vad} to increase comparability.
Unless otherwise specified, we utilize a VovNet-V2-99~\cite{lee2019energy} and an image resolution of $800 \times 320$ pixels.
For details on the backbone and FPN~\cite{lin2017feature} configuration, we refer the reader to~\cite{wang2023exploring, streampetr_github}.
Following~\cite{wang2023exploring}, we utilize streaming video training.
All models are trained for 24 epochs utilizing a batch size of eight on eight NVIDIA A100 GPUs with AdamW~\cite{loshchilov2017decoupled}, a learning rate of $2e^{-4}$ and a cosine annealing schedule.
\methodName\ performs object detection, tracking, map segmentation, and motion prediction, as well as open-loop planning.
Note that all tasks are performed jointly in one multi-task model.
Following~\cite{hu2023planning}, we train the model in a two-stage fashion: the stage-I model only performs object detection, tracking and map segmentation, while the stage-II model is optimized for all tasks in an end-to-end fashion with a frozen image backbone for numerical stability. In all perception experiments, we \mbox{append -I} or -II for clarity \eg \methodName-II.

\boldparagraph{Baselines}
Given that our approach follows the object-query propagation technique used by StreamPETR~\cite{wang2023exploring}, and considering that StreamPETR reaches \ac{sota} performance on the nuScenes benchmark~\cite{nuscenes_detection_benchmark}, we choose it as the main baseline for dynamic object perception.
To demonstrate the performance gains of our proposed architecture along the entire functional chain, we evaluate the performance of \methodName\ on downstream tasks for driving, such as motion prediction and open-loop planning. We choose the two recent \ac{sota} approaches UniAD~\cite{hu2023planning} and VAD~\cite{jiang2023vad} as baselines since they perform all tasks in an end-to-end trainable fashion while also following the two-stage training paradigm.
If not all metrics are reported in the corresponding publications, we utilize the published training logs and code to reproduce the results. For tasks without an official benchmark, we adopt the evaluation scheme of~\cite{hu2023planning, jiang2023vad}. It has to be noted that VAD~\cite{jiang2023vad} utilizes a ResNet-50 and a smaller detection range for the perception and motion prediction tasks. For all comparisons with VAD~\cite{jiang2023vad}, we configure our model to exactly follow their settings for a fair comparison.
We refer the reader to~\cite{jiang2023vad, vad_github} for additional details on the configuration and used detection ranges.

\subsection{Perception Sub-Task Results}
In this section, we analyze the performance of our \mbox{stage-I} model trained on perception tasks only, to allow for a fair comparison with \ac{sota} models specialized for perception.

\begin{table}
  \begin{center}
    \caption{\textbf{Object Detection}. \methodName\ outperforms task-specific models as well as end-to-end models on all metrics. We report performance for perception as well as stage-II results for end-to-end models. *Results taken from official repository~\cite{uniad_github}. $\mathparagraph$ indicates a version that uses a ResNet-101~\cite{he2016deep}, as in UniAD~\cite{hu2023planning}.}
    \label{table:results_detection}
    \begin{adjustbox}{width=\columnwidth,center}
    \begin{tabular}{lcccccccc}
      \toprule
      Name                                     & mAP$\uparrow$ & mATE$\downarrow$ & mAOE$\downarrow$ & mAVE$\downarrow$ & \highlight{NDS$\uparrow$} \\
      \midrule
      BEVFormer\cite{li2022bevformer}          & $41.6$        & $0.67$           & $0.37$     & $0.27$           & \highlight{$51.7$}        \\
      SteamPETR\cite{wang2023exploring}        & $48.2$        & $0.60$           & $0.37$     & $0.26$           & \highlight{$57.1$}        \\
      \uniadSOne*\cite{hu2023planning}         & $39.5$        & $0.66$           & $0.36$    & $0.40$           & \highlight{$50.6$}        \\
      \textbf{\methodNameSOne}$\mathparagraph$ & $48.2$        & $0.59$           & $\mb{0.32}$     & $0.28$           & \highlight{$57.4$}        \\
      \textbf{\methodNameSOne}                 & $\mb{49.5}$   & $\mb{0.57}$      & $0.39$  & $\mb{0.26}$      & \highlight{$\mb{57.8}$}   \\
      \midrule
      \uniadSTwo\cite{hu2023planning}          & $38.1$        & $0.68$           & $0.38$    & $0.38$           & \highlight{$49.8$}        \\
      \textbf{\methodNameSTwo}                 & $48.1$        & $0.57$           & $0.41$   & $0.28$           & \highlight{$56.6$}        \\
      \bottomrule
    \end{tabular}
    \end{adjustbox}
  \end{center}
\end{table}

\boldparagraph{Object Detection}
The object detection performance on the nuScenes validation set~\cite{caesar2020nuscenes} is shown in~\tabref{table:results_detection}. Compared to other multi-task models like UniAD~\cite{hu2023planning}, \methodName\ yields an improvement of $+10$ ($+\SI{20}{\percent}$) in terms of \ac{map} and $+7.2$ $(+\SI{12}{\percent})$ in \ac{nds} respectively. Compared to StreamPETR~\cite{wang2023exploring}, which is only trained on object detection, \methodName\ gains an improvement of $+1.3$ \ac{map} and $+0.7$ NDS, resulting in \ac{sota} performance for object detection. While our model builds on StreamPETR, we attribute the key improvements over StreamPETR to the newly introduced dynamic-static cross-attention that allows object-queries to benefit from the static scene structure.

\boldparagraph{Map Segmentation}
The results for map segmentation of different classes are shown in~\tabref{table:results_map_segmentation}. \methodName\ yields comparable performance to \ac{sota} approaches on all classes while improving the lane segmentation by $+2.9$ \ac{iou} as compared to UniAD using the same image backbone.
As indicated in \cite{hu2023planning}, the observed improvements in lane segmentation are vital for the accurate perception of dynamic agents. We want to highlight that our two-stream design allows the use of a single static \ac{bev} encoder instead of having a unified \ac{bev} encoder and an additional static map encoder. This results in $\SI{2.7}{M}$ ($-\SI{15}{\percent}$) fewer parameters for map segmentation as compared to UniAD. 

\begin{table}
  \caption{\textbf{Map Segmentation}. \methodName\ achieves competitive performance, especially improving the segmentation of lanes.
    $\mathparagraph$ indicates a version that uses a ResNet-101~\cite{he2016deep} as in UniAD~\cite{hu2023planning}, $^-$ a version only trained on the mapping task without dynamic agents. We report performance for perception as well as stage-II results for end-to-end models. *Results taken from official repository~\cite{uniad_github}. Segmentation \ac{iou}(\%) is reported for different classes.}
  \label{table:results_map_segmentation}
  \begin{adjustbox}{width=\columnwidth,center}
    \begin{tabular}{lcccc}
      \toprule
      Name                                         & \highlight{Lanes$\uparrow$} & Drivable$\uparrow$ & Divider$\uparrow$ & Crossing$\uparrow$ \\
      \midrule
      BEVFormer\cite{li2022bevformer}              & \highlight{$23.9$}          & $\mb{77.5}$        & -                 & -                  \\
      BEVerse\cite{zhang2022beverse}               & \highlight{-}               & -                  & $30.6$            & $\mb{17.2}$        \\
      \uniadSOne*~\cite{hu2023planning}            & \highlight{$31.3$}          & $69.1$             & $25.7$            & $13.8$             \\
      \textbf{\methodNameSOne$\mathparagraph$}     & \highlight{$34.2$}          & $69.7$             & $29.7$            & $13.8$             \\
      \textbf{\methodNameSOne}                     & \highlight{$34.6$}          & $70.5$             & $30.2$            & $12.8$             \\
      \textbf{\methodNameSOne}$\mathparagraph^{-}$ & \highlight{$\mb{35.6}$}     & $71.1$             & $\mb{32.3}$       & $15.1$             \\
      \midrule
      \uniadSTwo*\cite{hu2023planning}             & \highlight{$31.2$}          & $69.1$             & $25.9$            & $14.3$             \\
      \textbf{\methodNameSTwo}                     & \highlight{$34.1$}          & $70.0$             & $29.9$            & $12.2$             \\
      \bottomrule
    \end{tabular}%
  \end{adjustbox}
\end{table}

\boldparagraph{Multiple Object Tracking} The tracking performance of our model is shown in~\tabref{table:results_tracking}. In contrast to explicit tracking as in~\cite{hu2023planning, doll2023star} our model only performs implicit tracking through query propagation. We utilize the widely adopted tracking approach presented in~\cite{yin2021center} for a fair comparison to StreamPETR~\cite{wang2023exploring}. \methodName\ reaches \ac{sota} performance in terms of \ac{amota} and outperforms specialized tracking approaches like PF-Track~\cite{pang2023standing} by a large margin.
Compared to UniAD~\cite{hu2023planning}, our approach heavily improves the \ac{amota} by $+15.8$ ($+\SI{28}{\percent}$) and by $+2.5$ ($+\SI{4}{\percent}$) compared to the specialized StreamPETR, respectively. In particular, the dual-stream layout leads to a higher temporal consistency of tracked objects, manifesting in a reduction of \ac{ids} by $\SI{25}{\percent}$ compared to UniAD~\cite{hu2023planning} and StreamPETR~\cite{wang2023exploring}.
\begin{table}
  \begin{center}
    \caption{\textbf{Multiple Object Tracking}. \methodName\ achieves high temporal consistency, heavily outperforming previous \ac{sota} approaches.
      $\mathparagraph$ indicates a version that uses a ResNet-101~\cite{he2016deep}, as in UniAD~\cite{hu2023planning}, $^+$ tracking-by-detection approach reimplemented with BEVFormer~\cite{li2022bevformer} in~\cite{hu2023planning}. We report performance for perception as well as stage-II results for end-to-end models. *Results taken from official repository~\cite{uniad_github}.}
    \label{table:results_tracking}
    \begin{adjustbox}{width=\columnwidth,center}
      \begin{tabular}{lcccc}
        \toprule
        Name                                     & \highlight{AMOTA$\uparrow$} & AMOTP$\downarrow$ & Recall$\uparrow$ & IDS$\downarrow$ \\
        \midrule
        ImmortalTrack$^+$\cite{wang2021immortal} & \highlight{37.8}            & 1.11              & 47.8             & 936             \\
        PF-Track\cite{pang2023standing}          & \highlight{47.9}            & 1.22              & 59.0             & $\mb{181}$      \\
        StreamPETR\cite{wang2023exploring}       & \highlight{52.6}            & 1.12              & 59.9             & 886             \\
        \uniadSOne*\cite{hu2023planning}         & \highlight{39.3}            & 1.29              & 48.1             & 894             \\
        \textbf{\methodNameSOne}$\mathparagraph$ & \highlight{52.3}            & 1.12              & 60.7             & 726             \\
        \textbf{\methodNameSOne}                 & \highlight{$\mb{55.1}$}     & $\mb{1.08}$       & $\mb{60.7}$      & 663             \\
        \midrule
        \uniadSTwo*\cite{hu2023planning}         & \highlight{36.3}            & 1.34              & 45.9             & 1177            \\
        \textbf{\methodNameSTwo}                 & \highlight{52.6}            & 1.11              & 59.6             & 774             \\
        \bottomrule
      \end{tabular}
    \end{adjustbox}
  \end{center}
\end{table}

\boldparagraph{Runtime} \methodNameSOne\ perception takes an average runtime of $\SI{193}{\milli \second}$. The new Dynamic-Static Cross-Attention only adds $\SI{2.12}{\milli \second}$ per block, corresponding to $\SI{6}{\percent}$ of the total runtime (see Supplementary).
A small configuration with $\SI{107}{\milli \second}$ latency reaches $\SI{40.1}{\percent}$ \ac{map} and therefore still outperforms UniAD~\cite{hu2023planning} while being four times faster.

\subsection{Integration to End-to-End Pipelines}
To demonstrate the performance gains for downstream tasks like motion prediction and planning, we integrate our proposed dual-stream architecture into recent end-to-end trainable driving frameworks that reach \ac{sota} results: UniAD~\cite{hu2023planning} and VAD~\cite{jiang2023vad}. 
Due to the training focus on the final motion prediction and planning performance, we observe, similar to UniAD~\cite{hu2023planning}, a slight degradation in perception performance when training in the end-to-end setting as shown in~\tabref{table:results_detection},~\tabref{table:results_map_segmentation} and~\tabref{table:results_tracking}.

\boldparagraph{Motion Prediction}
The results for motion prediction are shown in~\tabref{table:results_prediction}. Our model significantly outperforms UniAD by $+6.8$ ($+\SI{12}{\percent}$) \ac{epa} for the vehicle class and by $+9.7$ ($+\SI{21}{\percent}$) for pedestrians, respectively.
A similar effect is observed for the vectorized framework VAD~\cite{jiang2023vad} where our model improves motion prediction by $+5.1$ ($+\SI{7}{\percent}$) \ac{epa} for vehicles. We attribute this to improved modelling of dynamic agents in the scene and improved motion queues by direct object to image attention.

\begin{table}
  \begin{center}
    \caption{\textbf{Motion Prediction}.
      \methodName\ remarkably improves the motion prediction task within different frameworks.
      V denotes the vehicle category and P pedestrians, respectively. *Results taken from official repository~\cite{uniad_github, vad_github}. Please note that VAD~\cite{jiang2023vad} and hence our integration \textsc{DualVAD} use a smaller detection range and that results are not directly comparable with other approaches.}
    \label{table:results_prediction}
    \begin{adjustbox}{width=\columnwidth,center}
      \begin{tabular}{lcc|cc|cc}
        \toprule
                                    & \multicolumn{2}{c|}{EPA$\uparrow$} & \multicolumn{2}{c|}{minADE$\downarrow$} & \multicolumn{2}{c}{minFDE$\downarrow$}                                           \\
        Name                        & \highlight{V}                      & P                                       & V                                      & P           & V           & P           \\
        \midrule
        VIP3D\cite{gu2023vip3d}     & \highlight{22.2}                   & -                                       & 2.05                                   & -           & 1.95        & -           \\
        UniAD*\cite{hu2023planning} & \highlight{45.6}                   & 35.5                                    & 0.71                                   & 0.78        & $1.02$      & 1.05        \\
        \textbf{\methodName}        & \highlight{52.4}                   & 45.2                                    & 0.68                                   & $\mb{0.63}$ & $1.08$      & $0.89$      \\
        \midrule
        VAD*~\cite{jiang2023vad}    & \highlight{64.7}                   & 47.4                                    & 0.68                                   & 0.67        & 0.92        & $\mb{0.84}$        \\

        \textbf{\textsc{DualVAD}}   & \highlight{$\mb{69.8}$}            & $\mb{47.9}$                             & $\mb{0.60}$                            & 0.67        & $\mb{0.83}$ & 0.85 \\
        \bottomrule
      \end{tabular}
    \end{adjustbox}
  \end{center}
\end{table}

\begin{table}
  \begin{center}
    \caption{\textbf{Open-Loop Planning}.
      \methodName\ achieves lowest L2 error and collision rate, especially for longer planning horizons.
      Avg describes the mean over the 1s, 2s and 3s values,
      \textsc{DualVAD} a version of our model that follows the VAD~\cite{jiang2023vad} framework.}%
    \label{table:results_planning}%
    \begin{adjustbox}{width=\columnwidth,center}
      \begin{tabular}{lccc|ccc}
        \toprule
                                   & \multicolumn{3}{c|}{L2 (m) $\downarrow$} & \multicolumn{3}{c}{Col (\%) $\downarrow$}                                                                                 \\
        Name                       & 1s                                      & 3s                                       & \highlight{Avg}         & 1s          & 3s          & \highlight{Avg}         \\
        \midrule
        UniAD\cite{hu2023planning} & 0.48                                    & 1.65                                     & \highlight{1.03}        & 0.05        & 0.71        & \highlight{0.31}        \\
        \textbf{\methodName}       & 0.56                                    & 1.55                                     & \highlight{1.03}        & $\mb{0.03}$ & $\mb{0.35}$ & \highlight{$\mb{0.17}$} \\      \midrule  
        VAD\cite{jiang2023vad}     & $0.41$                                  & $1.05$                                   & \highlight{0.72}        & 0.07        & $0.41$      & \highlight{0.22}        \\
        \textbf{\textsc{DualVAD}}  & $\mb{0.30}$                             & $\mb{0.82}$                              & \highlight{$\mb{0.55}$} & 0.11        & 0.36        & \highlight{0.22}        \\
        \bottomrule
      \end{tabular}
    \end{adjustbox}
  \end{center}
\end{table}

\begin{table*}
  \begin{center}
    \caption{\textbf{Ablation on the Interaction Design}. $\dagger$ denotes UniAD with the detection head of StreamPETR~\cite{wang2023exploring} instead of the tracking-by-attention head proposed in~\cite{zhang2022mutr3d}. $\emptyset$ denotes a variant of our approach without stream interaction and $\updownarrow$ a variant that uses bidirectional stream interaciton.}
    \label{table:interaction_design}
    \begin{adjustbox}{width=\textwidth,center}
    \begin{tabular}{lccccc|ccccc}
      \toprule
      Name                                     & w/Det  & w/Map  & \vtop{\hbox{\strut Temporal}\hbox{\strut BEV}} & \vtop{\hbox{\strut Obj2Img}\hbox{\strut Attn}} & \#Params & mAP$\uparrow$ & NDS$\uparrow$ & Lanes$\uparrow$ & AMOTA$\uparrow$ & IDS$\downarrow$ \\
      \midrule
      \uniadSOne~\cite{hu2023planning}         & \cmark & \cmark & \cmark                                         & \xmark                                         & -        & 39.5          & 50.6          & 29.3            & 39.3            & 894             \\
      StreamPETR~\cite{wang2023exploring}      & \cmark & \xmark & \xmark                                         & \cmark                                         & -        & 48.2          & 57.1          & -               & 52.6            & 886             \\
      \midrule
      \uniadSOne$\dagger$\cite{hu2023planning} & \cmark & \cmark & \cmark                                         & \xmark                                         & 111M     & 43.3          & 51.2          & 33.9            & 45.0            & 1764            \\
      \uniadSOne$\dagger$\cite{hu2023planning} & \cmark & \cmark & \xmark                                         & \xmark                                         & 111M     & 45.3          & 54.9          & 34.3            & 49.4            & 779             \\
      \textbf{\methodNameSOne}                 & \cmark & \cmark & \xmark                                         & \cmark                                         & 118M     & 46.9          & 56.1          & 31.7            & 51.6            & 658            \\
      \textbf{\methodNameSOne}$\emptyset$      & \cmark & \cmark & \cmark                                         & \cmark                                         & 113M     & 47.7          & 55.7          & 33.9            & 51.9            & 769             \\
      \textbf{\methodNameSOne}$\updownarrow$   & \cmark & \cmark & \cmark                                         & \cmark                                         & 120M     & 49.3          & 57.6          & 33.8            & 54.4            & $\mb{588}$      \\
      \textbf{\methodNameSOne}                 & \cmark & \cmark & \cmark                                         & \cmark                                         & 118M     & $\mb{49.5}$   & $\mb{57.7}$   & $\mb{34.6}$     & $\mb{55.1}$     & $663$           \\
      \bottomrule
    \end{tabular}
    \end{adjustbox}
  \end{center}
\end{table*}

\boldparagraph{Open-Loop Planning}
The evaluation of the open-loop planning performance is provided in~\tabref{table:results_planning}. While we report those results for completeness, we want to highlight the issues on open-loop planning in nuScenes~\cite{caesar2020nuscenes} recently identified in~\cite{zhai2023rethinking}. We integrate our approach into the planning modules proposed in~\cite{hu2023planning} and~\cite{jiang2023vad} that do not use the ego status as direct input to planning and follow their corresponding evaluation protocols~\cite{uniad_github, vad_github}.
As shown in~\tabref{table:results_planning}, \methodName\ reaches comparable performance in terms of $L2$ distance compared to UniAD~\cite{hu2023planning} and heavily reduces the collision rate up to a factor of two for longer planning horizons.
Similarly, for VAD~\cite{jiang2023vad} the $L2$ error is reduced by up to $\SI{0.23}{\meter}$ ($-\SI{21}{\percent}$) depending on the planning horizon.
This is in line with our model design since improving temporal consistency, especially for dynamic agents, might become more relevant for longer planning horizons.

\begin{figure*}
     \centering
     \begin{subfigure}[c]{0.7857\textwidth}
         \centering
         \includegraphics[width=\textwidth]{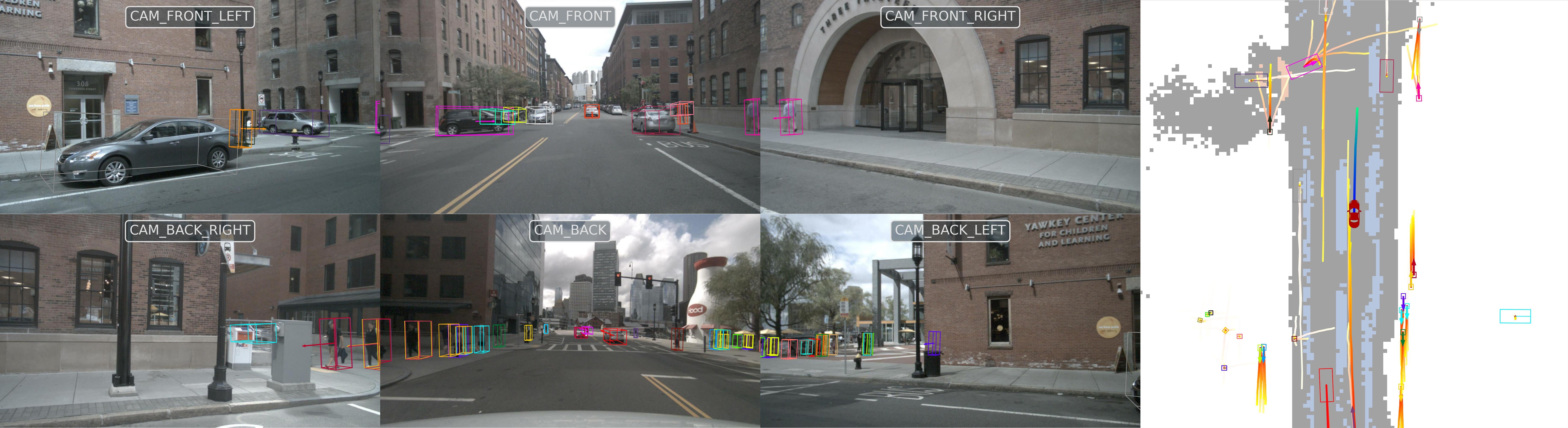}
         \caption{\methodName}
         \label{fig:qualitative_uniad}
     \end{subfigure}%
     \begin{subfigure}[c]{0.2143\textwidth}
         \includegraphics[width=\textwidth]{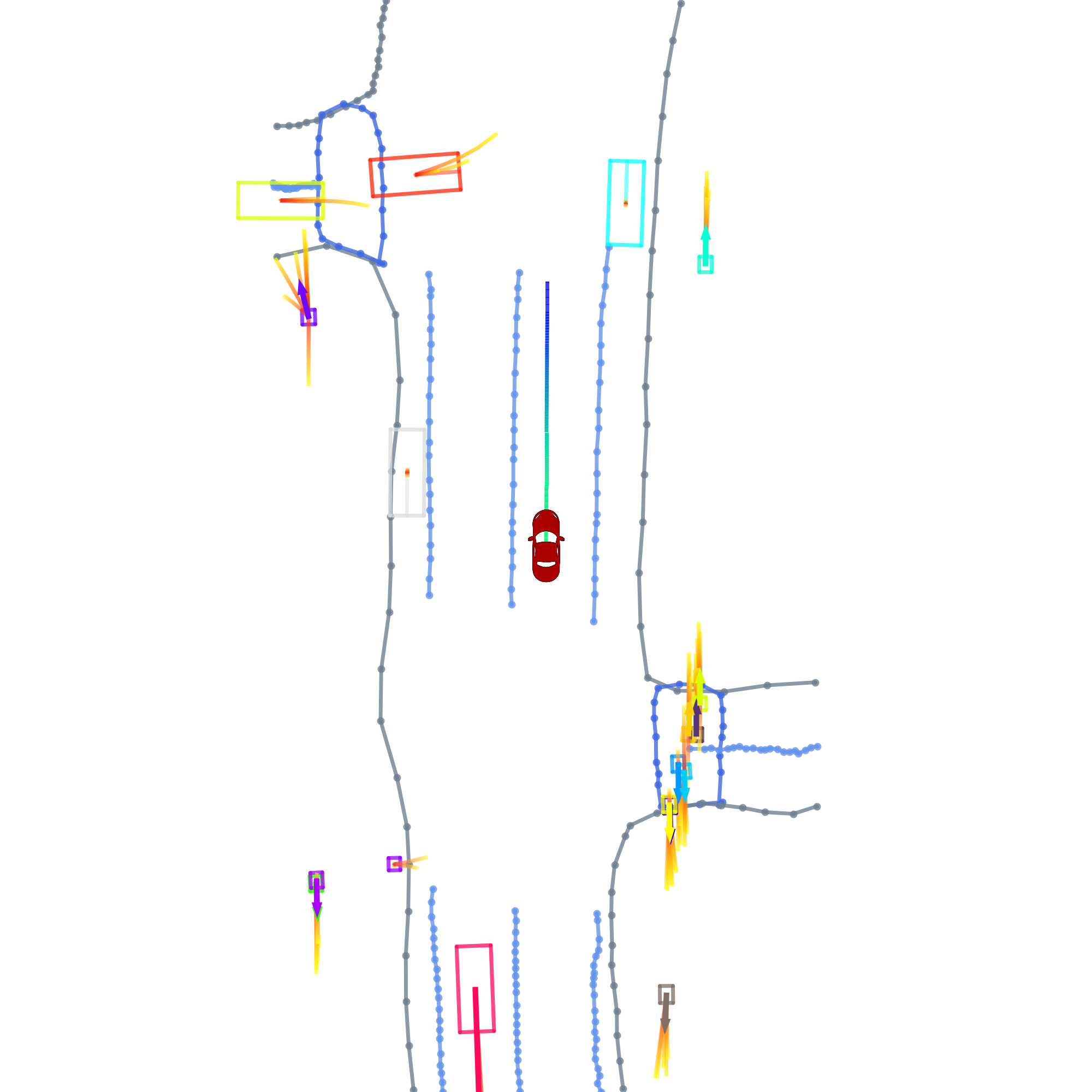}
         \caption{\textsc{DualVAD}}
         \label{fig:qualitative_vad}
     \end{subfigure}%
    \caption{\textbf{Qualitative Results}.~\figref{fig:qualitative_uniad} shows the output of \methodName\ for object tracking, map segmentation, motion prediction and planning.~\figref{fig:qualitative_vad} shows the same scene for the vectorized version \textsc{DualVAD} of our approach.}
    \label{fig:qualitative_results}
\end{figure*}

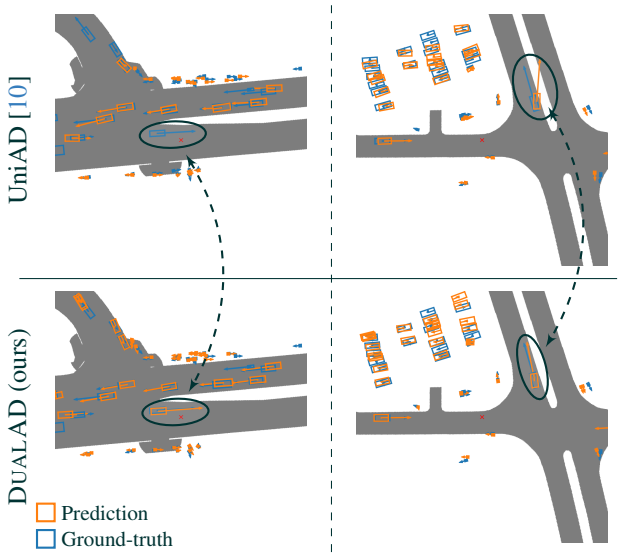
\begin{figure}[t]
  \centering
  \input{diagrams/qualitative_comparison.tikz}
  \caption{\textbf{Performance Comparison} of \methodName\ and UniAD~\cite{hu2023planning} for two different scenes. Predictions are shown in orange, ground-truth annotations in blue, ego location with a red cross. While highly dynamic agents cause perception errors such as track losses or distorted objects for UniAD, \methodName\ consistently captures them due to the proposed dual-stream design.}
  \label{fig:qualitative_comparison}
\end{figure}

\boldparagraph{Qualitative Results}
\figref{fig:qualitative_results} visualizes the performance of \methodName\ in a complex traffic scene. The proposed dual-stream design enables a temporally consistent perception of the surrounding area, including highly dynamic agents and allows for precise motion prediction and planning. A comparison to the perception of UniAD~\cite{hu2023planning} with respect to highly dynamic agents is shown in~\figref{fig:qualitative_comparison}. Additional examples for both integrated frameworks~\cite{hu2023planning, jiang2023vad} can be found in the supplementary.

\subsection{Ablations}
\boldparagraph{Effect of Interaction Design}
An evaluation of the different design choices of \methodName\ is shown in~\tabref{table:interaction_design}. First, using a version of our approach with two separate streams without the proposed dynamic-static cross-attention module especially decreases the object detection and tracking performance. This observation confirms the benefits of the interaction with the static branch for dynamic agents. Second, using a bidirectional stream interaction, where also the static \ac{bev}-queries attend to the dynamic agents, does not yield significant improvements. This is in line with our hypothesis that it is sufficient for the static world representation to perform cross-attention to the images paired with temporal self-attention. Third, we incorporate StreamPETR's query propagation~\cite{wang2023exploring} as used in our approach to UniAD~\cite{hu2023planning}, which yields consistent improvements but heavily increases the \ac{ids} which might be a result of the simple greedy tracker~\cite{yin2021center}. Fourth, \methodName\ benefits as expected from using temporal attention for the \ac{bev}-queries, but we observe the opposite for UniAD~\cite{hu2023planning}. A version without temporal attention within the unified \ac{bev}-grid in UniAD$\dagger$ has a performance for dynamic agent perception that is increased by $+3.7$ \ac{nds} and $4.4$ \ac{amota}, respectively, while \ac{ids} are drastically reduced by $\SI{55}{\percent}$. This supports our claim that the unified grid is not well-suited to propagate information about dynamic agents through time.

\begin{table}
  \begin{center}
    \caption{\textbf{Ablation on Temporal Consistency}.
      \methodName\ achieves consistent tracks even without sensor measurements from each sensor in each frame.
      We mimic the effect of non-synchronized sensors by using the front and back facing cameras in an alternating fashion (denoted by $\circleddash$).}
    \label{table:cam_drop}
    \begin{adjustbox}{width=\columnwidth,center}
      \begin{tabular}{lcccc}
        \toprule
        Name                                          & mAP$\uparrow$ & Lanes$\uparrow$ & AMOTA$\uparrow$ & \highlight{IDS$\downarrow$} \\
        \midrule
        \uniadSOne\cite{hu2023planning}               & 39.5          & 29.3            & 39.3            & \highlight{894}             \\
        \uniadSOne$\circleddash$\cite{hu2023planning} & 36.0          & 28.9            & 28.3           & \highlight{2062}            \\
        \textbf{\methodNameSOne}                      & $\mb{49.5}$   & $\mb{34.6}$     & $\mb{55.1}$     & \highlight{$\mb{663}$}      \\
        \textbf{\methodNameSOne$\circleddash$}        & 42.8          & 31.5            & 44.4            & \highlight{940}             \\
        \bottomrule
      \end{tabular}
    \end{adjustbox}
  \end{center}
\end{table}

\boldparagraph{Temporal Consistency, Non-Synchronized Sensors}
Since our proposed dual-stream design can flexibly propagate the belief state through time, we run \methodName~in a setting with non-synchronized sensors.
To do so, we split the camera images into two sets $\mathcal{C}_{front}$ containing the three front-facing cameras and  $\mathcal{C}_{back}$ for the rear cameras respectively. We use the inputs of $\mathcal{C}_{front}$ and $\mathcal{C}_{back}$ in an alternating fashion, leading to three camera inputs per time step and an effective refresh rate per camera of $\SI{1}{\hertz}$.
The resulting performance is shown in~\tabref{table:cam_drop}. 
The restriction of sensor data to one half of the scene per time step decreases the performance for both approaches.
We observe that the \ac{ids} of UniAD~\cite{hu2023planning} doubles, while \methodName\ keeps consistent tracks with only $\SI{29}{\percent}$ increase in \ac{ids}. This observation again highlights the effectiveness of the dual-stream design.

\boldparagraph{Effect on Highly Dynamic Agents}
Following our hypothesis that the dual-stream design should especially improve the detection of agents with higher velocities, we conduct an experiment in which we focus on highly dynamic agents. More specifically, we evaluate the object detection performance on objects of the type car, where both, the absolute and the relative velocity with respect to the ego vehicle, are higher than $\SI[quotient-mode=fraction, fraction-function=\dfrac]{10}{\meter/\second}$. The performance of UniAD~\cite{hu2023planning} drops to $36.3$ ($-\SI{38}{\percent}$) \ac{map} while \methodName\ drops to $47.3$ ($-\SI{25}{\percent}$) \ac{map}, yielding an increased performance delta of $5.5$\ac{map}.
We observe on the one hand that for both approaches the detection of highly dynamic objects is particularly challenging. On the other hand, these results confirm the importance of explicit motion modelling for the perception of dynamic agents as conducted in \methodName.

%% file: diagrams/qualitative_comparison.tikz
{
\tikzset{
    static_world_tasks/.style={
        rectangle,
        fill = sns_blue!10,
        draw = sns_blue,
        text=sns_dark_grey,
        font=\fontsize{8}{10}\selectfont,
        text width=60pt,
        align=center
    },
    dynamic_world_tasks/.style={
        rectangle,
        fill = sns_orange!10,
        draw = sns_orange,
        text=sns_dark_grey,
        font=\fontsize{8}{10}\selectfont,
        text width=60pt,
        align=center
    },
    task/.style={
        rectangle,
        fill = task_dark_anthracite,
        draw = task_dark_anthracite,
        text=white,
        fill opacity=0.8,
        text opacity=1.0,
        font=\fontsize{8}{10}\selectfont,
        text width=80pt,
        minimum height=18pt,
        minimum width=80pt,
        align=center
    },
    dual_transformer/.style={
        rectangle,
        fill = sns_purple!10,
        draw = sns_purple,
        text = sns_purple,
        minimum width=50pt,
        minimum height=50pt,
        align=center
    },
    shared_backbone/.style={
        rectangle,
        fill = sns_purple!10,
        draw = sns_purple,
        text = sns_purple,
        minimum width=40pt,
        minimum height=80pt,
        align=center,
        fill opacity=0.8,
        text opacity=1.0
    },
    label/.style={
        rectangle,
        rounded corners=3pt,
        inner sep=2pt,
        fill = sns_dark_grey!10,
        draw = sns_dark_grey,
        text=sns_dark_grey,
        font=\fontsize{8}{10}\selectfont,
        align=left
    },
    label_text/.style={
        text=sns_dark_grey,
        font=\fontsize{8}{10}\selectfont,
        align=left
    },
    stream_interaction/.style={
        text=sns_green,
        font=\fontsize{8}{10}\selectfont,
        align=center,
        text width=45pt,
        rectangle,
        rounded corners=3pt,
        dashed,
        draw=sns_green,
    },
    dashed_round_rectangle/.style={
        rectangle,
        rounded corners=3pt,
        dashed,
        draw=black,
        align=center
    },
    dirline/.style={
            draw,
            -latex',
        },
    double_dirline/.style={
            draw,
            latex'-latex'
        },
    brace/.style={decorate, decoration={brace, amplitude=5pt}},
	bracenode_left/.style={midway, left=2pt, rotate=90, anchor=south},
	bracenode_right/.style={midway, right=2pt, rotate=90, anchor=north},
	bracenode_top/.style={midway, anchor=south, yshift=4pt},
	bracenode_bottom/.style={midway, anchor=north, yshift=-4pt},
}

\begin{tikzpicture}
    \newcommand\dashedLineMargin{10pt}
    \newcommand\compensationDistance{80pt}
    \newcommand\compensationWidth{50pt}
    \newcommand\compastedStateDistance{230pt}
    \newcommand\streamDistance{120pt}
    \newcommand\temporalDistance{35pt}
    \newcommand\sharedBackboneDistance{80pt}
    \newcommand\imageSize{95pt}
    \newcommand\taskDistance{100pt}
    \newcommand\taskImageSize{80pt}

    \node[sns_dark_grey,rotate=90]
    (uniad_label)
    {{UniAD~\cite{hu2023planning}}};

    \node[right of=uniad_label, node distance=10pt, anchor=west]
    (uniad1)
    {
        \includegraphics[width=\imageSize]{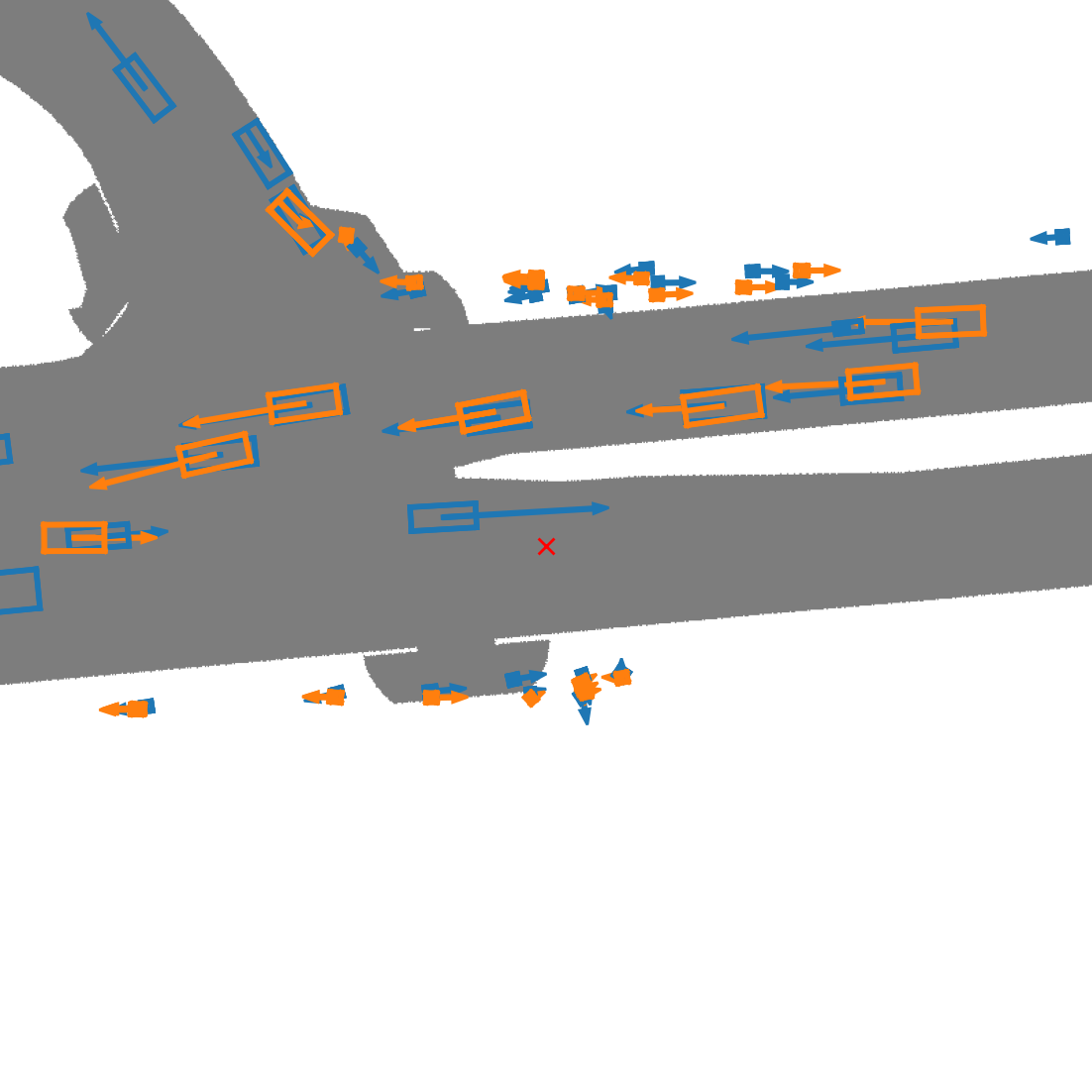}
    };
    \node[right of=uniad1, node distance=\imageSize-32pt,anchor=west]
    (uniad2)
    {
        \includegraphics[width=\imageSize]{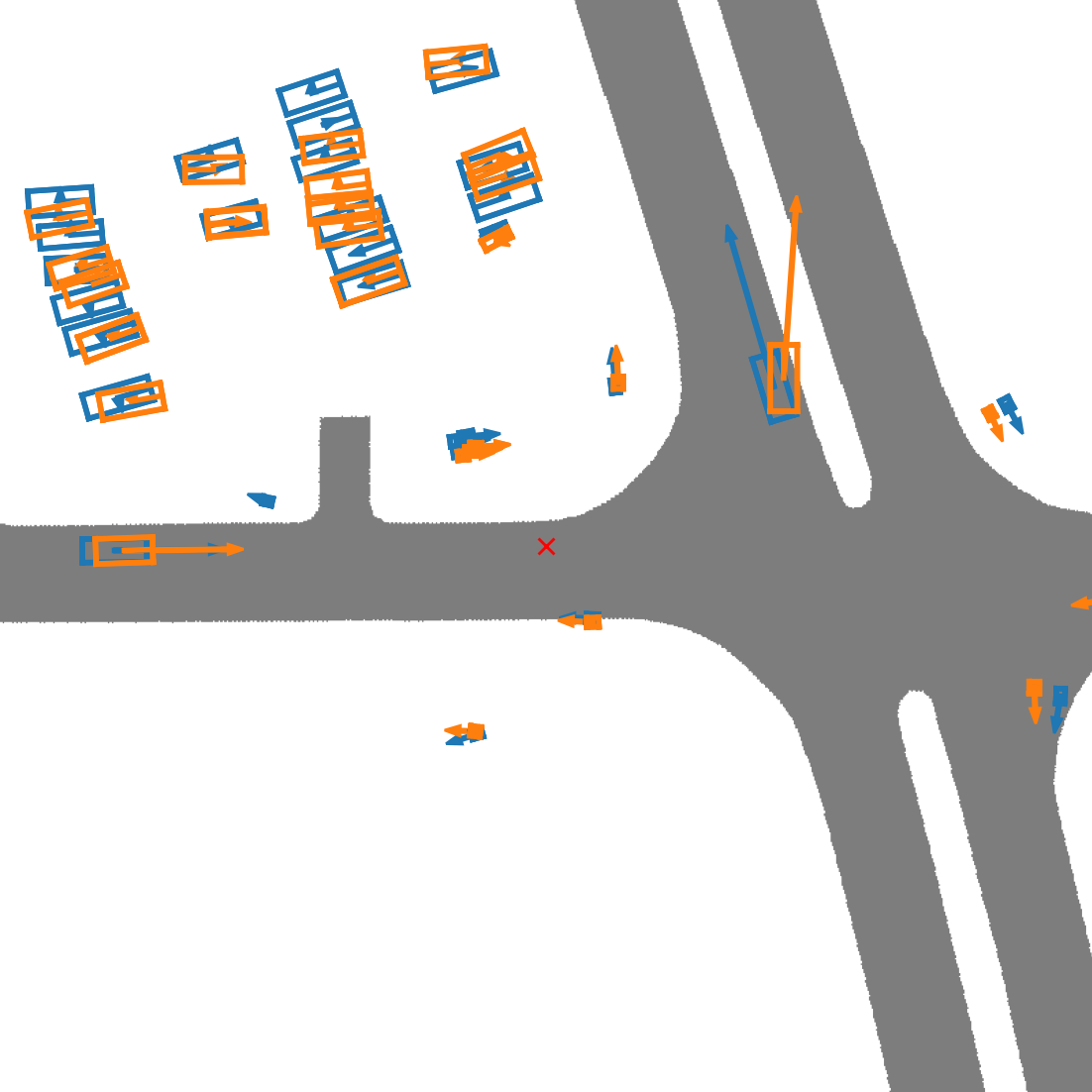}
    };

    \node[sns_dark_grey,rotate=90, left of=uniad_label, node distance=\imageSize+10pt]
    (dualad_label)
    {\textsc{DualAD} (ours)};

    \node[minimum width=\imageSize]
    (dualad1) at (uniad1 |- dualad_label)
    {
        \includegraphics[width=\imageSize]{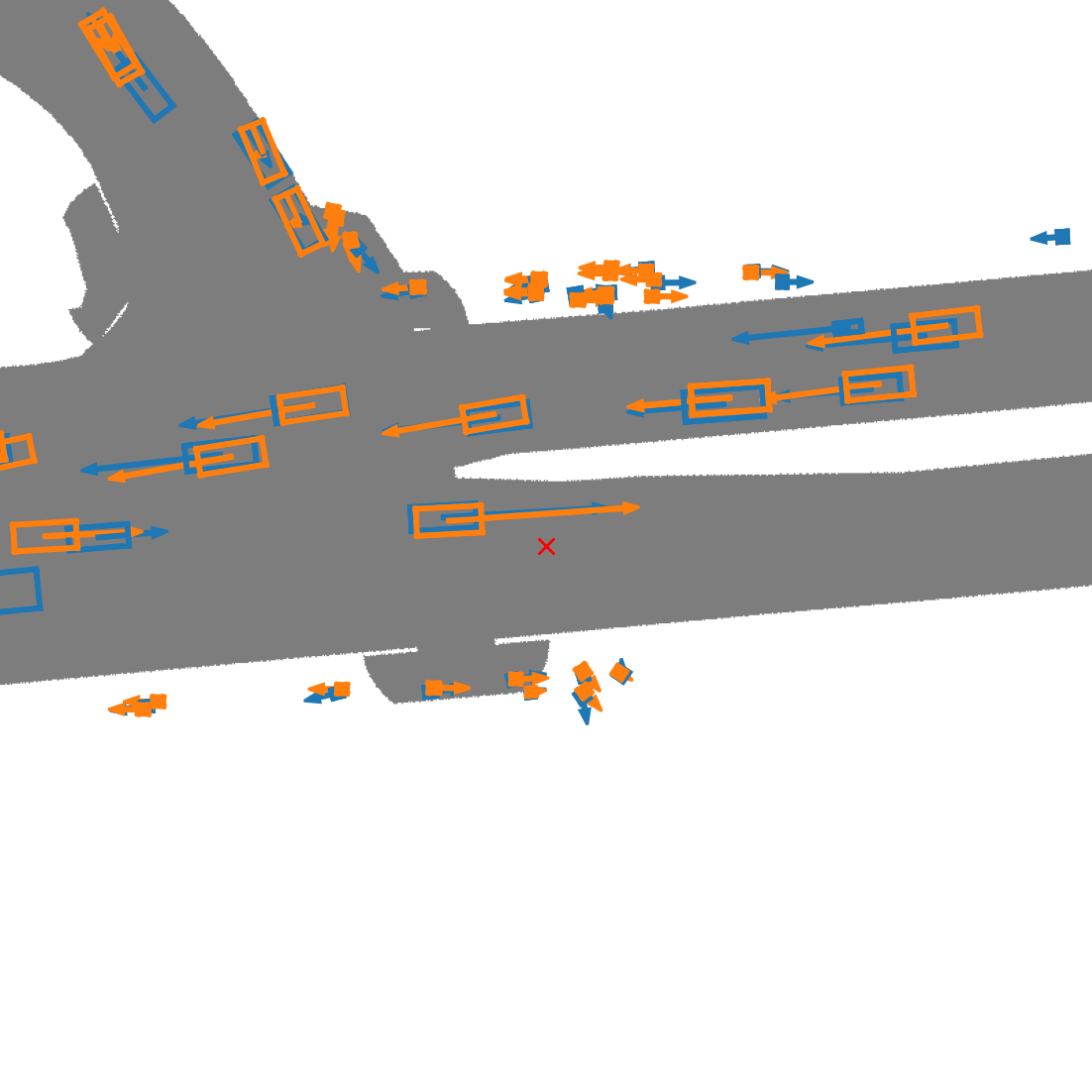}
    };
    \node[minimum width=\imageSize]
    (dualad2) at (uniad2 |- dualad_label)
    {
        \includegraphics[width=\imageSize]{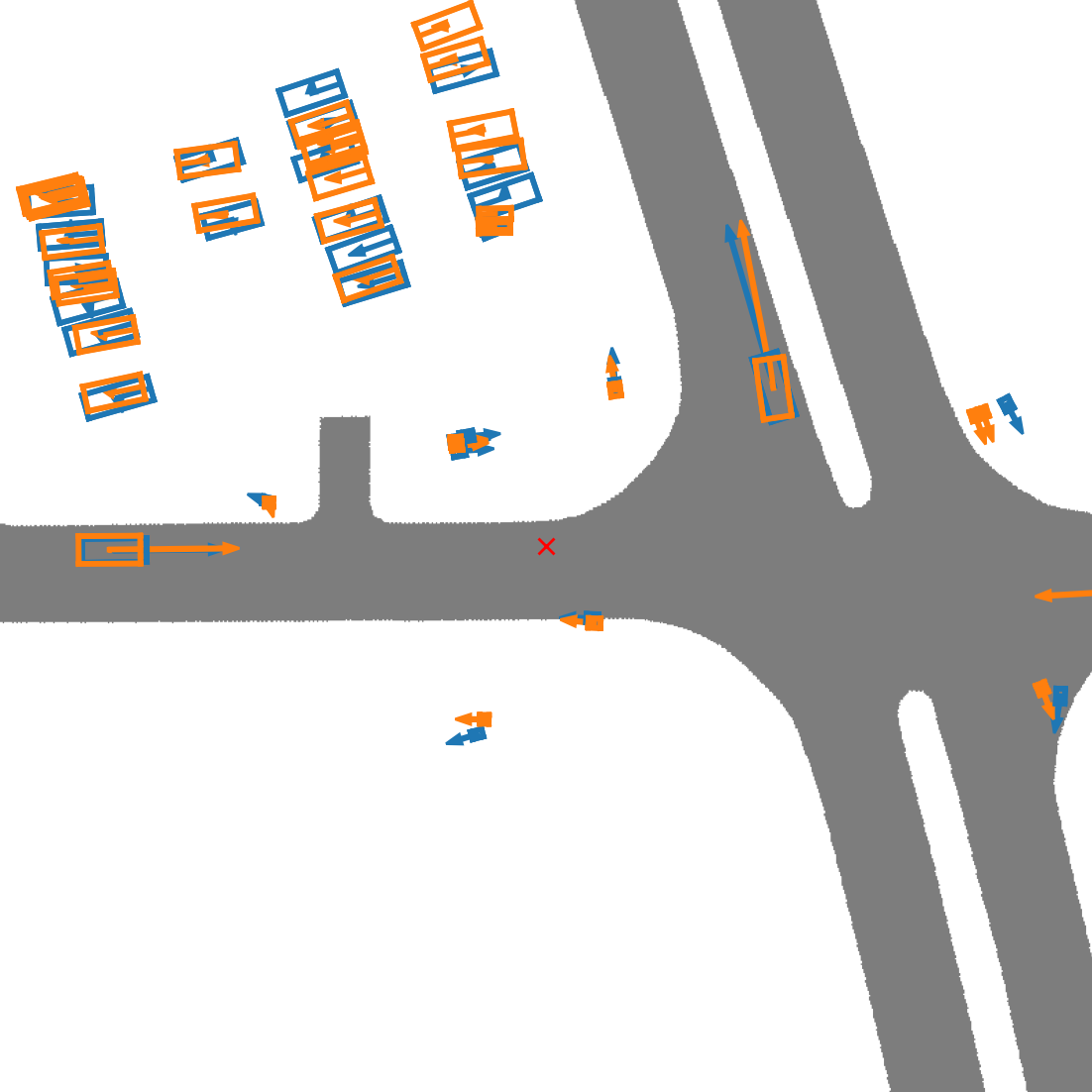}
    };

    \node[ellipse, minimum height=10pt, minimum width=25pt, draw=sns_dark_grey, thick,
            fill=sns_dark_grey!10, fill opacity=.1, rotate=2.5]
    (agenta_a_uniad) at ($(uniad1.center) + (-3pt, 2pt)$)
    {};
    \node[ellipse, minimum height=10pt, minimum width=25pt, draw=sns_dark_grey, thick,
            fill=sns_dark_grey!10, fill opacity=.1, rotate=2.5]
    (agenta_a_dualad) at ($(dualad1.center) + (-2pt, 2pt)$)
    {};

    \node[ellipse, minimum height=16pt, minimum width=25pt, draw=sns_dark_grey,
            fill=sns_dark_grey!10, fill opacity=.1, thick, rotate=105]
    (agenta_b_uniad) at ($(uniad2.center) + (20pt, 20pt)$)
    {};

    \node[ellipse, minimum height=10pt, minimum width=25pt, draw=sns_dark_grey,
            fill=sns_dark_grey!10, fill opacity=.1, thick, rotate=105]
    (agenta_b_dualad) at ($(dualad2.center) + (19pt, 19pt)$)
    {};

    \draw[dashed, sns_dark_grey] ($(dualad1.south)!0.5!(dualad2.south)$) -- ($(uniad1.north)!0.5!(uniad2.north)$);
    \draw[sns_dark_grey] ($(dualad_label)!0.5!(uniad_label)$) -- ($(dualad2.east)!0.5!(uniad2.east)$);

    \path[latex'-latex', dashed, sns_dark_grey, thick] ($(agenta_a_uniad.center) + (5pt, -7.5pt)$)
        edge[bend left=30] ($(agenta_a_dualad) + (5pt, 7.5pt)$);

    \path[latex'-latex', dashed, sns_dark_grey, thick] ($(agenta_b_uniad.center) + (5pt, -7.5pt)$)
        edge[bend left=30] ($(agenta_b_dualad) + (5pt, 7.5pt)$);

    \node[rectangle, draw, sns_blue, minimum height=5pt, minimum width=5pt, thick]
    (legend_rect_gt) at ($(dualad1.south west) + (0pt, 5pt)$)
    {};
    \node[right of=legend_rect_gt, label_text, xshift=2.5pt, align=left, text width=50pt]
    (legend_gt)
    {Ground-truth};

    \node[rectangle, draw, sns_orange, minimum height=5pt, minimum width=5pt, thick]
    (legend_rect_pred) at ($(legend_rect_gt) + (0pt, 10pt)$)
    {};
    \node[label_text, align=left, text width=50pt]
    (legend_pred) at (legend_gt |- legend_rect_pred)
    {Prediction};

\end{tikzpicture}
}

%% file: sections/5_conclusion.tex
\section{Conclusion}
This paper presents \methodName, a novel approach that explicitly models dynamic agents and static scene elements in a dual-stream design, where both can directly access the sensor information. 
This split explicitly accounts for object and ego motion within the dynamic stream, while only compensating for ego motion within the static stream.
The streams can interact by the newly introduced dynamic-static cross-attention, facilitating object detection by utilizing the inferred scene structure around the object. 

Our approach not only excels in early-stage perception tasks such as object detection and online map learning, but also demonstrates seamless integration with recent end-to-end models to tackle downstream tasks.
In our experimental evaluation, \methodName\ yields significant improvements over specialized models and reaches \ac{sota} performance for object detection, map segmentation, and multiple object tracking. Additionally, the integration into end-to-end models revealed improvements in motion prediction and planning, 
highlighting the importance of our dual-stream design for the entire functional chain.

Whilst our approach results in a robust and temporally consistent perception of the scene, the integration of other modalities such as LiDAR could boost the performance even further, especially combined with the potential of our model to flexibly move the belief state to different points in time to incorporate even unsynchronized sensors. The integration of additional information like traffic signs or traffic lights, as well as the integration of additional tasks such as depth-estimation or lane topology reasoning, remain promising research directions.

\boldparagraph{Acknowledgement} This work is a result of the joint research project STADT:up (19A22006O). The project is supported by the German Federal Ministry for Economic Affairs and Climate Action (BMWK), based on a decision of the German Bundestag. The author is solely responsible for the content of this publication.

%% file: sections/X_suppl.tex
\clearpage
\setcounter{page}{1}
\maketitlesupplementary
In this supplementary document, we first provide implementation details of our proposed approach. Furthermore, we present additional evaluation metrics for all perception tasks tackled by \methodName. Next, we discuss experimental findings regarding our design choices and temporal consistency.
Finally, we provide a detailed runtime analysis for different variants of our model and show additional qualitative results in the attached video file.

\section{Implementation Details}
Our work is built using the MMDetection3D framework\cite{mmdet3d2020}. Furthermore, we inherit various design choices from StreamPETR~\cite{wang2023exploring,streampetr_github}, UniAD~\cite{hu2023planning, uniad_github} and VAD~\cite{jiang2023vad, vad_github}. We truly thank all authors and contributors of those projects.
Our main model configuration closely follows StreamPETR~\cite{wang2023exploring, streampetr_github} since our dynamic stream design inherits the proposed query propagation through time as well as the geometric positional encodings for object-to-image cross-attention. All choices for the static stream are adopted from UniAD~\cite{hu2023planning}.

\boldparagraph{Data Augmentation}
We use the six surround camera images of nuScenes as input, down scaled to a resolution of $800 \times 320$ pixels. During training, we apply a random crop augmentation by choosing a random crop of $\SI{47}{\percent} - \SI{62.5}{\percent}$ of the image before down scaling.

\boldparagraph{Model Settings}
We use a VovNet-V2-99~\cite{lee2019energy} as image backbone and use the last two feature scales as input to the FPN~\cite{lin2017feature}. As in previous work, a latent dimension $L=256$ is adopted for all latent embeddings of our model. We use $|\mathcal{Q}_{\text{obj}}| = 900$ object queries consisting of the top-$k$ propagated from the previous time step with $k=256$ and 644 newly spawned objects queries respectively. For the BEV-queries we follow UniAD~\cite{hu2023planning} and use $|\mathcal{Q}_{\text{BEV}}| = 200 \times 200$. The used detection range is $ [\SI{-51.2}{\meter}, \SI{51.2}{\meter}]$ for $x$ and $y$ direction, resulting in an effective grid resolution of $\SI{0.512}{\meter}$.

The proposed dual-stream transformer utilizes six consecutive layers and performs self-attention within $\mathcal{Q}_{\text{obj}}$, cross-attention of $\mathcal{Q}_{\text{obj}}$, temporal self-attention of $\mathcal{Q}_{\text{BEV}}$ and the interpolated grid queries from the last frame~\cite{li2022bevformer, hu2023planning}, cross-attention from $\mathcal{Q}_{\text{BEV}}$ to image features as in~\cite{li2022bevformer} and dynamic-static cross-attention of $\mathcal{Q}_{\text{obj}}$ and $\mathcal{Q}_{\text{BEV}}$. For the dynamic object cross-attention to the image features we only choose the highest spatial resolution feature scale as in~\cite{wang2023exploring, streampetr_github}.

During training, we adopt query-denoising~\cite{li2022dn} and streaming video training as proposed in~\cite{wang2023exploring} to accelerate the convergence as well as Flash-Attention~\cite{dao2022flashattention} to reduce the memory requirements. With the aforementioned settings, the training for 24 epochs requires $\SI{18}{\giga \byte}$ of GPU memory and takes approximately one day for stage-I and two days for stage-2 on eight NVIDIA A100 GPUs.

\section{Performance Evaluation}
We provide evaluation results for various model configurations of \methodName. As in the main paper, we indicate all stage-I models that are trained on perception tasks only \eg object detection, map segmentation and multiple object tracking as \methodNameSOne\ and the configuration that was trained on all tasks in an end-to-end fashion as \methodNameSTwo\ respectively.
Furthermore, we adopt the notation introduced in~\tabref{table:interaction_design} to denote different configurations of \methodName. The version marked with $\emptyset$ does not use the proposed dynamic-static cross-attention, while $\updownarrow$ describes a version that uses bidirectional stream interaction by using global attention for the interaction from the static to the dynamic stream. The version of our model that is trained on the reduced sensor set by using front and back facing cameras in an alternating fashion only is indicated with $\circleddash$.

\boldparagraph{Object Detection}
A detailed evaluation of all metrics specified in the official nuScenes detection benchmark~\cite{nuscenes_detection_benchmark} is shown in~\tabref{sup:tab:detection}. For detailed metric definitions, we kindly refer to~\cite{caesar2020nuscenes, nuscenes_detection_benchmark}.
\begin{table*}
    \centering
    \caption{Object Detection Results.}
    \label{sup:tab:detection}

    \begin{tabular}{lccccccccc}
        \toprule
        Name                            & \vtop{\hbox{\strut Temporal}\hbox{\strut BEV}} & \vtop{\hbox{\strut Sensor}\hbox{\strut Drop}} & mAP$\uparrow$  & mATE$\downarrow$ & mASE$\downarrow$ & mAOE$\downarrow$ & mAVE$\downarrow$ & mAAE$\downarrow$ & \highlight{NDS$\uparrow$}  \\
        \midrule
        \methodNameSOne                 & \xmark                                         & \xmark                                        & 46.93          & 0.62             & 0.27             & 0.39             & 0.27             & \textbf{0.18}    & \highlight{56.16}          \\
        \methodNameSOne$\emptyset$      & \cmark                                         & \xmark                                        & 47.74          & 0.62             & 0.27             & 0.45             & 0.28             & 0.19             & \highlight{55.78}          \\
        \methodNameSOne$\updownarrow$   & \cmark                                         & \xmark                                        & 49.37          & 0.58             & 0.27             & 0.39             & 0.26             & 0.20             & \highlight{57.65}          \\
        \methodNameSOne$\mathparagraph$ & \cmark                                         & \xmark                                        & 48.21          & 0.60             & 0.27             & \textbf{0.32}    & 0.28             & 0.20             & \highlight{57.44}          \\
        \methodName$\circleddash$       & \cmark                                         & \cmark                                        & 42.86          & 0.65             & 0.28             & 0.47             & 0.32             & 0.19             & \highlight{52.22}          \\
        \methodNameSOne                 & \cmark                                         & \xmark                                        & \textbf{49.56} & 0.58             & \textbf{0.26}    & 0.40             & \textbf{0.26}    & 0.20             & \highlight{\textbf{57.81}} \\
        \methodNameSTwo                 & \cmark                                         & \xmark                                        & 48.16          & \textbf{0.57}    & 0.27             & 0.41             & 0.29             & 0.19             & \highlight{56.68}          \\
        \bottomrule
    \end{tabular}

\end{table*}

\boldparagraph{Map Segmentation}
The results for all model configurations on map segmentation are shown in table~\tabref{sup:tab:map_seg}. The evaluation is performed for four different classes as proposed in UniAD~\citep{hu2023planning} and we compute the \ac{iou} between predicted and ground truth segmentation maps.
\begin{table*}[]
    \centering
    \caption{Map Segmentation Results.}
    \label{sup:tab:map_seg}
    \begin{tabular}{lcccccc}
        \toprule
        Name                            & \vtop{\hbox{\strut Temporal}\hbox{\strut BEV}} & \vtop{\hbox{\strut Sensor}\hbox{\strut Drop}} & \highlight{Lanes$\uparrow$} & Drivable$\uparrow$ & Divider$\uparrow$ & Crossing$\uparrow$ \\
        \midrule
        \methodNameSOne                 & \xmark                                         & \xmark                                        & \highlight{31.73}           & 67.52              & 26.57             & 10.99              \\
        \methodNameSOne$\emptyset$      & \cmark                                         & \xmark                                        & \highlight{33.97}           & 69.35              & 29.49             & 12.33              \\
        \methodNameSOne$\updownarrow$   & \cmark                                         & \xmark                                        & \highlight{33.86}           & 67.78              & 29.11             & 12.18              \\
        \methodNameSOne$\mathparagraph$ & \cmark                                         & \xmark                                        & \highlight{34.26}           & 69.71              & 29.71             & \textbf{13.87}     \\
        \methodName$\circleddash$       & \cmark                                         & \cmark                                        & \highlight{31.53}           & 66.60              & 26.77             & 10.14              \\
        \methodNameSOne                 & \cmark                                         & \xmark                                        & \highlight{\textbf{34.68}}  & \textbf{70.50}     & \textbf{30.29}    & 12.82              \\
        \methodNameSTwo                 & \cmark                                         & \xmark                                        & \highlight{34.17}           & 70.01              & 29.96             & 12.25              \\
        \bottomrule
    \end{tabular}
\end{table*}

\boldparagraph{Multiple Object Tracking}
A detailed evaluation of all metrics specified in the official nuScenes tracking benchmark~\cite{nuscenes_tracking_benchmark} is shown in~\tabref{sup:tab:tracking}. For detailed metric definitions, we kindly refer to~\cite{caesar2020nuscenes, nuscenes_tracking_benchmark}.
\begin{table*}[]
    \centering
    \caption{Multiple Object Tracking Results.}
    \label{sup:tab:tracking}

    \begin{adjustbox}{width=\textwidth,center}
        \begin{tabular}{lccccccccccccc}
            \toprule
            Name                            & \vtop{\hbox{\strut Temporal}\hbox{\strut BEV}} & \vtop{\hbox{\strut Sensor}\hbox{\strut Drop}} & \highlight{AMOTA$\uparrow$} & AMOTP$\downarrow$ & RECALL$\uparrow$ & MT$\uparrow$  & ML$\downarrow$ & FAF$\downarrow$ & IDS$\downarrow$ & FRAG$\downarrow$ & TID$\downarrow$ & LGD$\downarrow$ \\
            \midrule
            \methodNameSOne                 & \xmark                                         & \xmark                                        & \highlight{51.63}           & 1.16              & 59.69            & 3006          & 2104           & 49.08           & 658             & 671              & 1.25            & 1.96            \\
            \methodNameSOne$\emptyset$      & \cmark                                         & \xmark                                        & \highlight{51.94}           & 1.13              & 59.27            & 3107          & 2148           & 48.37           & 769             & 657              & 1.24            & 1.84            \\
            \methodNameSOne$\updownarrow$   & \cmark                                         & \xmark                                        & \highlight{54.39}           & 1.09              & \textbf{61.11}   & 3232          & 2077           & 46.76           & \textbf{588}    & \textbf{580}     & 1.14            & 1.70            \\
            \methodNameSOne$\mathparagraph$ & \cmark                                         & \xmark                                        & \highlight{52.32}           & 1.13              & 60.74            & 3272          & \textbf{1908}  & 49.46           & 726             & 695              & 1.09            & 1.67            \\
            \methodName$\circleddash$       & \cmark                                         & \cmark                                        & \highlight{44.39}           & 1.22              & 53.96            & 2658          & 2476           & 53.57           & 940             & 936              & 1.44            & 2.01            \\
            \methodNameSOne                 & \cmark                                         & \xmark                                        & \highlight{\textbf{55.09}}  & \textbf{1.09}     & 60.71            & \textbf{3279} & 2031           & \textbf{46.21}  & 663             & 588              & 1.12            & 1.70            \\
            \methodNameSTwo                 & \cmark                                         & \xmark                                        & \highlight{52.57}           & 1.11              & 59.62            & 3159          & 2166           & 46.25           & 774             & 593              & \textbf{1.07}   & \textbf{1.61}   \\
            \bottomrule
        \end{tabular}
    \end{adjustbox}
\end{table*}

\boldparagraph{Motion Prediction}
A detailed evaluation of motion prediction results for all dynamic classes of the nuScenes dataset~\cite{caesar2020nuscenes} is shown in~\tabref{sup:tab:motion}. As in UniAD~\cite{hu2023planning} we adopt a confidence threshold $c_{\text{motion}} = 0.4$ during inference to select object queries that are passed to the motion head.
\begin{table}
    \centering
    \caption{Motion prediction results of \methodNameSTwo\ for all object categories on the nuScenes benchmark~\cite{nuscenes_tracking_benchmark}.}
    \label{sup:tab:motion}
    \begin{tabular}{lcccc}
        \toprule
        Name       & \highlight{EPA}$\uparrow$ & minADE$\downarrow$ & minFDE$\downarrow$ & miss rate$\downarrow$ \\
        \midrule
        Car        & \highlight{54.97}         & 0.35               & 0.39               & 0.035                 \\
        Truck      & \highlight{43.12}         & 0.37               & 0.38               & 0.017                 \\
        Bus        & \highlight{42.31}         & 0.51               & 0.56               & 0.057                 \\
        Trailer    & \highlight{26.79}         & 0.55               & 0.53               & 0.017                 \\
        Pedestrian & \highlight{45.28}         & 0.46               & 0.61               & 0.003                 \\
        Motorcycle & \highlight{39.02}         & 0.32               & 0.37               & 0.011                 \\
        Bicycle    & \highlight{36.89}         & 0.28               & 0.30               & 0.002                 \\
        \bottomrule
    \end{tabular}
\end{table}

\subsection{Discussion of design choices}
The extensive ablations on various tasks and configurations of our proposed approach (see~\tabref{sup:tab:detection},~\tabref{sup:tab:map_seg},~\tabref{sup:tab:tracking}) validate our different design choices.
Our model consistently benefits from temporal information and the proposed dynamic-static cross-attention. Adding another cross-attention block to perform bidirectional interaction does not significantly improve the performance of static map perception or overall temporal consistency, which is in line with our hypothesis that map segmentation might not benefit from dynamic agent perception.
We leave the investigation of other interaction designs and other dense tasks that depend on the dynamic agent perception \eg free-space estimation for future work.

The stage-II configuration of our approach yields a slightly decreased perception performance when compared to the stage-I model.
This could result from the fact that in stage-II the model might focus on certain scene parts that are more relevant for the currently planned trajectory. Additionally, a fast detection of highly dynamic agents and temporal consistency might be crucial for longer planning horizons, which is in line with the improvements of the stage-II model in terms of \ac{tid} and \ac{lgd} as shown in~\tabref{sup:tab:tracking}.

The \methodNameSOne$\circleddash$ version of our model that only has access to front or back facing cameras in an alternating fashion maintains high temporal consistency by query propagation even without sensor data for some areas in the scene. We refer to the attached video for a qualitative example. However, the initial detection of newly appeared object is not possible if no sensor data for the corresponding scene area is available or consistent tracking might be challenging, especially for highly dynamic or hardly visible agents in the scene. Since our base model especially improves over previous approaches in such challenging cases, this explains the drop in perception performance by $\SI{-6.7}{\ac{map}}$ and $\SI{-10.7}{\ac{amota}}$ respectively (see~\tabref{sup:tab:detection},~\tabref{sup:tab:tracking}).

\subsection{Runtime Analysis}
\label{sub:sec:runtime}
We evaluate the runtime of the stage-II configuration of \methodName.
The results of the entire system as well as the runtime of the intermediate task modules are shown in~\tabref{sup:tab:runtime}. \methodName\ runs with $\SI{4.12}{FPS}$ on a single NVIDIA A100 GPU.
The dual stream transformer uses a significant amount of the model's total runtime due to the expensive attention operations from object queries and \ac{bev}-queries to sensor data. Since all downstream tasks use the resulting representations, the task heads only add a small amount of additional runtime.
Please note that our codebase contains various operations which could be further optimized.
However, improving the runtime and memory requirements of end-to-end approaches remains a challenging topic for large scale application of such approaches.

\begin{table}
    \centering
    \caption{Runtime evaluation of \methodNameSTwo\ on a single NVIDIA-A100 for 500 frames of the nuScenes validation set. Misc describes various non-optimized computations \eg bounding box decoding and positional encodings.}
    \label{sup:tab:runtime}
    \begin{tabular}{lc}
        \toprule
        Module                  & Runtime (ms) $\downarrow$ \\
        \midrule
        Image Backbone          & 19                        \\
        Dual Stream Transformer & 59                        \\
        Detection Head          & 17                        \\
        Map Head                & 23                        \\
        Motion Head             & 24                        \\
        Planning Head           & 39                        \\
        Misc                    & 80                        \\
        \midrule
        Total                   & 242                       \\
        \bottomrule
    \end{tabular}

\end{table}

\subsection{Integration to VAD~\cite{jiang2023vad}}
The version of our model that is based on VAD~\cite{jiang2023vad} is denoted as \textsc{DualVAD}, please note that we report the performance of the stage-II model to allow for a fair comparison with the provided model in~\cite{vad_github}.
In contrast to the other configurations, VAD relies on a ResNet-50~\cite{he2016deep} as image backbone, an input resolution of $1280 \times 720$ pixels~\cite{jiang2023vad, vad_github} and a shorter detection range around the ego vehicle of $ [\SI{-30}{\meter}, \SI{30}{\meter}]$ in $x$ and  $[\SI{-15}{\meter}, \SI{15}{\meter}]$ in $y$ respectively. A detailed evaluation of the perception performance is given in~\tabref{sup:tab:vad_perception}. \textsc{DualVAD} outperforms VAD~\cite{jiang2023vad} by $+2.7$ \ac{map} for dynamic object perception and achieves a slightly higher vectorized map perception performance while also heavily improving downstream tasks such as motion prediction (see~\tabref{table:results_prediction}) and open-loop planning (see~\tabref{table:results_planning}). The runtime of \textsc{DualVAD-II} is shown in~\tabref{sup:tab:runtime_vad}. In this configuration, our model runs at $\SI{3.32}{FPS}$ on a single NVIDIA A100 GPU. Due to the larger input image size, the runtime of the dual stream transformer increases significantly as compared to our base configuration.
\begin{table}
    \centering
    \caption{Runtime evaluation of \textsc{DualVAD-II} on a single NVIDIA-A100 for 500 frames of the nuScenes validation set. Misc describes various non-optimized computations \eg bounding box decoding and positional encodings.}
    \label{sup:tab:runtime_vad}
    \begin{tabular}{lc}
        \toprule
        Module                  & Runtime (ms) $\downarrow$ \\
        \midrule
        Image Backbone          & 11                        \\
        Dual Stream Transformer & 114                       \\
        Detection Head          & 58                        \\
        Map Head                & 4                         \\
        Motion Head             & 7                         \\
        Planning Head           & 3                         \\
        Misc                    & 104                       \\
        \midrule
        Total                   & 301                       \\
        \bottomrule
    \end{tabular}

\end{table}

\begin{table*}
    \centering
    \caption{Perception Results for VAD~\cite{jiang2023vad} based models. *Results taken
        from official repository. mAP$_{\text{Map}}$ denotes the mAP of vectorized map perception as defined in~\cite{liao2022maptr, jiang2023vad}.}
    \label{sup:tab:vad_perception}

    \begin{tabular}{lccccccc|c}
        \toprule
        Name                         & mAP$\uparrow$  & mATE$\downarrow$ & mASE$\downarrow$ & mAOE$\downarrow$ & mAVE$\downarrow$ & mAAE$\downarrow$ & \highlight{NDS$\uparrow$}  & mAP$_{\text{Map}}$$\uparrow$ \\
        \midrule
        VAD~\cite{jiang2023vad}*     & 33.92          & 0.59             & 0.28             & \textbf{0.53}    & 0.40             & \textbf{0.23}    & \highlight{46.02}          & 47.5               \\
        \textbf{\textsc{DualVAD-II}} & \textbf{36.64} & \textbf{0.59}    & \textbf{0.27}    & 0.57             & \textbf{0.35}    & 0.23             & \highlight{\textbf{48.00}} & \textbf{47.9}      \\
        \bottomrule
    \end{tabular}

\end{table*}

\subsection{Qualitative Results}
Together with this document, we provide a video that shows qualitative results of our approach for various scenes from the nuScenes validation set. Those include complex traffic scenes, a setting with unsynchronized sensors, challenging lighting and adverse weather conditions and results for the vectorized map representation. \methodNameSTwo\ demonstrates robust and consistent performance for all perception tasks, as well as downstream performance for motion prediction and open-loop planning.